\documentclass[Afour,sageh,times]{sagej}
\usepackage{tabularx}
\usepackage{multirow}
\usepackage{enumitem}

  \usepackage{multirow}
  \graphicspath{{img/}}
\usepackage{color}

\usepackage{amsmath}
\usepackage{amssymb}
\usepackage{gensymb}
\usepackage{flushend}

\usepackage{array}

  \usepackage[caption=false,font=normalsize,labelfont=sf,textfont=sf]{subfig}
\usepackage{url}
\newcommand{\normal}{\hat{\bm{n}}} %\normal}
\renewcommand{\vec}[1]{{\bm{#1}}}
\newcommand{\bg}{b_\mathrm{g}}
\newcommand\Cs{$\mathcal{C}$-space }

\newcommand\Csdot{$\mathcal{C}$-space. }
\newcommand\Cscoma{$\mathcal{C}$-space, }

\newcommand\Cf{\mathcal{C}_\text{free}}

\usepackage{algorithm}
\usepackage[noend]{algorithmic}

\usepackage{bm}
\usepackage{tabularx}
\usepackage{xcolor}

\usepackage{booktabs, multirow} % for borders and merged ranges
\usepackage{soul}% for underlines
\usepackage{makecell}
\usepackage[
  pdftitle={Motion Generation With Environmental Constraints},
  pdfauthor={El\H{o}d P\'all and Oliver Brock},
  pdfkeywords={grasping, open-loop grasping, environmental constraints, vision-free grasping, bin-picking}
  bookmarks=true,hidelinks]{hyperref}
  
\hypersetup{draft}
\usepackage{flushend} 

\begin{document}

%======================================================================
% IJRR specifics
%======================================================================
\def\volumeyear{2021}
\def\journalname{The International Journal of Robotics Research}
\setcounter{secnumdepth}{3}

%======================================================================
% PDF Metadata (from https://elektro.robotics.tu-berlin.de/index.php/Publications)
%======================================================================

%\hypersetup{pdftitle={Motion Generation With Environmental Constraints},
%	pdfauthor={El\H{o}d P\'all and Oliver Brock},
%	pdfkeywords={Robotics, Motion Planning, Motion Generation, Motion Control, State Uncertainty}
%}

%======================================================================
% Title
%======================================================================

\runninghead{P\'all et al.}
\title{Motion Generation With Environmental Constraints}
\author{El\H{o}d P\'all\affilnum{1}and Oliver Brock\affilnum{1,2}}

\affiliation{\affilnum{1}Robotics and Biology Laboratory, Technische Universit{\"a}t Berlin, Germany \affilnum{2}Science of Intelligence, Research Cluster of Excellence, Berlin, Germany}
\corrauth{Oliver Brock, Robotics and Biology Laboratory, Technische Universit{\"a}t Berlin, Marchstra{\ss}e 23, 10587 Berlin, Germany.}
\email{oliver.brock(at)tu-berlin.de}

% As a general rule, do not put math, special symbols or citations
% in the abstract or keywords.
\begin{abstract}
Robot motion planning faces challenges in high-dimensional spaces and uncertain environments, often constrained by the need for collision-free motions. We advocate an alternative approach, Environmental Constraint Exploitation (ECE), where deliberate contact with the environment simplifies planning by reducing dimensionality and computational complexity. By integrating ECE into motion planning algorithms, we bias exploration to task-relevant regions and leverage contact for uncertainty reduction to improve robustness during execution. We evaluate ECE benefits with RRT-based planners and demonstrate their practical benefits in a real-world application. This work consolidates and extends prior research, showcasing how ECE simplifies motion planning while enhancing adaptability and performance in complex environments.
\end{abstract}

% Note that keywords are not normally used for peerreview papers.
%\begin{IEEEkeywords}
%\end{IEEEkeywords}

% make the title area
\maketitle

% For peer review papers, you can put extra information on the cover
% page as needed:
% \ifCLASSOPTIONpeerreview
% \begin{center} \bfseries EDICS Category: 3-BBND \end{center}
% \fi
%
% For peerreview papers, this IEEEtran command inserts a page break and
% creates the second title. It will be ignored for other modes.
%\IEEEpeerreviewmaketitle

%======================================================================
\section{Introduction}
%======================================================================
Robot motion planning is one of the foundational and most extensively investigated research areas in robotics \citep{lavalle_planning_2006, barraquand1991robot}. In this paper, we explore an aspect of motion planning that has remained substantially unexplored: motion planning in contact with the environment. Leveraging the environment provides substantial advantages for the efficiency of motion planning and the range of problems to which motion planning can be successfully applied.

Standard definitions of motion planning permit the boundaries of the robot and the obstacle to intersect (for example,\citep{lavalle_planning_2006}, p. 156). This means that the definition does not rule out motion in contact between a robot and its environment. Pioneering work leveraged such contact to generate successful plans even when the uncertainty of motion execution is too large to reach the goal region without contact \citep{lozano-perez_automatic_1984}. But following this seminal, early work, the majority of researchers developed motion planning algorithms that avoid contact between the robot and the environment. We believe that this overlooks a fundamental opportunity.

The two most important challenges in developing motion planning algorithms are a) high dimensionality of the configuration space and b) uncertainties of various kinds (sensing, actuation, modeling). The deliberate exploitation of contact during robot motion can alleviate the challenges associated with both of these challenges. First, contact can reduce or even eliminate uncertainty, as information available about the environment can be used to reduce uncertainty about the robot's state.  Second, the surface of obstacles can serve as a lower-dimensional and simpler representation of the connectivity of the C-space, similar to bug algorithms 
\citep{barraquand1991robot}.

In this paper, we present motion planning algorithms that exploit contact as a central algorithmic element. We view contact as the exploitation of environmental constraints \cite{eppner_exploitation_2015}. An environmental constraint can be any feature of the environment that limits the motion of the robot, either physically or via a feature in the environment exploited by a controller. We also present their application in three different domains: motion planning, tactile localization, and grasping. For each of the algorithms and each of the applications, we explain how environment constraint exploitation (ECE) either greatly increases the planning efficiency or enables solving problems that otherwise would be beyond a practical solution.

The algorithms and experimental results presented in this paper illustrate the power of contact exploitation in motion planning. They provide a coarse exploration of the landscape of contact-based motion planning opportunities. %\todo{Can we say something more about the future potential? -- We answer this at the end of the intro \\ Maybe even something about a contrast to deep learning motion generation? In any case, something more here... -- feel free to propose something, but I don't see the need for it here yet.}

\subsection{Novel Contributions}

The research presented here collects several conference papers for the first time in an archival format while situating them in a larger picture of EC-based motion generation:

\begin{itemize}
\item \cite{sieverling_interleaving_2017} devised an efficient contact-based motion planning approach in \Cs under uncertainty without explicitly representing ECs by random sampling motion in free space and contact. 

\item \cite{pall_contingent_2018} proved that even if transitioning between EC regions is probabilistic due to a large amount of motion uncertainty, contact sensing can reduce the uncertainty during planning, and probabilistic transitions can be handled efficiently by reusing already computed partial solutions.

\item \cite{pall_analysis_2021} discovered a novel EC and characterized its use for grasping from piles of nearly identical objects with an explorative empirical study.
\end{itemize}

 As novel contributions, we explain how the specific properties of planning problems match with different implementations of ECE, as summarized in Table~\ref{tab: 1}. We also detail how to integrate different ECE implementations into classical motion planning algorithms and evaluate them across relevant planning problems.

First, we focus on integrating ECE into classical motion planning algorithms. We explain how the Contact Exploiting RRT (CERRT) planner computes a partial cover of EC regions between a start and goal by interleaving motion in free and contact spaces in Section~\ref{sec: cerrt}. Then, we extend the previously published CERRT planner by integrating workspace information from ECE-graphs into planning to maintain computational practicality in high-dimensional and \textit{complex} $\mathcal{C}$-spaces. Afterward, we explain how the Contingency CERRT planner computes a probabilistic ECE graph in Section~\ref{sec: concerrt}. As a novel contribution, we extend the previously published algorithm with workspace information-guided exploration. Finally, we evaluate all previously published and novel motion planners with new experiments comparing performance along the spectra of \Cs complexity and motion uncertainty. 

Secondly, we focus on a practical application of ECE and analyze its benefits through a real-world bin-picking application. In our previous work, we discovered a new EC and showed its benefits for grasping. Now, we extend our understanding of the novel EC by characterizing it with a new hypothesis-driven study in Section~\ref{sec: gece}.

The above-mentioned sections can be read individually to gain insights into problem-specific ECE implementations and applications. Still, they build on each other to support our main argument: contact-based environmental constraint exploitation simplifies motion planning. Even though the paper focuses primarily on motion planning, the ECE concept generally applies to simplifying control, perception, and building robotic systems.

\begin{table}[tb]\centering
\scriptsize
\begin{tabular}{ccccc}\toprule
\multicolumn{4}{c}{\textbf{ECE Implementation in Motion Planning}} \\\midrule
\multicolumn{2}{c}{\multirow{2}{*}{}} & \multicolumn{2}{c}{\textbf{\Cs complexity}} \\\cmidrule{3-4}
& &low &high \\\cmidrule{2-4}
\multirow{2}{*}{\makecell{\textbf{Motion}\\ \textbf{uncertainty}}} & low & 
\makecell{%Sequencing funnels\\
Section~\ref{sec: cerrt}} & 
\makecell{%EC-context guided exploration\\
Section~\ref{sec: ceet}}  \\\cmidrule{2-4}
&high & 
\makecell{%Contact-event-based contingencies\\
Section~\ref{sec: concerrt} }&
\makecell{Section~\ref{sec: concerrt} + \ref{sec: ceet}} \\\midrule%\hline%\bottomrule
\multicolumn{4}{c}{\textbf{ECE Applications for Motion Generation}} \\\midrule
\multicolumn{2}{c}{Motion planning} & \multicolumn{2}{c}{Section~\ref{sec: cerrt}}  \\
\multicolumn{2}{c}{Tactile localization} & \multicolumn{2}{c}{Section~\ref{sec: concerrt}} \\
\multicolumn{2}{c}{Grasping} & \multicolumn{2}{c}{Section~\ref{sec: gece}} \\
\bottomrule
\end{tabular}
\caption{Paper's structure with respect to main challenges in motion generation and contributions supporting general applicability of \emph{environmental constraint exploitation}.}\label{tab: 1}
\end{table}

%\begin{table}[!htp]\centering
%\caption{\todo{combined like Tab.I or separate with this?} Applications using environmental constraint exploitations}\label{tab: }
%\scriptsize
%\begin{tabular}{lrr}\toprule
%\textbf{Applications} & \textbf{Section}\\\midrule
%Motion planning & \ref{sec: cerrt}  \\
%Tactile localization & \ref{sec: concerrt} \\
%Grasping & \ref{sec: gece} \\
%\bottomrule
%\end{tabular}
%\end{table}

%======================================================================
\section{Related Work}
\label{sec:related work}
%======================================================================
With environmental constraint exploration, we overcome both challenges of motion planning: maintaining computation practicable for planning and computing motion plans that are robust against uncertainty. Thus, we divide related work into three parts. First, we present methods that reduce computation complexity on the spectrum between structural-context unaware and aware planning methods in increasing order of information usage since we use ECEs' structural context to simplify planning. Secondly, we discuss uncertainty handling methods on the spectrum between uncertainty-unaware planning and complete search in belief space, in increasing order of their treatment of uncertainty because we use ECE to reduce uncertainty. Finally, we present EC types and ways these were exploited for motion generation. 

\subsection{Motion Planning in Complex Environments.} 

We want to simplify planning in complex environments by representing a robot's workspace with ECs because such a representation is low-dimensional and can efficiently guide configuration space exploration. Thus, we discuss planning methods on the spectrum from exploiting prior workspace information to exploration without using any priors. 

\emph{Exploitation}-based planners compute a plan relying solely on information available before planning without using any information gained during planning. For example, a visibility map is computed from the geometrical properties of the environment. As another example, virtual potential fields or navigation functions exploit structural information encoded in their potential function. Computing a visibility map or a navigation function is difficult for high-dimensional non-linear \Cscoma but afterward, motion planning is trivially, for example, by following a gradient.       

\emph{Exploration}-based planners use no prior information but collect it via configuration space exploration. For example, the original multi-query PRM planner performs pure exploration when using uniform random sampling. However, its single-query sibling, the RRT planner explores \Cs biased to large unexplored regions implicitly. 

\emph{Guided-exploration}-based planners use a variety of informed sampling techniques that use a fixed or adaptive heuristic~\citep{burns_single-query_2007, hauser_multi-modal_2010, kingston_informing_2020} exploiting information gained during exploration to efficiently explore difficult regions of the configuration space. For example, narrow passages can be explored with bridge sampling. A commonly used bias for single-query planners is the goal bias. One can seldom replace a random sample with the goal configuration or try connecting new nodes to the goal. Both heuristics increase planning efficiency to connect a given start to a goal. Other biasing techniques target issues related to the Voronoi bias of RRT planners or escaping bug traps. For example, a planner may choose to omit expanding nodes in a tree after their expansion has failed multiple times, a planner can also use a bi-directional search one from the start and one from goal configuration~\citep{kuffner_rrt-connect_2000}, or even grow forests of search trees~\citep{hang_pre-grasp_2019}.

Structural information of a workspace is valuable to guide exploration and consequently increase a planner's efficiency in finding a solution. For example, a workspace can be divided into free and contact regions and use different sampling techniques or local planners tailored for the respective region~\citep{koval_pre-_2016, guan_efficient_2018}. Another approach is to leverage the environment's geometry to discretize the state space into regions, then sequence a subset of regions, and search for a motion plan through the selected regions~\cite {erdmann_using_1986, lozano-perez_automatic_1984, goldberg_orienting_1993, bhatia_sampling-based_2010}. A workspace decomposition can be learned~\citep{chamzas_learning_2021} or approximated~\citep{rickert_balancing_2014,rajendran_context-dependent_2019, liu_motion_2020} and used to bias exploration toward task-relevant regions of the workspace~\citep{rickert_balancing_2008}. 

We agree with ~\citet{rickert_balancing_2008} that exploration and exploitation should be balanced to combine priors with newly gained information. While their Exploring Exploiting Tree (EET) planner used free space information only and was unaware of uncertainty, we integrate EC-based free- and contact-region information into a Contact-EET (CEET) planner to guided exploration of \Cs to task-relevant free space and EC exploitation to reducing state uncertainty when needed.

\subsection{Motion Planning under Uncertainty}
We propose using ECE as manipulation funnels and for contact sensing, to reduce state uncertainty during planning. So, we present uncertainty handling planning approaches. 

%----------------------------------------------------------------------
Traditional sampling-based planners generate a collision-free path from a geometric model of the environment and robot kinematics by assuming that the world model and execution are perfect. \citep{elbanhawi_sampling-based_2014} reviewed the classical sampling-based motion planning methods, which methods can efficiently search the high-dimensional configuration space. Such planners are mostly uncertainty-unaware but some replan from scratch or discard portions of the explored space where execution failed. We share the view of ~\citet{elbanhawi_sampling-based_2014} that uncertainty-aware planning is still relevant in robotic research.

%---------------------------------------------------------------------
Conformant planning considers state uncertainty. Such a planner generates a fixed sequence of robust actions where actions are guaranteed to lead to a goal, even under uncertainty. Information that is accessible during execution is integrated into the plan to transition between actions. The transitioning events can be proprioception, the duration of motion, or any other sensor events. A classical way is to compute all regions from which compliant actions lead to a goal, so-called pre-images~\citep{lozano-perez_automatic_1984}, and then chain them to a sequence of actions. This approach can bring objects with unknown positions into the desired state without any sensing~\citep{erdmann_exploration_1988, goldberg_orienting_1993}. However, pre-images are impractical for high-dimensional \Cs computationally. In sampling-based approaches, information types ranging from artificial information regions to rangefinders and camera images were used~\citep{bry_rapidly-exploring_2011, platt_jr_belief_2010, van_den_berg_lqg-mp:_2011}, respectively, to estimate the relative position of a robot from obstacles and transition between actions. The combined exploration of free \Cs and in contact~
\citep{phillips-grafflin_planning_2020, sieverling_interleaving_2017, guan_efficient_2018, wirnshofer_robust_2018} enabled planners to search for interconnected manipulation funnels. 

With contingent planning~\citep{pryor_planning_1996}, a planner uses sensing to generate reactive behavior by implementing a decision tree or graph that branches based on observations. Contingent planners can handle more uncertainty than conformant planners since these can combine the benefits of manipulation funnels and contact sensing. Such planners were applied in a broad class of problems, for instance, part orientation for arbitrary shapes and realistic friction models~\citep{amagai_implementation_2001, zhou_probabilistic_2017} using visual or contact observation. One way to add contingencies to a motion plan is to reverse and retry motions that do not lead to the desired outcome~\citep{phillips-grafflin_planning_2020}. 

%----------------------------------------------------------------------
%\subsection{Optimal Planning Under Uncertainty}
%----------------------------------------------------------------------
POMDP solvers are a generic approach to compute globally optimal contingency plans. The solution is a contingent plan that maps a belief space to optimal actions. 
Point-based POMDP solvers can approximate optimal solutions~\citep{kurniawati_sarsop_2008} for low-dimensional, discrete state-, action-, and observation-spaces. For continuous state space problems (e.g., manipulation and grasping), some methods discretized \Cs into contact manifolds~\citep{hsiao_grasping_2007, koval_pre-_2016, koval_configuration_2020} and use contact-sensing as feedback. There exist solvers for continuous state space~\citep{bai_monte_2010} based on sampling and Monte Carlo simulation, for continuous action space~\citep{seiler_online_2015} relying on numerical optimization, and continuous state-action space by combining sampling and optimization-based techniques~\citep{vien_pomdp_2015}. However, these approaches can not easily be applied to high-dimensional configuration space or with increased contact manifolds.

We leverage manipulation funnels and contact events for both conformant and contingent planning because these reduce uncertainty. Contact events will be integrated increasingly into motion planning to handle increased amounts of uncertainty. However, we omit solution optimality from planning because efficient optimization methods exits~\citep{toussaint_dual_2014, posa_direct_2014, posa_optimization_2016, toussaint_sequence-constraints_2022} that can optimize our planner's solutions with contact-exploiting motions as well.

\subsection{Types of Environmental Constraints for Motion Generation}

We consider ECs to be a recurring structure or property of the environment associated with robot actions, so we categorize related work based on the environment's structural regularities used to simplify motion generation. 

%EC types: \\
%\textbf{geometrical ECs} and their use (ctr and planning)

First and most popular, \emph{geometrical ECs} are contact regions with consistent geometrical properties such as surfaces, edges, or corners offering manipulation funnels leveraged with deliberate motion on contact manifolds. Such manipulation funnels reduce uncertainty to a lower-dimensional manifold. Another benefit is that the geometrical properties of an environment define the perceivable direction of interaction forces. Geometrical regularities, like static surfaces and concave/convex edges, were used to stabilize an object for robotic grasping~\citep{deimel_exploitation_2016}. Moreover, a surface can help to pivot or reorient an object~\citep{odhner_open-loop_2013, chavan-dafle_prehensile_2015}, and a concave edge can expose a portion of the object for pinch grasping~\citep{kappler_representation_2010, hang_pre-grasp_2019}. \citep{mandery_analyzing_2015} also observed such ECE usage in human's whole-body posture for locomotion and manipulation. For motion planning, geometrical EC can be exploited similarly as visibility maps, for example using pre-images~\citep{lozano-perez_automatic_1984, erdmann_using_1986} or ECE- affordance graph~\citep{eppner_planning_2015}. Such ECs can be explored with sampling-based RRT-like planners~\citep{phillips-grafflin_planning_2020, sieverling_interleaving_2017, guan_efficient_2018, wirnshofer_robust_2018} or used as a descritized observation map for POMDP solvers~\citep{hsiao_grasping_2007, koval_pre-_2016, koval_configuration_2020}.

%\textbf{gravity as EC} and their use (ctr and planning)
Secondly, we can consider dynamic object properties as another type of EC, \emph{dynamic EC}, because dynamic property also constrains an object's motion. For example, an object's inertia in combination with gravity was used to reorient an object~\citep{ mason_progress_1999, dafle_extrinsic_2014, woodruff_planning_2017}.  In the given examples, the effect of inertia and gravity complemented the forces provided by a robot, and so the robot's dexterity was increased.

%\textbf{deformable ECs} and their use (ctr)
A further contact-based EC type is \emph{compliant ECs}. \citet{bhatt_surprisingly_2021} showed that compliant parts of a robot hand provide similar constraints as geometrical ECs for in-hand manipulation. They used some fingers of a soft hand to move an object on the palm. Other fingers provided a vertical constraint to restrict the object's motion and reduce uncertainty similar to geometrical ECs. However, compliant ECs provide force daping during interaction making an object's motion smoother as opposed to geometrical ECs where contact changes discretely. Nonetheless, even continuously changing forces can be thresholded to obtain contact events for further uncertainty reduction with sensing.

%non-contact-based ECs:\\
%\textbf{visual ECs} and their use (ctr)
Finally, ECs can be non-contact-based when features of the environment are used to generate virtual forces restricting the motion of a robot or an object and reducing uncertainty. For example, visual or acoustic ECs are features extracted from visual or acoustic measurements that can be used for visual servoing~\citep{hill1979real} or acoustic servoing~\citep{schoenwald_improved_1986}, respectively.

Next, we define motion generation problems with ECE in the focus, then integrate geometrical ECE benefits to \Cs motion planning, finally, we show the existence of a new EC manifesting in movable object in a homogeneous pile providing the same benefits when leveraged for grasping as geometrical ECs.

\section{Planning Problem Definitions}
\label{sec: problem definition}
%======================================================================

Before we dive into ECE implementation details, we first explain our proposed planning problem categorization and then provide formal problem definitions. We divide the motion planning problem into two sub-problems: 1) sequencing ECEs between a start and goal regions. It is enough to solve this sub-problem if the goal is reached by executing local policies associated with selected ECEs. Otherwise, we need to 2) plan a motion trajectory through the selected sequence of ECEs between the given start and goal, which is then followed during execution.

When searching for a motion plan through ECEs, we further differentiate between problem categories along two spectra. One spectrum is the \Cs complexity that increases as the dimensionality and geometrical complexity of the environment increases. The other spectrum is motion uncertainty. When motion uncertainty is low, error accumulates slowly, and local policies can reach their target reliably; however, high motion uncertainty results in probabilistic outcomes requiring different uses of ECE. Depending on problem characteristics, different ECE features are better suited for efficient planning. 

\subsection{ECE Sequencing}
\label{sec: prob def specilized ECE}
%------------------------------------------------------------------------------

We define the ECE sequencing problem as a graph search problem, where nodes are ECEs $\Upsilon$ and edges are directional transition operators $\mapsto$ between two ECEs. First, we build an ECE-graph for an instance of an environment and task, then search for a path in the graph, i.e., a sequence of ECEs that bring a robot (or manipulandum) from a set of initial states to a desired goal region. 

We define an ECE $\Upsilon$ to be a contiguous subset of all configurations of a body (a manipulandum or a sensorized part of a robot) on an EC and the exerted force onto the body by the environment:
$$\Upsilon \subset \mathcal{M}_{F} \times \mathcal{F},$$
where  $\mathcal{F}$ is the 6D (or 3D for planer objects) wrench space of the body that interacts with the EC and $\mathcal{M}_{F}$ is the lower-dimensional contact manifold in \Cs specified for an EC with a motion constraint function~$F$:
$$
\mathcal{M}_{F} =\left\lbrace \bm{q} \in \mathcal{C} | F(\bm{q},\dot{\bm{q}}) = 0 \right\rbrace ,
$$
where $\bm{q}$ and $\dot{\bm{q}}$ are a robot's (or manipulandum's) configuration and the first-order derivative.

We define the condition of existence of an edge $\mapsto$ between two nodes as a nonempty intersection of the two contact manifolds and a change in the contact force:
$$\Upsilon  \mapsto \Upsilon' \overset{\text{iff}}{\Rightarrow} \left\{\begin{matrix}
 \mathcal{M}_{F} \cap \mathcal{M}'_{F} &  \neq  & \emptyset \\ 
\mathcal{F} - \mathcal{F}' &  \neq & 0_n\\
\end{matrix}\right. , $$
where $0_n$ is an $n$ dimensional vector of zeros, $n$ is 6 for 3D bodies or 3 for 2D planer objects.

The planning problem is: given a start and goal state, a known environment, a task, ECs, and associated local policies, build the ECE-graph and find a path $\Pi = \{\Upsilon_{start},\:..., \Upsilon_{goal}\} $ connecting the start with the desired goal. Note that we assume to know the existing ECs or are able to detect them visually for a given environment and task. 

\subsection{Contact-Based Motion Planning}
\label{sec: prob def specilized ECE}
%------------------------------------------------------------------------------

We cast the motion planning problem as a belief space planning problem~\citep{van_den_berg_lqg-mp:_2011} to account for motion uncertainty. However, the possibility for the robot to establish contact with the environment makes a belief non-Gaussian or even multi-modal. 

We plan in an $n$-dimensional configuration space $\mathcal{C} \in \mathbb{R}^n$ and define $\mathcal{C}_{\text{valid}}$ the \emph{valid} configuration where a robot is within its joint limits and does not collide with the environment with its parts that do not sense contact. We differentiate between free and contact spaces to choose an appropriate local policy. Therefore, we decomposed $\mathcal{C}_{\text{valid}}$ into $\mathcal{C}_\text{free}$ free space and $\partial \mathcal{C}_\text{free}$ configurations in contact at the boundary of free space. We define a task-relevant subspace $\mathcal{C}_{\text{task}} \subset \mathcal{C}_\text{valid}$ where $\forall q \in \mathcal{C}_{\text{task}}$ is also in the space covered by $\Pi$, a solution of ECE sequencing.  

All motions have an uncertain outcome, and the robot can not fully perceive its configuration but must estimate it from noisy sensor measurements. So, we use a motion model with independent joint noise $\delta \hat{\mathbf{q}} = \delta \mathbf{q}+\mathcal{N}(0,\sqrt{|\delta \mathbf{q}|}\bm{\sigma}_\text{motion}).$ Moreover, the initial configuration is not known accurately, and we model the initial state uncertainty with a Gaussian distribution $\mathcal{N}(\mathbf{q}_0, \bm{\sigma}_\text{init})$ around a configuration $\vec{q}_0$ and variance $\bm{\sigma}_\text{init}$.

We need to constrain uncertainty accumulation to reach a goal region with high probability. Thus, we maintain a history of past motions and their effect on state uncertainty. Since we omit explicitly representing funnels, we can not account for motion noise in the shape of funnels like with pre-images~\citet{lozano-perez_automatic_1984}. Therefore, instead of planning in \Cscoma we plan in belief space $\mathcal{B}$, where each belief $b \in \mathcal{B}$ is a probability distribution over configurations with a mean configuration $\bm{\mu}_b$, uncertainty covariance matrix $\bm{\Sigma}_b$, and a fully observable contact state $o \in \mathcal{O}$. We define different contact sensor models to observe the active contact(s) at a given configuration $\bm{q}$:

1) \emph{Tactile sensor model} assumes binary contact sensing on different parts of the robot: $\mathcal{O}_{\text{tactile}} (\vec{q})=\{o_1, \ldots, o_k\}$, where each contact observation $o_i = \text{sensor}_i$ is a sensor patch indicating contact in the given configuration.

2) \emph{Force sensor model} assumes the robot can detect the contact normal: $\mathcal{O}_{\text{force}} ({\vec{q}})=\{o_1, \ldots, o_k\}$. Each observation $o_i = (\text{sensor}_i, \normal_i)$ is a pair of sensor patch and surface normal $\normal_i$. 

3) \emph{Oracle sensor model} assumes to detect the contacting surfaces of the robot's sensorized part and the environment, and the contact normal: $\mathcal{O}_{\text{force}} ({\vec{q}})=\{o_1, \ldots, o_k\}$, where $o_i =(\text{surface}_\text{robot},\:\text{surface}_\text{EC},\:\normal)$.  

The contact state is the relevant environmental feature of the EC used to detect the entrance of a funnel and instantiate the local policy. For example, the contact normal indicates which surface is reached in the local neighborhood of the robot, and it can be used to generate a sliding motion along the respective surface.

Next, we define $\mathcal{B}_\text{valid}$ as the space of all valid beliefs where $\forall b \in \mathcal{B}_\text{valid}$ lies mostly in the valid configuration space to respect additional constraints as defined for $\mathcal{C}_{\text{valid}}$ during planning:
\begin{equation}
\int_{\mathbf{q}\in\mathcal{C}_{\text{valid}}}b(\mathbf{q})d\mathbf{q}>1-\epsilon.
\label{eq: valid cs}
\end{equation}

Similarly to $\mathcal{B}_\text{valid}$, we define $\mathcal{B}_\text{task}$ as the space of all task beliefs which can be expressed with Equation~\eqref{eq: valid cs} by integrating $\vec{q}$ over $\mathcal{C}_{\text{task}}$. 

%\textbf{The planning problem is the following}: given a start and goal beliefs $b_0, b_g \in \mathcal{B}_\text{valid}$, search for a motion plan $\pi$ through $\mathcal{C}_\text{free} \cup \partial\mathcal{C}_\text{free}$ that brings a robot from $b_0$ inside $b_g$. 
%
%
%\textbf{The planning problem is:} given a start and goal beliefs $b_0, b_g \in \mathcal{B}_\text{task}$, search for a policy $\pi: {\mathcal{B}}_\text{task} \rightarrow \mathcal{U}$ that brings the robot from the start to the goal belief state. 

We assume to have perfect knowledge about the environment, the robot's kinematic model and its contact-state is fully observable. While we assume that an accurate geometrical model of the environment is given and use it to approximate a task-relevant subspace, a visual-perception-based approach can also approximate $\mathcal{C}_{\text{task}}$ as shown by~\citet{eppner_exploitation_2015} when they used ECs for single object grasping. 

The planning problem is now: given a start and goal belief $b_0, b_g \in \mathcal{B}_\text{valid}$, find a policy $\pi: \mathcal{B}_\text{task}\rightarrow \mathcal{U}$ that brings the robot to the goal belief state with high probability. Note that we only care about finding feasible policies and do not consider optimality, unlike the POMDP solver. However, the policy computed with the following planner(s) can be used as an input for contact-based trajectory optimization algorithms~\citep{toussaint_dual_2014, posa_direct_2014, posa_optimization_2016, toussaint_sequence-constraints_2022}.

In the following three sections (\ref{sec: cerrt}, \ref{sec: concerrt}, and \ref{sec: ceet}), we explain what benefits of ECE should be integrated into a planning algorithm along the spectra of \Cs complexity and motion uncertainty. Finally, Section~\ref{sec: gece} provides a case study where we only solve ECE sequencing to robustly grasp from piles of nearly identical objects.

%======================================================================
\section{Planning in Low Complexity $\bm{\mathcal{C}}$-Spaces and Under Low Motion Uncertainty}
\label{sec: cerrt}
%======================================================================

In this section, we introduce a contact-exploiting variant of the RRT planner (CERRT) \citet{sieverling_interleaving_2017}. Just like RRT, CERTT uses diffusion to explore the \Cs until a solution path is found. However, each of the expansion steps consists of contact-exploiting motions, akin to the manipulation funnels defined by Mason \citep{mason_mechanics_1985}. The continuous expansion of the CERRT planner's tree effectively tiles the C-space with sequences of funnels. The result is a planner that is efficient on motion planning problems with moderate uncertainty and in relatively low $\mathcal{C}$-spaces complexity. 

With rapid exploration of random manipulation funnels and their connectivity, we simplify planning because we tile only a subspace with funnels between a given start and a given goal. Since we explore manipulation funnels, we can limit state uncertainty accumulation over a trajectory. However, a tree structure allows only deterministic funnel connectivity, and so, it can only handle moderate uncertainty accumulation. On the other hand, just like RRT, random funnel exploration becomes inefficient in high-dimensional spaces, and it is hindered by complex environmental features, such as narrow passages.    

% This section addresses \Cs planning, where the state space has low complexity and motion uncertainty is also low. Even though we consider low complexity $\mathcal{C}$-spaces, we devise a planning algorithm that scales to higher \Cs dimensions, like 7 DoF of modern robot arms. Since we consider low motion uncertainty, i.e., slow accumulation of inaccuracies, uncertainty affects mainly goal region reachability without leading actions to undesired collisions.

% We combine sampling-based planning with ECE to maintain computational practicality in high dimensional $\mathcal{C}$-spaces and to ensure goal reachability. For the latter, we leverage ECE as manipulation funnels because these funnels implicitly collapse state uncertainty along one dimension when moving on a contact manifold. 

% We simplify planning compared to tiling the entire \Csdot Moreover, we omit explicitly computing a funnel tile in \Cs to further simplify planning. For this, we rely on local planners during planning to explore manipulation funnels and their connectivity; during execution, we use complaint control and contact sensing to detect, use, and traverse funnels. 

\subsection{Efficient Funnel Sequencing in Configuration Space}
%------------------------------------------------------------------------------

\begin{figure}[tb]
    \centering
     \def\svgwidth{1.0\linewidth}
        %% Creator: Inkscape inkscape 0.92.3, www.inkscape.org
%% PDF/EPS/PS + LaTeX output extension by Johan Engelen, 2010
%% Accompanies image file 'uncertainty_reduction_funnels.pdf' (pdf, eps, ps)
%%
%% To include the image in your LaTeX document, write
%%   \input{<filename>.pdf_tex}
%%  instead of
%%   \includegraphics{<filename>.pdf}
%% To scale the image, write
%%   \def\svgwidth{<desired width>}
%%   \input{<filename>.pdf_tex}
%%  instead of
%%   \includegraphics[width=<desired width>]{<filename>.pdf}
%%
%% Images with a different path to the parent latex file can
%% be accessed with the `import' package (which may need to be
%% installed) using
%%   \usepackage{import}
%% in the preamble, and then including the image with
%%   \import{<path to file>}{<filename>.pdf_tex}
%% Alternatively, one can specify
%%   \graphicspath{{<path to file>/}}
%% 
%% For more information, please see info/svg-inkscape on CTAN:
%%   http://tug.ctan.org/tex-archive/info/svg-inkscape
%%
\begingroup%
  \makeatletter%
  \providecommand\color[2][]{%
    \errmessage{(Inkscape) Color is used for the text in Inkscape, but the package 'color.sty' is not loaded}%
    \renewcommand\color[2][]{}%
  }%
  \providecommand\transparent[1]{%
    \errmessage{(Inkscape) Transparency is used (non-zero) for the text in Inkscape, but the package 'transparent.sty' is not loaded}%
    \renewcommand\transparent[1]{}%
  }%
  \providecommand\rotatebox[2]{#2}%
  \newcommand*\fsize{\dimexpr\f@size pt\relax}%
  \newcommand*\lineheight[1]{\fontsize{\fsize}{#1\fsize}\selectfont}%
  \ifx\svgwidth\undefined%
    \setlength{\unitlength}{895.44918823bp}%
    \ifx\svgscale\undefined%
      \relax%
    \else%
      \setlength{\unitlength}{\unitlength * \real{\svgscale}}%
    \fi%
  \else%
    \setlength{\unitlength}{\svgwidth}%
  \fi%
  \global\let\svgwidth\undefined%
  \global\let\svgscale\undefined%
  \makeatother%
  \begin{picture}(1,0.36800613)%
    \lineheight{1}%
    \setlength\tabcolsep{0pt}%
    \put(0,0){\includegraphics[width=\unitlength,page=1]{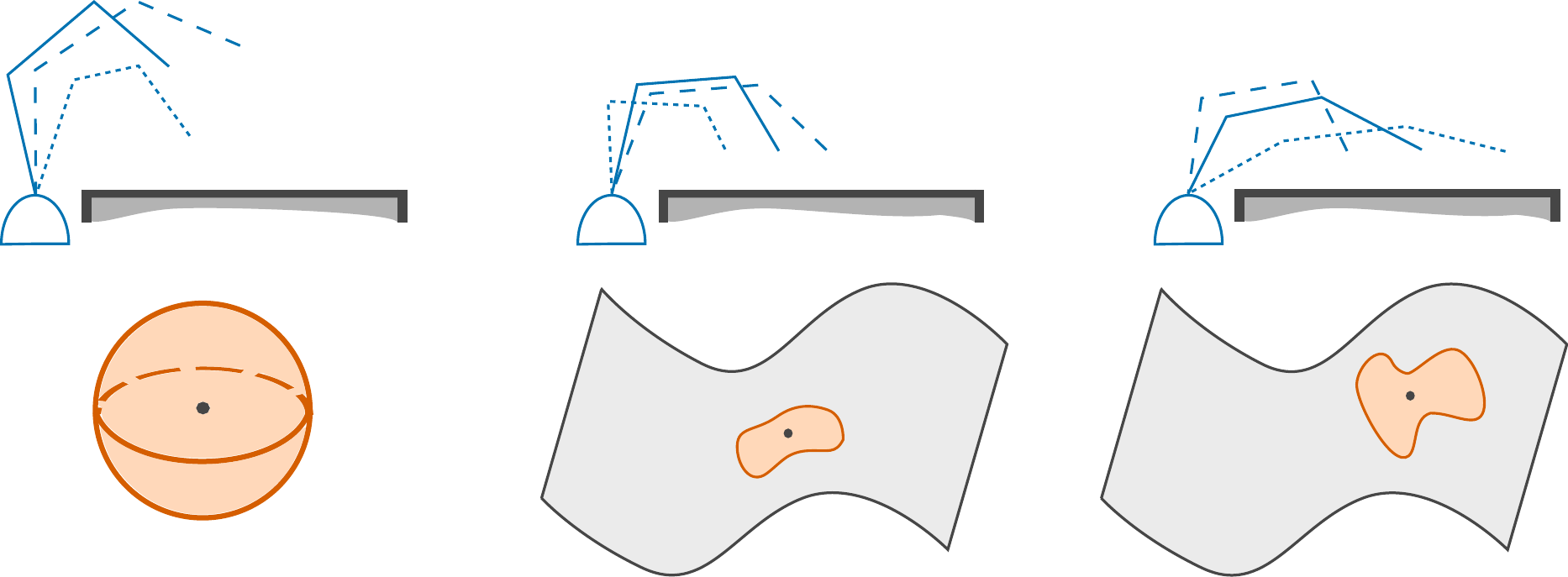}}%
    \put(0.17950113,0.28812753){\makebox(0,0)[lt]{\lineheight{1.25}\smash{\begin{tabular}[t]{l}$\xrightarrow[]{\text{guarded move}}$\end{tabular}}}}%
    \put(0.1150641,0.14159559){\color[rgb]{0,0,0}\makebox(0,0)[lt]{\lineheight{0}\smash{\begin{tabular}[t]{l}$b$\end{tabular}}}}%
    \put(0.36193532,0.04560321){\makebox(0,0)[lt]{\lineheight{1.25}\smash{\begin{tabular}[t]{l}$\mathcal{M}_F$\end{tabular}}}}%
    \put(0.52990838,0.11697749){\color[rgb]{0,0,0}\makebox(0,0)[lt]{\lineheight{0}\smash{\begin{tabular}[t]{l}$b'$\end{tabular}}}}%
    \put(0.71903445,0.04550669){\makebox(0,0)[lt]{\lineheight{1.25}\smash{\begin{tabular}[t]{l}$\mathcal{M}_F$\end{tabular}}}}%
    \put(0.89340027,0.14590714){\color[rgb]{0,0,0}\makebox(0,0)[lt]{\lineheight{0}\smash{\begin{tabular}[t]{l}$b''$\end{tabular}}}}%
    \put(0.62621424,0.28812815){\makebox(0,0)[lt]{\lineheight{1.25}\smash{\begin{tabular}[t]{l}$\xrightarrow[]{\text{slide}}$\end{tabular}}}}%
    \put(0,0){\includegraphics[width=\unitlength,page=2]{uncertainty_reduction_funnels.pdf}}%
  \end{picture}%
\endgroup%

    \caption{Illustrations of uncertainty reduction with contact-exploiting motions: in the top row, three sketches show an RRR robot in three possible configurations with dashed, dotted, and continuous lines as links. From left to right, the robot executes contact-exploiting motions. Consequently, state uncertainty is reduced as illustrated at the bottom row: an initial uniform uncertainty on all three joins is illustrated with $b$ sphere, then the sphere is projected onto the contact manifold $\mathcal{M}_{F}$ collapsing uncertainty along one dimension. Finally, by moving in contact, uncertainty accumulation is constrained along the manifold's dimension.}
    \label{fig: manifold and ece}
\end{figure}

We sequence manipulation funnels to reduce state uncertainty and reach the goal region with high probability. To efficiently sequence funnels in a high-dimensional \Cscoma, we represent funnels with local policies, or in the motion planning context, also called local planners. Then, we sequence local planners between a given start and goal. By sequencing local policies between a start and goal region, we partially covered the state space with manipulation funnels, computing a region where the goal can be reached while uncertainty is reduced. By implicitly computing a partial cover, we greatly simplify planning because it is computationally intractable to compute exactly funnel regions and their intersections in high-dimensional \Cs and complex environments.

To sequence manipulation funnels, we need to define their respective local planners: To traverse a funnel without touching its wall, we use a free-space motion, for example, the \textit{connect} action~\citep{kuffner_rrt-connect_2000} that is a straight line joint space motion in free space. To deliberately contact a funnel's wall, we use a \textit{guarded move}~\citep{will_experimental_1975} that is a straight line joint space motion in free space ending when a sensorized part of the robot contacts the environment. Finally, we use a \textit{sliding} action to move in contact. These three local planners and their effect on state uncertainty are shown in Figure~\ref{fig:  manifold and ece}. Note that we are not limited to these three types of local planners, but these are the minimum to search for motions through a funnel and to transition between funnels; in later sections, we present other actions as well. %In our implementation below, we use these three action types where sliding is implemented for flat surfaces.   

We can efficiently explore valid motions through funnels and transition between them by random sampling configuration and local planners using a multi- or single-query motion planner, depending on the amount and source of uncertainty. Single-query planning is advised for the current problem definition, where actuation and proprioception are inaccurate because state uncertainty depends on past actions. After all, single-query planners encode the history of actions. So next, we explain the implementation details for such a planner.  In contrast, multi-query planners are not suited to maintain action history, and so such planners are applicable when state uncertainty is independent of past action.

\subsection{Implementation of Sampling-Based Manipulation Funnel Sequencing}
\label{sec: alg cerrt}
%------------------------------------------------------------------------------

We present the Contact Exploiting RRT (CERRT) motion planner~\citep{sieverling_interleaving_2017} that sequences manipulation funnels between a given start and goal belief states by randomly exploring motions in free space and in contact.

%that instantiates manipulation funnel exploration to maintain \Cs connectivity between a given start and goal independent of \Cs conditionality. We focusing on manipulation funnel sequencing and for further implementation details about simulating contact-exploiting local planners, we refer to the respective publication~\citep{sieverling_interleaving_2017}. 

The CERRT planner (Algorithm~\ref{alg:cerrt}) is in the family of Partical-RRT planners~\citep{melchior_particle_2007} because it represents a belief with a set of particles and the associated contact state $b = (Q, O)$, where $Q = \{\vec{q}_0,...,\vec{q}_N\}$ and $O = \{{o}_0,...,{o}_m\}$. A particle-based representation is desired because an initial Gaussian belief becomes non-Gaussian or multi-model when projected on contact manifolds.

Like RRT planners, CERRT samples a random configuration with goal bias then choose the nearest neighbor of the sample, simulates an action toward the sample, and tries to connect a new valid node to the goal. However, there are three main differences due to considering state uncertainty and using funnels to reduce it. First, the nearest neighbor method combines Euclidean joint distances and a measure of belief uncertainty to minimize state uncertainty propagation and increase \Cs exploration. In the original implementation, it is formulated as follows:
\begin{align*}
 b_\text{near} =  \underset{b \in T}{argmin}&  \left[ \gamma \cdot d_{\bm{\Sigma}}(b) +(1-\gamma)\cdot d_{\bm{q}}(b) \right],
\end{align*}
where $d_\vec{q} = \| \vec{q}_\text{rand} - \bm{\mu}_b \|_2 $ is the Euclidean joint distance between a random sample $\vec{q}_\text{rand}$ and the mean configuration of a belief, $d_{\bm{\Sigma}} = \sqrt{\text{Tr}(\bm{\Sigma})}$ is uncertainty measure of a belief. The two terms are weighted with the $\gamma$ parameter. $\gamma$ adjusts the admissible state uncertainty propagation during exploration because large values of $\gamma$ penalize more uncertainty over joint distances, so nodes with low uncertainty are extended. In contrast, small values of $\gamma$ weigh more joint space distances than uncertainty, so even high uncertainty nodes are expanded.   

Secondly, the planner chooses between free-space and contact-exploiting actions biased by $\gamma$. The biasing heuristic is tailored to the previous distance metric: When $\gamma = 1$, the metric penalizes state uncertainty, and only guarded motion or sliding actions are chosen, i.e., motion in contact is explored, reducing state uncertainty. On the other hand, when $\gamma = 0$, the distance metric ignores uncertainty, and only the connect action is chosen, i.e., $\Cf$ is explored without using any contact exploitation. 

Thirdly, since CERRT can choose between free-space and contact-exploiting motions, it requires appropriate local planners. All local planners forward propagate a belief toward a random sample by simulating the chosen local planner for each particle in the nearest belief. The connect and guarded move actions are identical to the one in the RRT-Connect~\citep{kuffner_rrt-connect_2000}, and sliding uses the task projection technique~\citep{stilman_task_2007}.

\begin{algorithm}[tbp]
    \caption{CERRT planner}
	\label{alg:cerrt}
	\begin{algorithmic}[1]
		\small
		\REQUIRE ${b}_0, {b}_{g}, \epsilon_{\mathrm{goal}}, \gamma$
		\ENSURE $T = (V, E)$
%					\STATE $\vec{\mathcal{Q}}_{\mathrm{init}} \leftarrow \operatorname{SAMPLE}(\mathcal{N}(\vec{q}_{\mathrm{start}}, \vec{\sigma}_{\mathrm{start}}), N)$\COMMENT{sample start particles}\label{alg:cerrt-start}
		\STATE $V \leftarrow b_0$\COMMENT{Init tree vertexes with start state}
		\STATE $E \leftarrow \emptyset$\COMMENT{Init tree edges with empty set}
		\WHILE[search until goal reached]{true}
					\STATE $\vec{q}_{\text{rand}} \leftarrow \operatorname{RANDOM\_CONFIG}()$ \COMMENT{Sample random $\vec{q}$ with goal bias}
					\STATE ${b}_{\text{near}} \leftarrow \operatorname{NEAREST\_NEIGHBOUR}( \vec{q}_{\text{rand}},T,\gamma)$\COMMENT{Find closest belief in tree}
					\STATE $u \leftarrow \operatorname{SELECT\_ACTION}(\vec{q}_{\text{rand}}, {b}_{\text{near}},\gamma)$\COMMENT{Choose between free space or contact-exploiting motion}

			\STATE ${b}_{\mathrm{new}} \leftarrow \operatorname{SIMULATE}({b}_{\text{near}}, u, \vec{q}_{\text{rand}})$\COMMENT{Simulate action}

			\IF[Check if belief is valid]{${b}_{\text{new}} \in \mathcal{B}_\text{valid}$}
				\STATE $V \leftarrow V \cup \{{b}_{\text{new}}\}$
				\STATE $E \leftarrow E \cup \{({b}_{\text{near}}, {b}_{\text{new}})\}$
				\STATE $b_\text{connect} \leftarrow \operatorname{SIMULATE}(b_{\text{new}},\text{connect},\bm{\mu}_{b_\text{goal}})$\COMMENT{Try connecting to goal}
				\IF[Check if resulting belief is in goal region]{$\lVert b_\text{connect} - b_\text{goal} \rVert < \epsilon_{\text`{goal}}$}
					\STATE $V \leftarrow V \cup \{{b}_{\text{connect}}\}$
					\STATE $E \leftarrow E \cup \{({b}_{\text{new}}, {b}_{\text{connect}})\}$
					\RETURN $G$
				\ENDIF

			\ENDIF
\ENDWHILE
	\end{algorithmic}	
\end{algorithm}

In summary, probabilistic goal reachability due to motion uncertainty can be mitigated by leveraging contact-exploiting motion that collapses state uncertainty on manipulation funnel manifolds. Consequently, the presented approach applies to planning problems where the goal can be expressed relative to the environment. Moreover, computational practicality can be maintained even in high-dimensional configuration spaces using a sampling-based approach for manipulation funnel exploration that omits explicitly computing a funnel coverage.

%======================================================================
\section{Planning under High Motion Uncertainty}
\label{sec: concerrt}
%======================================================================

Here, we introduce a contact-contingency planner, a variant of the contact-exploiting RRT planner, the Contingent~CERRT (ConCERRT) planner~\citep{pall_contingent_2018}. Like CERRT, ConCERRT rapidly explores random manipulation funnels between a given start and goal regions. However, it generates contact contingencies: tree branches conditioned on distinct contact events that may occur under execution uncertainty and differentiate between reached manipulation funnels. Consequently, the planner is efficient on motion planning problems with increased uncertainty and relatively low \Cs complexity. 

With contact event-based contingency planning, we simplify motion planning under increased uncertainty. We handle high uncertainty using funnels to collapse state uncertainty along one dimension and using contact sensing to partition a funnel's exit into regions. These regions are entrances to other funnels. Even though each reachable partition requires computing a contingency plan, we simplify contingency planning with dynamic programming explained next. 
% by reusing solved contingencies for the unsolved ones.

% This section addresses \Cs planning under high motion uncertainty and low \Cs complexity. Since we consider high motion uncertainty, belief states become increasingly uncertain as a robot moves, leading to probabilistic motion outcomes. Thus, we focus on uncertainty reduction and scalability to high-dimensional \Cs planning.

% Similarly to the previous section, we combine sampling-based planning to maintain computational practicality with manipulation funnels to reduce uncertainty. As opposed to the previous section where the planner computed a \textit{conformant plan} i.e., sequence of manipulations funnels and motions through them, now, we need to compute a \textit{contingency plan} that allows branches in the plan when it is probable to reach different funnels with a single action.  

% We handle high uncertainty using funnels by collapsing state uncertainty along one dimension and using contact sensing by partitioning a funnel's exit into regions. These regions are entrances of different funnels. Even though each reachable partition requires computing a contingency plan, we simplify contingency planning with dynamic programming by reusing solved contingencies for the unsolved ones.

\subsection{Efficient Contingent Funnel Sequencing in Configuration Space}
\label{sec: concerrt insight}
%------------------------------------------------------------------------------

\begin{figure}[tbp]
\centering
\def\svgwidth{1\linewidth}
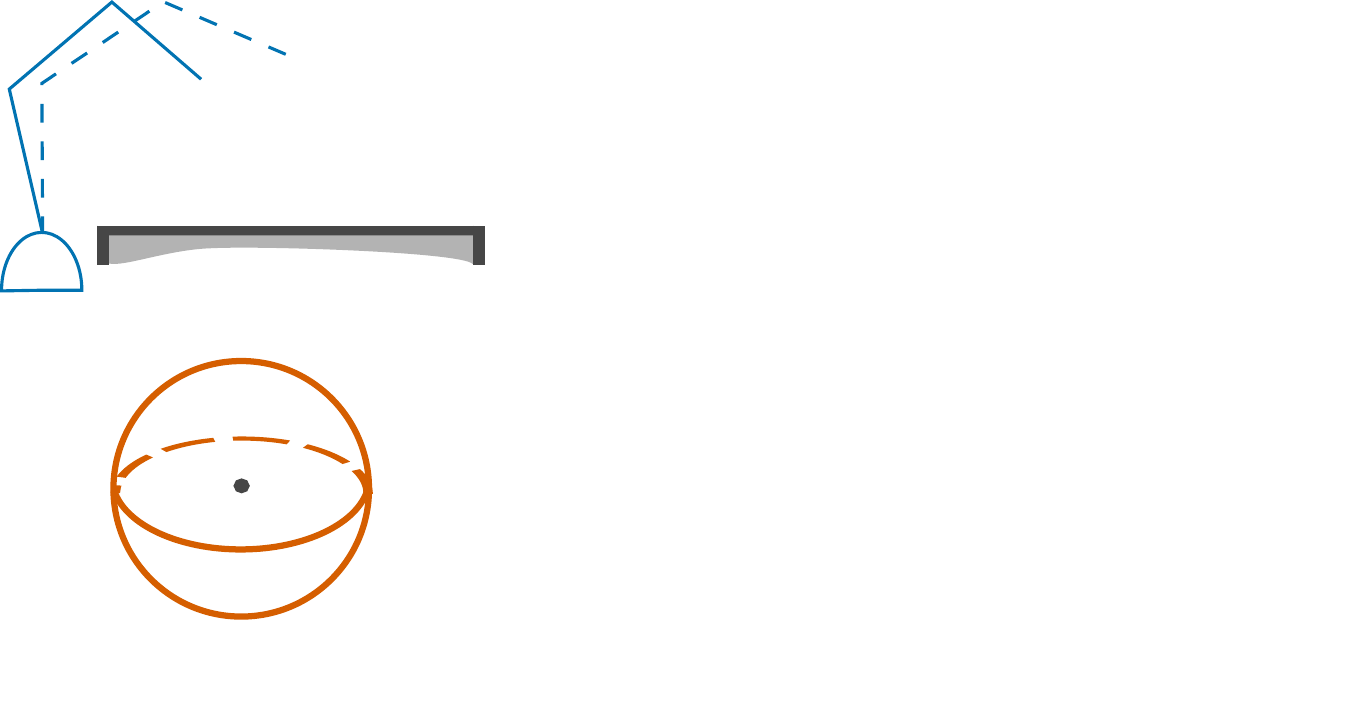
    \caption{Illustration of uncertainty reduction using contact sensing: an RRR robot's initial state uncertainty is visualized in the top row with dashed, dotted, and continuous lines and in the bottom row with orange volume. Due to initial uncertainty and inaccurate motion, the robot may touch the surface with one or both fingers. If the fingers can sense contact, the manifold projected to the manifold $\mathcal{M}_F$ can be partitioned into three regions: contact with left $b_l$, right $b_r$, or both $b_{l,r}$ fingers. Each partition alone has lower state uncertainty than combined.}
    \label{fig: partitioning example}
\end{figure}

For contingency planning, we anticipate possible motion outcomes that may occur during execution by anticipating contact events when different manipulation funnels are reachable with the same inaccurate motion. Since manipulation funnels are associated with a distinct geometrical feature of the environment, the contact event is measurably different when reached. If the robot can sense contact events, it can reduce its uncertainty by partitioning its belief state into smaller beliefs than the original, where each partition has a consistent contact state as illustrated in Figure~\ref{fig: partitioning example}. Since contact events are triggered when reaching a funnel, uncertainty is further reduced by collapsing to the funnel's wall. Such partitions can arise from tactile or force sensors as well. 

Using belief-space partitioning seemingly increases computational complexity because every new partition is a new belief state that must eventually be connected to the goal. However, this effort can be limited in practice.

We simplify contingency planning in two ways. First, we assume fully observable contact sensing to limit branching in the final policy. Still, each branching is a new planning problem that needs to be solved. Secondly, we simplify solving each new planning problem with dynamic programming~\citep{bellman1957dynamic} by reusing solved contingencies as goal regions.

\begin{figure}[tbp]
\centering
    \subfloat[first iteration\label{subfig-1: policy}]{%
	\def\svgwidth{.47\linewidth}
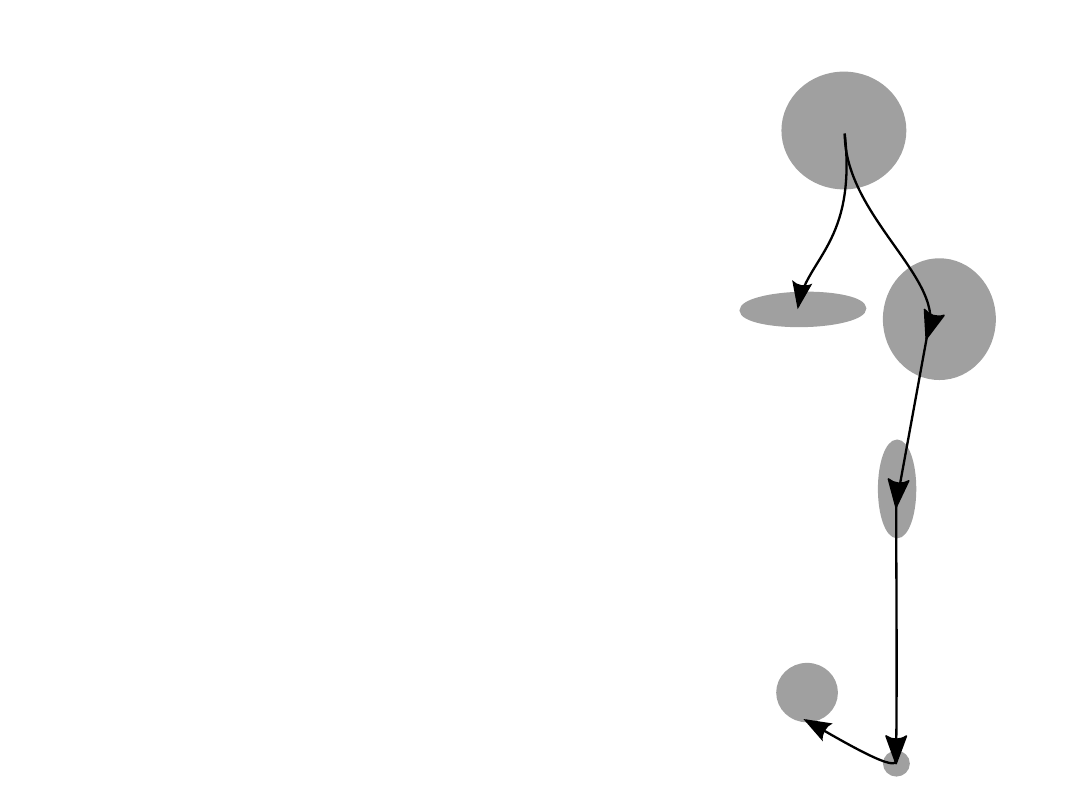
    }
    \subfloat[second iteration\label{subfig-1: policy}]{%
	\def\svgwidth{.47\linewidth}
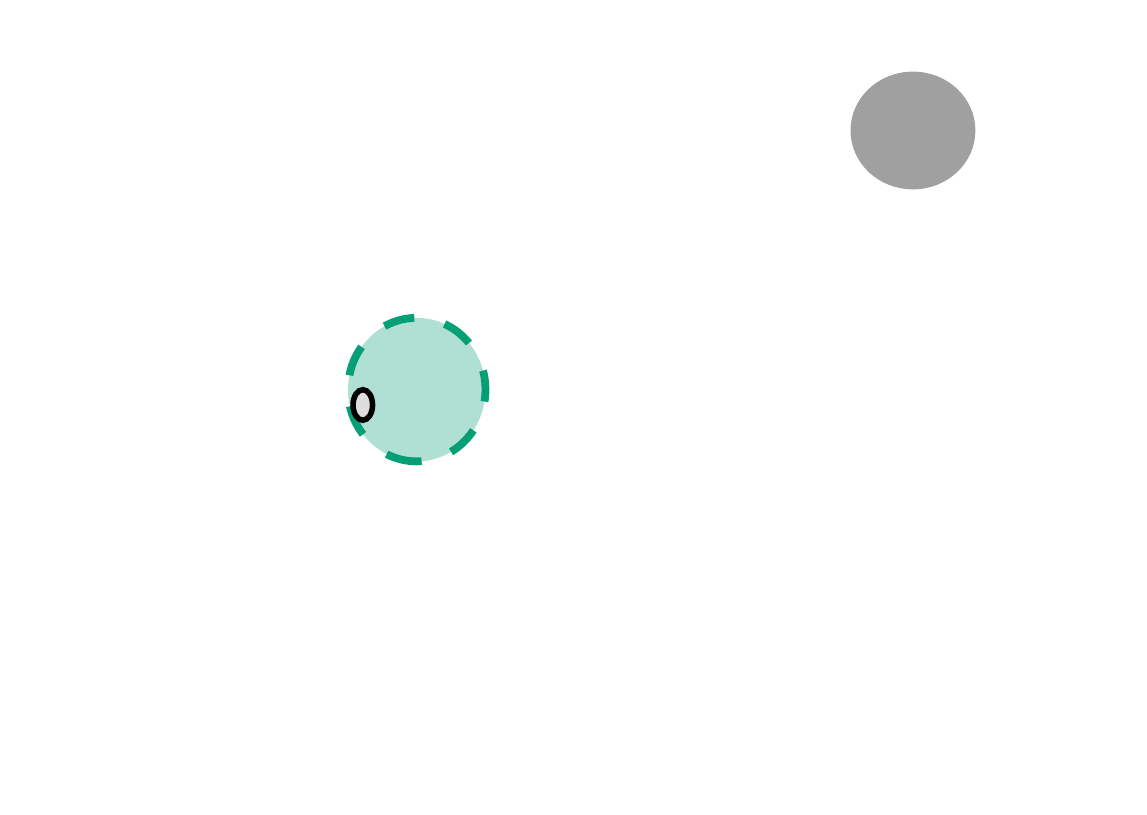
    }

    \caption{Reusing solved contingencies as goal region simplifies planning for unsolved contingencies illustrated with two iterations of the ConCERRT planner: during the \textit{a)~first~iteration}, the initial search tree $T_{b_0}$ connects $b_0$ and $b_g$ beliefs (left); the resulting policy $\pi$ consists of one path from start to goal and has one unconnected partition $b_1$ (right). In the \textit{b)~second~iteration}, $T_{b_1}$ tree is expanded from $b_1$ and connects to a belief in $\pi$ via $b_{g_2}$ with a sliding motion (left); this action is added to the final policy $\pi'$ (right).}
    \label{fig:policy_construction}
\end{figure}

To reuse solved contingencies as goal regions, the planner incrementally builds a policy $\pi$. The working principle of belief partitioning and incremental policy building is illustrated in Figure~\ref{fig:policy_construction}. We maintain two separate lists of belief states:
\begin{itemize} 
\item$\mathcal{B}_{\text{open}}$ contains all belief states that are yet to be connected to the goal. It is initialized with the initial belief $b_0$ and increases with new belief partitions in the policy. When $\mathcal{B}_{\text{open}}$ is empty, the policy provides 100\% success probability.
\item$\mathcal{B}_{\text{connected}}$ contains all beliefs that are already connected to the goal. Initially, it only contains the goal belief $b_g$. However, over time, $\mathcal{B}_{\text{connected}}$ is extended with new beliefs connected to a goal belief.
\end{itemize}

We can run a conformant planner for every state in $\mathcal{B}_{\text{open}}$ as a separate tree search, attempting to connect to any state in $\mathcal{B}_{\text{connected}}$. When a node from $\mathcal{B}_{\text{open}}$ connects with a node from $\mathcal{B}_{\text{connected}}$, we add the resulting action sequence to the policy, the visited nodes on that path to $\mathcal{B}_{\text{connected}}$, and nodes on unsolved branches on the path to $\mathcal{B}_{\text{open}}$. 

This parallel search using a forest of trees might seem like an overhead. However, the effort is limited because the algorithm shifts from exploration to exploitation~\citep{rickert_balancing_2014}. Initially, the algorithm must explore most of \Cs as $\mathcal{B}_{\text{connected}}$ contains only $b_g$. However, as $\mathcal{B}_{\text{connected}}$ increases with each solved contingency, the nodes in $\mathcal{B}_{\text{connected}}$ cover more and more of the \Cscoma creating opportunities for exploitation which decrease the planning complexity of later iterations.

\subsection{Implementation of Contingent Manipulation Funnel Sequencing}
\label{sec: alg concerrt}
%------------------------------------------------------------------------------

We present the Contingent Contact Exploiting RRT (ConCERRT) motion planner~\citep{pall_contingent_2018} that sequences manipulation funnels considering probabilistic transitions between funnels and computing contingencies for anticipated motion outcomes. The new planner uses the conformant CERRT planner~\citep{sieverling_interleaving_2017} from Section~\ref{sec: alg cerrt} to expand each belief in $\mathcal{B}_{\text{open}}$, but other conformant contact-exploiting belief-space planners can be used as we will show in Section~\ref{sec: ceet}. 

The ConCERRT planner, in Algorithm~\ref{alg:contingency planner}, initially samples a fixed number of particles from the initial belief $b_0$, similarly to CERRT, and then, it adds them as root to the initial search tree. In every iteration, ConCERRT cycles through all elements of $\mathcal{B}_\mathrm{open}$ and expands the respective tree. The expansion works similar to the CERRT planner. It samples a random configuration, finds the nearest neighbor in the current search tree, chooses an action, simulates the effects of that action, adds the resulting state to the tree, and tries to connect the new state to a goal. After each tree expansion, it updates the policy, the open beliefs, and the closed beliefs.

\begin{algorithm}[htb]
    \caption{ConCERRT planner}
	\label{alg:contingency planner}
	\begin{algorithmic}[1]
		\small
		\REQUIRE $\vec{b}_0, \vec{b}_{\mathrm{g}}, \epsilon_\text{goal}, \gamma$
		\ENSURE $\pi$
		\STATE $\mathcal{B}_\mathrm{open} \leftarrow \vec{b}_0$ \COMMENT{ Init set of open belief.}
		\STATE $\mathcal{B}_\mathrm{connected} \leftarrow \vec{b}_{\mathrm{g}}$ \COMMENT{Init set of closed beliefs.}
		\STATE $T_{\vec{b}_0}.V \leftarrow \vec{b}_0$ \COMMENT{Init first tree vertexes.}
		\STATE $T_{\vec{b}_0}.E \leftarrow \emptyset$ \COMMENT{Init first tree edges.}
		\STATE $\pi \leftarrow \emptyset$						\COMMENT{Init policy.}
		\WHILE[Search until all contingencies are solved.]{$P(\pi) < 1$}	
			\FORALL[Expand each unsolved contingency.]{$ \vec{b} \in \mathcal{B}_\mathrm{open} $}		
				\STATE $T_{\vec{b}} \leftarrow T_{\vec{b}}.\operatorname{EXPAND}(\mathcal{B}_\mathrm{connected})$	\COMMENT{Update tree.}
				\STATE $\pi \leftarrow \pi.\operatorname{UPDATE}(T_{\vec{b}})$ \COMMENT{Update policy with partial solutions.}
				\STATE $\mathcal{B}_{\mathrm{open}} \leftarrow \mathcal{B}_{\mathrm{open}}.\operatorname{UPDATE}( T_{\vec{b}})$ \COMMENT{Remove solved contingencies.}
				\STATE $\mathcal{B}_{\mathrm{connected}} \leftarrow \mathcal{B}_\mathrm{connected}.\operatorname{UPDATE}(T_{\vec{b}})$ \COMMENT{Add new contingencies.}
			\ENDFOR
		\ENDWHILE		
		\RETURN $\pi$
	\end{algorithmic}	
\end{algorithm}
%    \vspace{-1.5em}

Even though the expansion method is based on the CERRT planner, using similar steps that are inside the while loop in Algorithm ~\ref{alg:cerrt}, we change \textit{nearest neighbor selection} (line 5 in Alg.~\ref{alg:cerrt}), add a new \textit{belief partitioning} method after a validating belief propagation (after line 8 in Alg.~\ref{alg:cerrt}), and extend the \textit{goal connect} method (line 11 in Alg.~\ref{alg:cerrt}) as well. We explain each modification in detail below.

First, the \textit{nearest neighbor selection}, similar to CERRT, computes a norm that balances joint distance and uncertainty with $\gamma$ and uses the same spatial distance term $d_{\vec{q}}(b)$. However, the distance term over uncertainty includes the trace norm of a belief $b$ and the trace norm of all sibling beliefs $Sib(b)$, i.e., the partitions that could be reached from the same action and need to be solved if this branch connects to the goal region. So, ${b}_\text{near}$ in tree $T_i$ for a random sample $\vec{q}_\text{rand}$ is chose as follows:
\begin{align*}
 b_\text{near} =  \underset{b \in T_i }{argmin}&  \left[\gamma \cdot \left(d_{\bm{\Sigma}}(b)+\sum_{b' \in Sib(b)}d_{\bm{\Sigma}}(b')\right) \right. \\
 & ~\left. +(1-\gamma)\cdot d_{\bm{q}}(b)\right].
\end{align*}

Secondly, we partition a new belief $b_\text{new}$ after verifying that $b_\text{new}$ is inside the valid belief space. For partitioning, we apply a contact sensor model by computing for each particle the observable contact state $\mathcal{O}(\vec{q})=\{o_0, \ldots, o_k\}$. We then cluster the belief $b_\text{new}$ into $\{b^o_{0}, \ldots, b^o_{n}\}$, such that particles with the same measurement are in the same belief. The implementation is different for the two sensor models: For the \emph{tactile} sensor model, we cluster based on the sensor patches in contact. For the \emph{force} sensor model, we cluster two particles into different beliefs if the difference between their measured normals is larger than $15\degree$. We estimate the transition probabilities as $p(b^o_{i}|b_\text{near},u) \approx \frac{|\mathcal{Q}(b^o_{i})|}{N_\mathrm{particles}}$. Then, for each $b^o_{i}$, we sample the missing particles and execute the same steps as in Algorithm~\ref{alg:cerrt} from line 9 to 14, and then return the tree.

The last difference to CERRT is using a different goal connect method that connects each new belief $b^o_i$ to any belief in $\mathcal{B}_\mathrm{connected}$. To do so, we simulate a noisy connect action towards every $b_\mathrm{goal} \in \mathcal{B}_\mathrm{connected}$ resulting in a new distribution $b'$. We check if $b'$ lies within the goal belief by testing if ${d_M(\vec{q})<\epsilon_M = 2}$ for all $\vec{q}\in b'$, where $d_M(\vec{q})$ is the Mahalanobis distance between $\vec{q}$ and $b_ \mathrm{goal}$. If this test succeeds, ConCERRT UPDATEs the policy $\pi$ with all beliefs on the solution path, $\mathcal{B}_\mathrm{connected}$ with all new beliefs that were connected to the goal, and $\mathcal{B}_\mathrm{open}$ with all new partitions that are not yet connected to the goal. Compared to the previously published ConCERRT algorithm~\citep{pall_contingent_2018}, this implementation includes an additional belief partitioning during \textit{goal connect} to find partial solutions faster. We found that belief partitioning during goal connect was fundamental when using structural context to simplify planning detailed in Section~\ref{sec: ceet}. 

%\begin{algorithm}[htb]
%	\caption{$T$.EXPAND()}
%	\label{alg:expand}
%	\begin{algorithmic}[1]
%		\small
%		\REQUIRE $\mathcal{B}_{connected}$
%%		\ENSURE true, false	
%%		\STATE  solved $\leftarrow$ false		
%		\STATE $q_\mathrm{rand} \leftarrow \operatorname{RANDOM\_CONFIG}()$
%		\STATE $b_\mathrm{near}\leftarrow \operatorname{NEAREST\_NEIGHBOUR}(q_\mathrm{rand}, T) $
%		\STATE $u \leftarrow \operatorname{SELECT\_ACTION}(q_\mathrm{rand}, b_\mathrm{near})$
%
%		\STATE $b'  \leftarrow  \operatorname{SIMULATE}(q_\mathrm{rand}, b_\mathrm{near}, u)$
%		\IF{$\operatorname{IS\_VALID}(b')$}
%			\STATE $\mathcal	{B}_\mathrm{contingencies} \leftarrow \operatorname{BELIEF\_PARTITIONING}(b')$
%			 \FORALL{$b'' \in \mathcal{B}_\mathrm{contingencies}$}
%		 			\STATE $T \leftarrow T.\operatorname{ADD\_BELIEF}(b'')$
%		 			\STATE $T \leftarrow T.\operatorname{GOAL\_CONNECT}(b'', \mathcal{B}_\mathrm{connected})$
%
%			 \ENDFOR
%		\ENDIF			
%		\RETURN $T$
%	\end{algorithmic}
%\end{algorithm}

In summary, increased amounts of uncertainty can be handled by leveraging contact-exploiting actions and integrating contact sensing into planning to collapse uncertainty on funnel walls and differentiate between funnels reachable with the same action. Moreover, using a sampling-based approach for manipulation funnel exploration and dynamics programming technique for reusing contact-event-based contingencies maintains computational practicality even in high-dimensional configuration space and under significant motion uncertainty. 

%======================================================================
\section{Planning in High Complexity $\bm{\mathcal{C}}$-Spaces}
\label{sec: ceet}
%======================================================================

This section introduces a workspace-aware variant of the CERRT planner, the CET planner, and a contact-exploiting variant of the EET planner, the CEET planner, both published by~\cite{pall2023motion}. Both novel planners combine the best of both worlds (CERRT and EET): sequence manipulation funnels and guide sequencing by workspace information. The resulting planners are efficient on motion planning problems with moderate uncertainty and increased \Cs complexity.

With guided manipulation funnel sequencing, we simplify planning in complex $\mathcal{C}$-spaces because we avoid exploring \Cs regions that are disconnected in the workspace, and also avoid searching for funnel transitions in \Cs between distant or disconnected geometrical features of the environment. However, like CERRT, the conformant tree expansion allows only deterministic funnel transitions, and so, these planners can only handle moderate uncertainty accumulation.

% addresses motion planning in complex $\mathcal{C}$-spaces under low motion uncertainty. We consider a \Cs complex when the environment is composed of multiple obstacles and narrow passages because these make $\Cf$ fragmented into multiple regions with narrow passages, and computing or even approximating the connectivity of such a $\Cf$ is complicated. However, not all open $\Cf$ regions, surfaces, or narrow passages must be explored to find a motion plan between a given start and goal. 

% Similarly to the previous two sections, we follow a sampling-based planning approach and also use manipulation funnels to reduce uncertainty and reach a goal region with high probability. In contrast, we plan in a complex \Cscoma and so, we want to simplify planning by only exploring its relevant parts. We focus on relevant parts using work-space information about the task to prioritize task-relevant funnel exploration and use the environment's geometrical connectivity to bias funnel sequencing. 

%Moreover, we limit exploration to funnels that are task relevant and find actual connections between those funnels using a sampling-based approach as presented in Section~\ref{sec: cerrt}.

%
%\subsection{Problem Definition: Task-Relevant Belief-Space Planning}
%\label{sec: pd ceet}
%%------------------------------------------------------------------------------
\begin{figure}[tbp]
	\centering
	\def\svgwidth{1.0\linewidth}
	%% Creator: Inkscape inkscape 0.92.3, www.inkscape.org
%% PDF/EPS/PS + LaTeX output extension by Johan Engelen, 2010
%% Accompanies image file 'guarded_slide.pdf' (pdf, eps, ps)
%%
%% To include the image in your LaTeX document, write
%%   \input{<filename>.pdf_tex}
%%  instead of
%%   \includegraphics{<filename>.pdf}
%% To scale the image, write
%%   \def\svgwidth{<desired width>}
%%   \input{<filename>.pdf_tex}
%%  instead of
%%   \includegraphics[width=<desired width>]{<filename>.pdf}
%%
%% Images with a different path to the parent latex file can
%% be accessed with the `import' package (which may need to be
%% installed) using
%%   \usepackage{import}
%% in the preamble, and then including the image with
%%   \import{<path to file>}{<filename>.pdf_tex}
%% Alternatively, one can specify
%%   \graphicspath{{<path to file>/}}
%% 
%% For more information, please see info/svg-inkscape on CTAN:
%%   http://tug.ctan.org/tex-archive/info/svg-inkscape
%%
\begingroup%
  \makeatletter%
  \providecommand\color[2][]{%
    \errmessage{(Inkscape) Color is used for the text in Inkscape, but the package 'color.sty' is not loaded}%
    \renewcommand\color[2][]{}%
  }%
  \providecommand\transparent[1]{%
    \errmessage{(Inkscape) Transparency is used (non-zero) for the text in Inkscape, but the package 'transparent.sty' is not loaded}%
    \renewcommand\transparent[1]{}%
  }%
  \providecommand\rotatebox[2]{#2}%
  \newcommand*\fsize{\dimexpr\f@size pt\relax}%
  \newcommand*\lineheight[1]{\fontsize{\fsize}{#1\fsize}\selectfont}%
  \ifx\svgwidth\undefined%
    \setlength{\unitlength}{651.06330872bp}%
    \ifx\svgscale\undefined%
      \relax%
    \else%
      \setlength{\unitlength}{\unitlength * \real{\svgscale}}%
    \fi%
  \else%
    \setlength{\unitlength}{\svgwidth}%
  \fi%
  \global\let\svgwidth\undefined%
  \global\let\svgscale\undefined%
  \makeatother%
  \begin{picture}(1,0.47017186)%
    \lineheight{1}%
    \setlength\tabcolsep{0pt}%
    \put(0,0){\includegraphics[width=\unitlength,page=1]{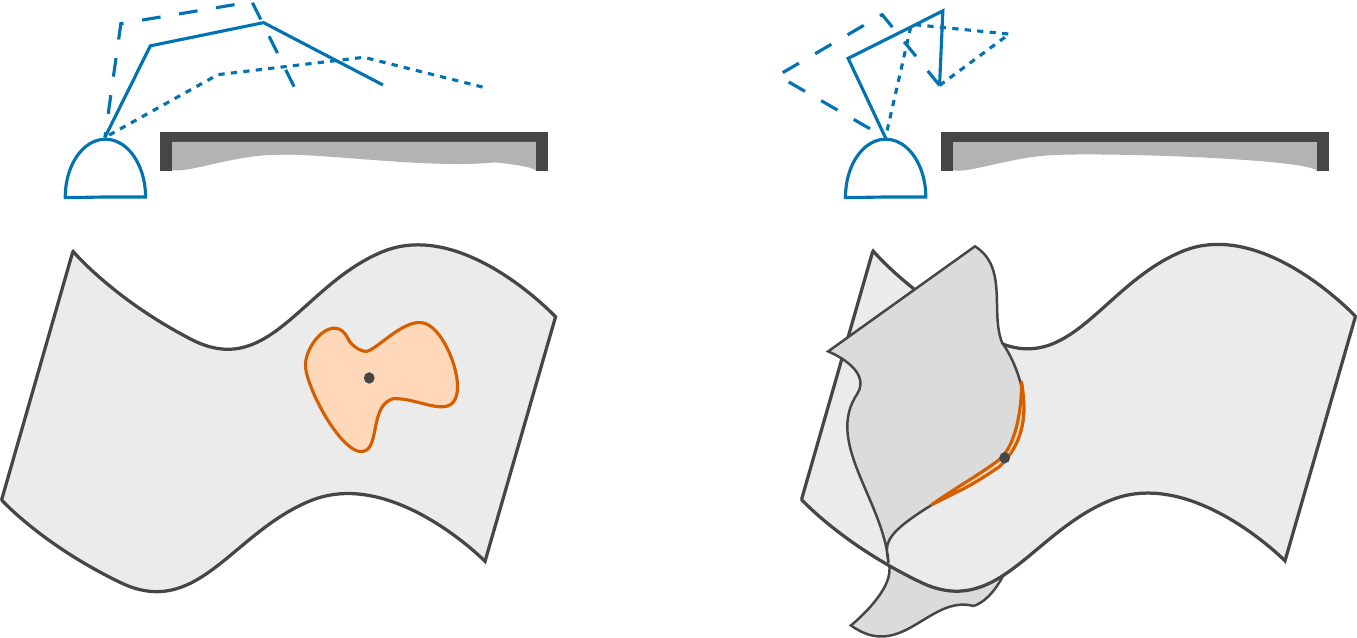}}%
    \put(0.02379766,0.09535983){\makebox(0,0)[lt]{\lineheight{1.25}\smash{\begin{tabular}[t]{l}$\mathcal{M}_F$\end{tabular}}}}%
    \put(0.26361417,0.23344702){\color[rgb]{0,0,0}\makebox(0,0)[lt]{\lineheight{0}\smash{\begin{tabular}[t]{l}$b$\end{tabular}}}}%
    \put(0.32809815,0.42925801){\makebox(0,0)[lt]{\lineheight{1.25}\smash{\begin{tabular}[t]{l}$\xrightarrow[]{\text{guarded slide}}$\end{tabular}}}}%
    \put(0.84292519,0.13584358){\makebox(0,0)[lt]{\lineheight{1.25}\smash{\begin{tabular}[t]{l}$\mathcal{M}_F$\end{tabular}}}}%
    \put(0.75166823,0.10755527){\color[rgb]{0,0,0}\makebox(0,0)[lt]{\lineheight{0}\smash{\begin{tabular}[t]{l}$b'$\end{tabular}}}}%
    \put(0.63071984,0.16004378){\makebox(0,0)[lt]{\lineheight{1.25}\smash{\begin{tabular}[t]{l}$\mathcal{M}_{F'}$\end{tabular}}}}%
    \put(0,0){\includegraphics[width=\unitlength,page=2]{guarded_slide.pdf}}%
  \end{picture}%
\endgroup%

    \caption{We illustrate a contact-exploiting action to explore manipulation funnel connectivity of the \textit{guarded-slide} actions (top row) and its uncertainty reduction effect (bottom row) on an RRR robot arm. The action moves the arm along an initially touched surface until the contact is lost, collapsing uncertainty to a line at the intersection of the horizontal and vertical surfaces or the intersection of two contact manifolds (bottom).}
    \label{fig: guarded move example}
\end{figure}

\subsection{Efficient Sequencing of Task-Relevant Funnels in Configuration Space}
\label{sec: insight ceer}
%------------------------------------------------------------------------------

We use workspace information to efficiently explore and sequence task-relevant funnels directly in configuration space. Similar to Section~\ref{sec: cerrt}, we implicitly sequence funnels by interleaving motion in free space and in contact. But now, we guide motions toward task-relevant regions of \Cscoma and so, we sequence task-relevant funnels. Since we only explore task-relevant regions of a complex \Cscoma we simplify planning and maintain computational practicality even in large volumes of \Cscoma like for a robot arm on a mobile base or humanoids. 

\begin{figure}[tbp]
\centering
%\def\svgwidth{0.46\linewidth}
%\input{Part2/img/sketches/region1_illustration.pdf_tex}
%\hfill
%\def\svgwidth{0.31\linewidth}
%\input{Part2/img/sketches/region2_illustration.pdf_tex}
%\hfill
\def\svgwidth{0.47\linewidth}
%% Creator: Inkscape inkscape 0.92.3, www.inkscape.org
%% PDF/EPS/PS + LaTeX output extension by Johan Engelen, 2010
%% Accompanies image file 'region2_sol_illustration.pdf' (pdf, eps, ps)
%%
%% To include the image in your LaTeX document, write
%%   \input{<filename>.pdf_tex}
%%  instead of
%%   \includegraphics{<filename>.pdf}
%% To scale the image, write
%%   \def\svgwidth{<desired width>}
%%   \input{<filename>.pdf_tex}
%%  instead of
%%   \includegraphics[width=<desired width>]{<filename>.pdf}
%%
%% Images with a different path to the parent latex file can
%% be accessed with the `import' package (which may need to be
%% installed) using
%%   \usepackage{import}
%% in the preamble, and then including the image with
%%   \import{<path to file>}{<filename>.pdf_tex}
%% Alternatively, one can specify
%%   \graphicspath{{<path to file>/}}
%% 
%% For more information, please see info/svg-inkscape on CTAN:
%%   http://tug.ctan.org/tex-archive/info/svg-inkscape
%%
\begingroup%
  \makeatletter%
  \providecommand\color[2][]{%
    \errmessage{(Inkscape) Color is used for the text in Inkscape, but the package 'color.sty' is not loaded}%
    \renewcommand\color[2][]{}%
  }%
  \providecommand\transparent[1]{%
    \errmessage{(Inkscape) Transparency is used (non-zero) for the text in Inkscape, but the package 'transparent.sty' is not loaded}%
    \renewcommand\transparent[1]{}%
  }%
  \providecommand\rotatebox[2]{#2}%
  \newcommand*\fsize{\dimexpr\f@size pt\relax}%
  \newcommand*\lineheight[1]{\fontsize{\fsize}{#1\fsize}\selectfont}%
  \ifx\svgwidth\undefined%
    \setlength{\unitlength}{934.90866834bp}%
    \ifx\svgscale\undefined%
      \relax%
    \else%
      \setlength{\unitlength}{\unitlength * \real{\svgscale}}%
    \fi%
  \else%
    \setlength{\unitlength}{\svgwidth}%
  \fi%
  \global\let\svgwidth\undefined%
  \global\let\svgscale\undefined%
  \makeatother%
  \begin{picture}(1,0.7271198)%
    \lineheight{1}%
    \setlength\tabcolsep{0pt}%
    \put(0,0){\includegraphics[width=\unitlength,page=1]{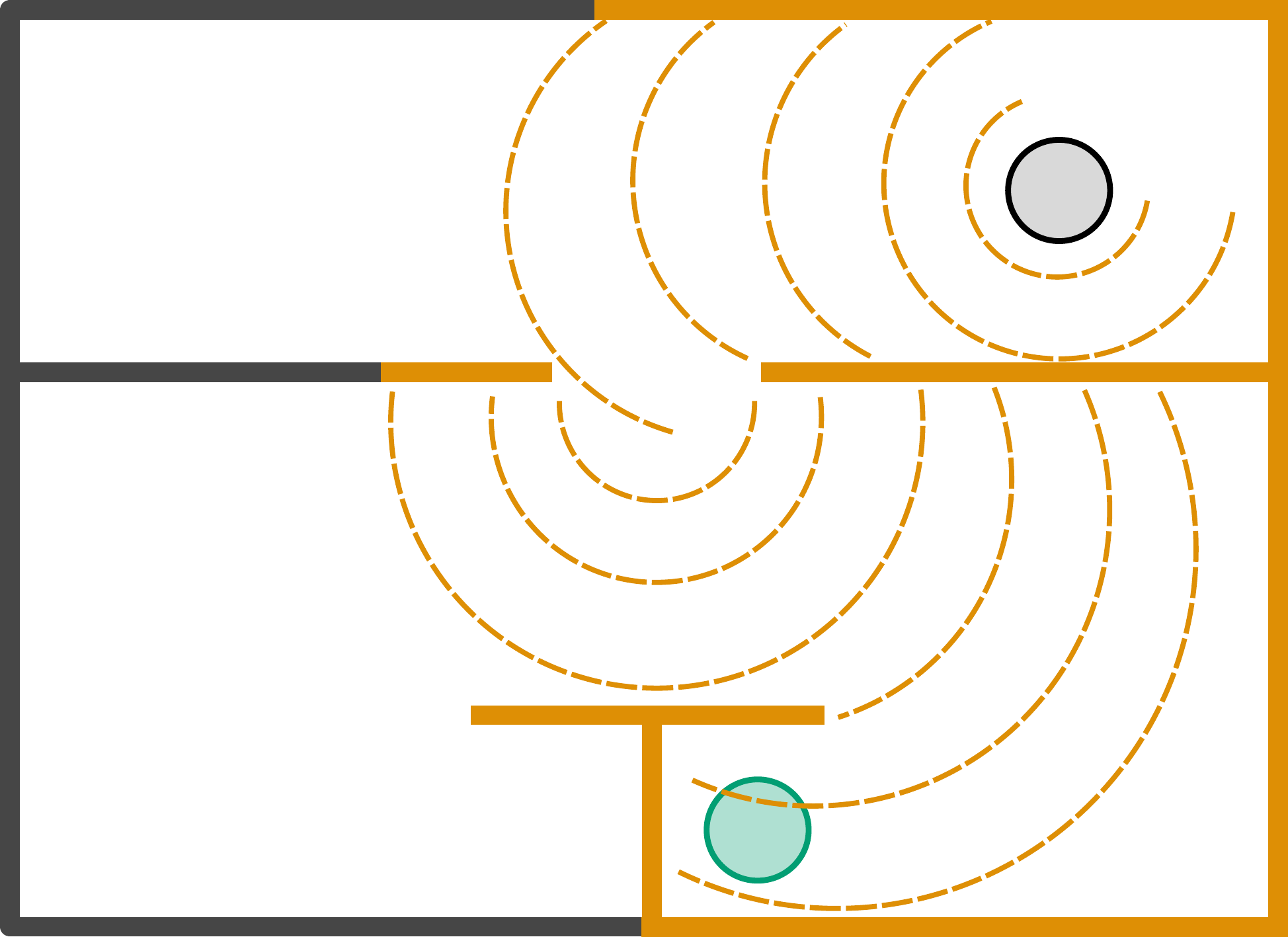}}%
    \put(0.56172435,0.06508022){\makebox(0,0)[lt]{\lineheight{1.25}\smash{\begin{tabular}[t]{l}$b_g$\end{tabular}}}}%
    \put(0,0){\includegraphics[width=\unitlength,page=2]{region2_sol_illustration.pdf}}%
    \put(0.79548984,0.55994179){\makebox(0,0)[lt]{\lineheight{1.25}\smash{\begin{tabular}[t]{l}$b_0$\end{tabular}}}}%
    \put(0,0){\includegraphics[width=\unitlength,page=3]{region2_sol_illustration.pdf}}%
    \put(0.79642104,1.43925157){\makebox(0,0)[lt]{\lineheight{1.25}\smash{\begin{tabular}[t]{l}S\end{tabular}}}}%
    \put(-0.35813978,1.42514651){\makebox(0,0)[lt]{\lineheight{1.25}\smash{\begin{tabular}[t]{l}S\end{tabular}}}}%
    \put(0,0){\includegraphics[width=\unitlength,page=4]{region2_sol_illustration.pdf}}%
%    \put(1.93362639,-0.23293722){\makebox(0,0)[lt]{\lineheight{1.25}\smash{\begin{tabular}[t]{l}G\end{tabular}}}}%
%    \put(1.56972535,-0.3348216){\makebox(0,0)[lt]{\lineheight{1.25}\smash{\begin{tabular}[t]{l}S\end{tabular}}}}%
    \put(0,0){\includegraphics[width=\unitlength,page=5]{region2_sol_illustration.pdf}}%
  \end{picture}%
\endgroup%

\hfill
\def\svgwidth{0.47\linewidth}
%% Creator: Inkscape inkscape 0.92.3, www.inkscape.org
%% PDF/EPS/PS + LaTeX output extension by Johan Engelen, 2010
%% Accompanies image file 'shpereTunnels.pdf' (pdf, eps, ps)
%%
%% To include the image in your LaTeX document, write
%%   \input{<filename>.pdf_tex}
%%  instead of
%%   \includegraphics{<filename>.pdf}
%% To scale the image, write
%%   \def\svgwidth{<desired width>}
%%   \input{<filename>.pdf_tex}
%%  instead of
%%   \includegraphics[width=<desired width>]{<filename>.pdf}
%%
%% Images with a different path to the parent latex file can
%% be accessed with the `import' package (which may need to be
%% installed) using
%%   \usepackage{import}
%% in the preamble, and then including the image with
%%   \import{<path to file>}{<filename>.pdf_tex}
%% Alternatively, one can specify
%%   \graphicspath{{<path to file>/}}
%% 
%% For more information, please see info/svg-inkscape on CTAN:
%%   http://tug.ctan.org/tex-archive/info/svg-inkscape
%%
\begingroup%
  \makeatletter%
  \providecommand\color[2][]{%
    \errmessage{(Inkscape) Color is used for the text in Inkscape, but the package 'color.sty' is not loaded}%
    \renewcommand\color[2][]{}%
  }%
  \providecommand\transparent[1]{%
    \errmessage{(Inkscape) Transparency is used (non-zero) for the text in Inkscape, but the package 'transparent.sty' is not loaded}%
    \renewcommand\transparent[1]{}%
  }%
  \providecommand\rotatebox[2]{#2}%
  \newcommand*\fsize{\dimexpr\f@size pt\relax}%
  \newcommand*\lineheight[1]{\fontsize{\fsize}{#1\fsize}\selectfont}%
  \ifx\svgwidth\undefined%
    \setlength{\unitlength}{963.99994617bp}%
    \ifx\svgscale\undefined%
      \relax%
    \else%
      \setlength{\unitlength}{\unitlength * \real{\svgscale}}%
    \fi%
  \else%
    \setlength{\unitlength}{\svgwidth}%
  \fi%
  \global\let\svgwidth\undefined%
  \global\let\svgscale\undefined%
  \makeatother%
  \begin{picture}(1,0.72614111)%
    \lineheight{1}%
    \setlength\tabcolsep{0pt}%
    \put(0,0){\includegraphics[width=\unitlength,page=1]{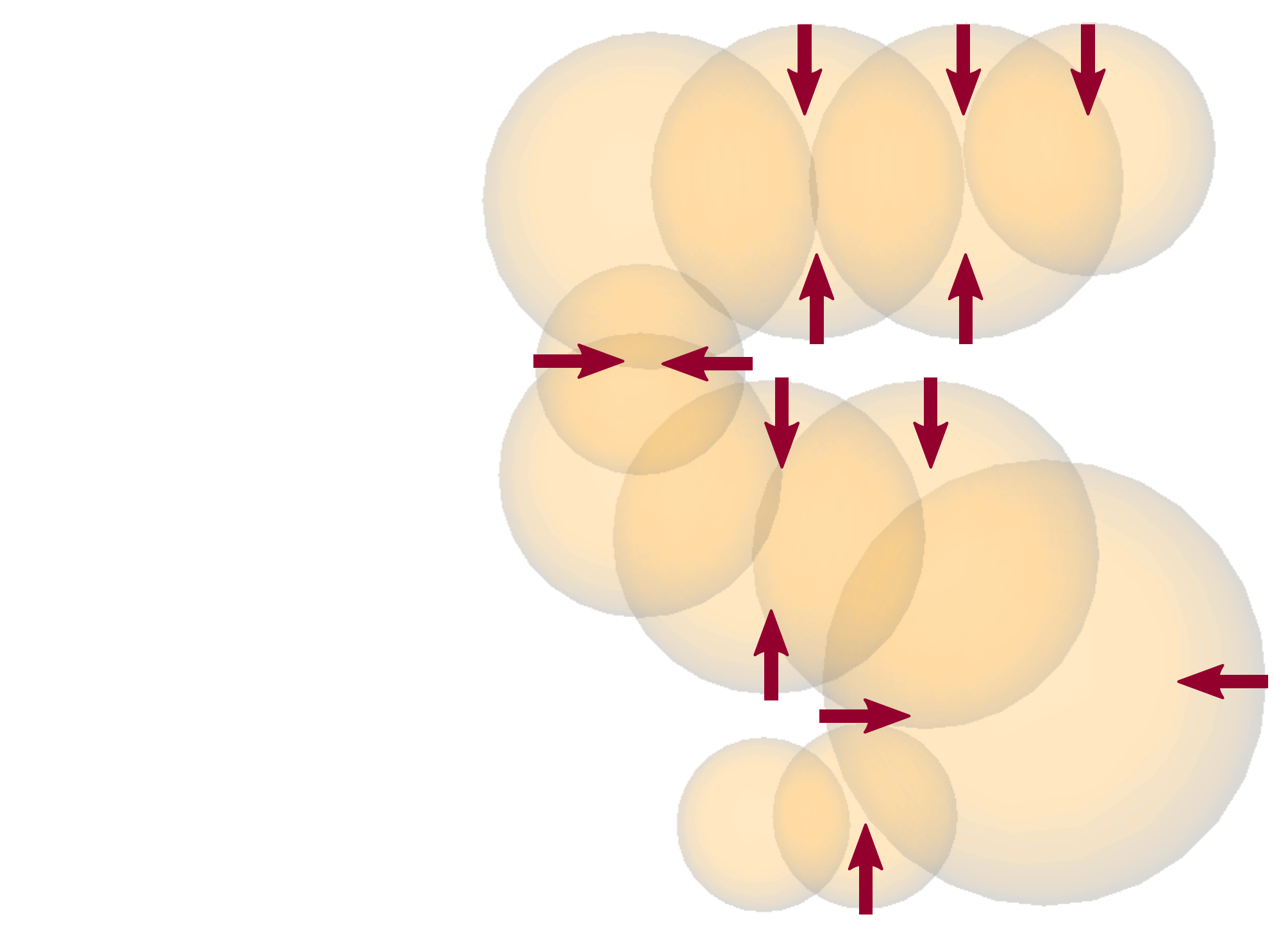}}%
    \put(0.87277715,0.62159162){\color[rgb]{0,0,0}\makebox(0,0)[lt]{\lineheight{1.25}\smash{\begin{tabular}[t]{l}$s_0$\end{tabular}}}}%
    \put(0,0){\includegraphics[width=\unitlength,page=2]{shpereTunnels.pdf}}%
    \put(0.51616079,0.08826253){\color[rgb]{0,0,0}\makebox(0,0)[lt]{\lineheight{1.25}\smash{\begin{tabular}[t]{l}$s_{10}$\end{tabular}}}}%
    \put(0,0){\includegraphics[width=\unitlength,page=3]{shpereTunnels.pdf}}%
  \end{picture}%
\endgroup%

    \caption[Task-relevant workspace region illustration.]{\textit{Left:} Illustration of wavefront expansion to identify task-relevant free and contact regions (orange) shown in a 2D navigation problem give $b_0$ start belief and $\bg$ goal. \textit{Right:} A sequence of ten spheres approximating the task-relevant free space region and surface normals (arrows) indicating nearby surfaces of the spheres.}
    \label{fig: regions illustration}
\end{figure}

Workspace information enables computationally efficient guided exploration because task-relevant free-space regions can be efficiently computed by decomposing a robot’s workspace using a wavefront planner, as demonstrated by \citet{rickert_balancing_2014}. It is computationally less expensive to obtain from workspace decomposition than from \Cs because the workspace is generally lower dimensional than \Csdot Moreover, workspace information allows efficient funnel sequencing as well because funnel regions (i.e., contact surfaces) and their connectivity are directly encoded in the geometrical connectivity of the workspace. Therefore, we can filter task-relevant funnels and search for transitions only between physically connected funnels to find a sequence between a given start and goal region. Suppose waves are expended from a given start to a goal, as illustrated in Figure~\ref{fig: regions illustration}, then, the waves cover a task-relevant free space region and touch task-relevant surfaces.

Workspace information can guide $\mathcal{C}$-space exploration because a joint-space configuration can be moved efficiently toward a workspace pose when the robot's kinematic model is known. We use the kinematic model to check if a configuration is in a given workspace region for sampling and a robot's Jacobian to move a configuration toward a pose. Since we use biased sampling and the Jacobian to approximate task-relevant workspace regions, we denote the approximated free-space $\widetilde{\mathcal{C}}_\text{task}$ and contact $\widetilde{\delta\mathcal{C}}_\text{task}$ regions.

Workspace decomposition can efficiently approximate task-relevant \Cs regions, but since this is only an approximation, incomplete information can lead to local minimums. Thus, we must balance the exploitation of incomplete information with exploration to gain new information~\citep{rickert_balancing_2008}. Moreover, we want to balance exploring motion in contact and free space because contact-space exploration requires more computation than free-space exploration due to frequent collision checking.

We propose to balance exploitation and exploration and balance $\widetilde{\mathcal{C}}_\text{task}$ and $\widetilde{\delta\mathcal{C}}_\text{task}$ exploration similarly. The balancing strategy is similar because we \emph{shift} from exploitation to exploration and also \emph{shift} from exploring $\widetilde{\mathcal{C}}_\text{task}$ to $\widetilde{\delta\mathcal{C}}_\text{task}$. For the former, we initially exploit workspace information to quickly reach the next sphere. We start exploring alternative paths to the next sphere if pure exploitation fails. For the latter, we initially explore $\widetilde{\mathcal{C}}_\text{task}$ to reduce collision checking. If we cannot find a path in $\widetilde{\mathcal{C}}_\text{task}$, we start exploring $\widetilde{\delta\mathcal{C}}_\text{task}$ to reduce state uncertainty and improve \Cs connectivity. 

\begin{figure*}[tb]
	\centering	
	\def\svgwidth{0.95\linewidth}
	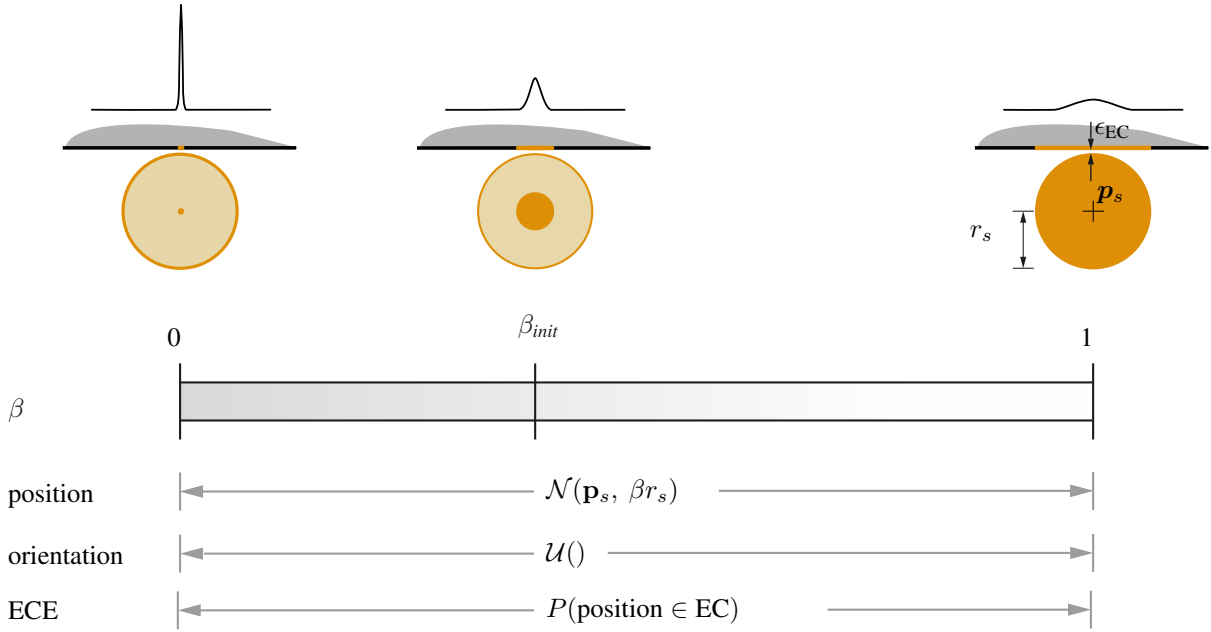
    \caption[Balancing exploitation-exploration and free- and contact-space exploration.]{The value of $\beta$ balances a) exploitation and exploration of workspace information and b) free and contact space exploration. a) to balance exploitation and exploration, a pose sample is drawn from a sphere $s$ (orange circles) where the position is from a normal distribution with variance proportional to $\beta$ and the orientation is from a uniform distribution. b) to balance free and contact space exploration, the previously drawn sample is projected with $\beta$ probability to a surface within $\epsilon_\text{EC}$ distance away from $s$.}%todo{fix img when using IJRR template}
    \label{fig: beta explora and expliot}
\end{figure*}

We start shifting from exploitation to exploration and from exploring $\widetilde{\mathcal{C}}_\text{task}$ to $\widetilde{\delta\mathcal{C}}_\text{task}$ for the same reason, i.e., when progress toward the next sphere is hindered. Hence, we use a single variable $\beta \in [0,~1]$ to measure progress and to balance both exploitation and exploration and exploring $\widetilde{\mathcal{C}}_\text{task}$ and $\widetilde{\delta\mathcal{C}}_\text{task}$, as illustrated in Figure~\ref{fig: beta explora and expliot}.

With $\beta=0$, the planner purely exploits workspace information and behaves like a potential field planner using a global navigation function approximated with the sequence of spheres $S$. Thus, we drag a robot's tool frame toward the next sphere's center $\vec{p}_s$ in free space, and so, we explore a small region of $\widetilde{\mathcal{C}}_\text{task}$.

When pure exploitation fails, $\beta$ increases, and exploration starts. Initially, we sample a pose with high probability close to the center of $s$ by drawing an orientation from a uniform distribution and a position from a Gaussian distribution $\mathcal{N}(\mathbf{p}_s, \beta r_s)$ around the centers of $s$ with a variance proportional to $\beta$. Hence, $\beta$ also indicates how closely the planner follows the potential field defined by task-relevant workspace information. We balance $\widetilde{\mathcal{C}}_\text{task}$ and $\widetilde{\delta\mathcal{C}}_\text{task}$ exploration by projecting the previously sampled pose onto a randomly selected neighboring surface of $s$ with $\beta$ probability.

When $\beta = 1$, samples are likely drawn anywhere inside a sphere and projected on a neighboring surface; $\beta >1 $ indicates that the planner cannot reach the next sphere and is stuck. To unstuck the planner, we will use random exploration until a new sphere is reached, but note that other strategies (e.g., backtracking to a previous sphere) can also be used.

\subsection{Implementation: Contact EET Planners}
\label{sec: alg ceet}
%------------------------------------------------------------------------------

Based on the previous insights, we provide two algorithmic instantiations. Both algorithms compute a conformant motion plan using manipulation funnels to collapse and bound uncertainty. Still, the two algorithms use different amounts of workspace information to guide exploration: the Contact Exploring Tree (CET) planner uses task-relevant regions to bias sampling, and the Contact Exploring/Exploiting Tree (CEET) planner uses the same regions to bias sampling and action selection. 

We extract task-relevant regions from a given geometrical model of an environment using sphere-based wavefront expansion in workspace~\citep{brock_decomposition-based_2001}. First, we compute a tree of workspace spheres in free space between a given start and goal. Then, we extract a sequence of spheres connecting the start and goal, and we save neighboring surfaces and contact normals for each sphere, as shown in Figure~\ref{fig: regions illustration}. The extracted \emph{spheres} $s_{i=1..n}$ approximate a task-relevant sub-space in the robot's workspace, and configurations within these spheres approximates $\widetilde{\mathcal{C}}_\text{task}$. We approximate $\widetilde{\delta\mathcal{C}}_{s_i}$ of sphere $s_i$ by projecting $\widetilde{\mathcal{C}}_{s_i}$ on \emph{neighboring surfaces} of the respective sphere. By knowing the \emph{sequence} of spheres between a start and a goal, we also know how a robot should move within a sphere or on neighboring surfaces.

\textbf{The CET planner} is identical with the CERRT planner (Algorithm~\ref{alg:cerrt}) except that it restricts randomly drawn configuration to be inside $S$ forcing to 
 explore $\widetilde{\mathcal{C}}_\text{task}$. Even though the planner uses only a fraction of the information from workspace decomposition, it provides significant benefits when planning in complex environments because it reduces the search space volume while also exploring task-relevant funnels due to the random free-space or contact-exploiting action selection.  

\textbf{The CEET planner} presented in Algorithm~\ref{alg:ceet planner} guides exploration by leveraging all three types of information obtained from workspace decomposition: $\widetilde{\mathcal{C}}_\text{task}$ from $S$ (i.e., inner volume of funnels), $\widetilde{\delta\mathcal{C}}_{s_i}$ from neighboring surfaces of $S$ (walls of funnels), and $s_i \in S$ connectivity (sequence of funnels). The CEET planner combines manipulation funnel exploitation, like the CERRT planner (Section~\ref{sec: cerrt}), and workspace-information guided exploration/exploitation, like the Exploring/Exploiting Tree (EET) planner~\citep{rickert_balancing_2014}.

Similarly to the planners presented in Section~\ref{sec: alg cerrt} and~\ref{sec: alg concerrt}, CEET is a belief-space planner approximating a belief with a set of particles. Like CERRT and CET, it is a conformant single-query planner growing a tree from a given start belief to a goal. However, in every iteration, it expands a node toward the next sphere $s_\text{next}$. If the resulting belief $b_\text{new}$ is valid (Equation~\eqref{eq: valid cs}), it is added to the tree. Next, the algorithm tries to connect $b_\text{new}$ and $b_g$, and upon success, returns the tree. Otherwise, it checks if a new sphere (e.g., the next sphere or one even closer to the goal) is reached. When a new sphere is reached, it updates $s$, resets $\beta = \beta_\text{init}$, and initialize the stuck counter $k_s$ with zero. However, if $b_\text{new}$ is invalid or a new sphere is not reached, it increases $\beta$ to shift toward exploration and use more contact-exploiting actions. We increase $\beta$ faster when expansion fails because it strongly indicates that uncertainty hinders progress. 

If $\beta > 1$ at the end of the while loop, the planner reaches the exploration limit. Hence, it resets $\beta$ and increases the stuck counter $k_s$. The number of times we explore the same sphere depends on the size of the current sphere relative to the next one. Moving from a smaller sphere toward a larger one is like exiting a narrow passage. In this case, we hope to have more feasible paths than entering a narrow passage. Thus, we restart the exploration of the same sphere multiple times. In contrast, we restart exploration fewer times when we go from a larger sphere to a smaller one (entreating a narrow passage). When $k_s$ reaches this adaptive threshold, the planner starts randomly exploring \Cs similarly to CERRT. It switches from random exploration back to guided exploration when a yet unvisited sphere is reached.

\begin{algorithm}[tbh]
    \caption{CEET planner}
	\label{alg:ceet planner}
	\begin{algorithmic}[1]
		\small
		\REQUIRE ${b}_0, {b}_{\mathrm{g}}$
		\ENSURE $T = (V,E)$
		\STATE $V \leftarrow  {b}_0 $ \COMMENT{initialize tree}
		\STATE $S, EC \leftarrow \operatorname{WAVEFRONT}(\mu_{{b}_0},\mu_{\bg})$\COMMENT{compute sphere tree and associated ECs%~\citep{brock_decomposition-based_2001}
		}
		\STATE $s \leftarrow  S_\text{begin}$ \COMMENT{take first sphere}		
		\STATE $\beta \leftarrow \beta_\text{init}$ \COMMENT{initialize exploration-exploitation balance}		
		\STATE $k_s \leftarrow 0$ \COMMENT{initialize sphere exploration counter}		
		\WHILE{true}		
		
%			\IF{$k_s = 0$}
%				\STATE $b_\text{new} \leftarrow \operatorname{EXPLOIT}(T, s)$\COMMENT{only exploit workspace information}
%			\ELSIF{$k = 1$}
%				\STATE $b_\text{new} \leftarrow \operatorname{EXPLORE}(T, s, EC = \emptyset, \beta)$\COMMENT{explore $\Cf$ region only to find a robust action}
%			\ELSE		
			\STATE $b_\text{near},\: b_\text{new} \leftarrow \operatorname{EXPAND}(V, s_\text{next}, EC_{s_\text{next}} \cup EC_{s}, \beta)$ \COMMENT{expend one node in the tree}
%			\ENDIF
		
			\IF{$ \operatorname{IS\_VALID}(b_\text{new})$}%is in the task-relevant space with b_\text{new} \subset \mathcal{B}_\text{task} \wedge 
				\STATE $V \leftarrow V \cup \{b_\text{new}\}$ \COMMENT{add new node to the tree}
				\STATE $E \leftarrow E \cup \{(b_\text{near},b_\text{new})\}$ \COMMENT{add new edge to the tree}
				\IF{$\operatorname{GOAL\_CONNECT}(b_\text{new}, \bg)$}
					\RETURN $T$ \COMMENT{the goal is reached from the new belief}
				\ENDIF
				
				\IF[new belief reached an new sphere]{$\vec{T}_{b_\text{new}} \in s_\text{unvisited} ~\vee~ {s_\text{unvisited} \in S} $}
					\STATE $s \leftarrow s_\text{unvisited}$ \COMMENT{update current sphere}
					\STATE $\beta \leftarrow \beta_\text{init}$\COMMENT{reset exploration-exploitation balance}		
					\STATE $k_s \leftarrow 0$\COMMENT{reset sphere exploration counter}
				\ELSE
					\STATE $ \beta \leftarrow \beta(1+\alpha)$ \COMMENT{increase exploration when sphere is not reached}						
				\ENDIF
	
			\ELSE%[expansion failed]
				\STATE $ \beta \leftarrow \beta(1+\alpha)^2$ \COMMENT{increase exploration when node expansion failed}
			\ENDIF
		
			\IF[exploration limit reached]{$\beta > 1 $}
				\STATE $\beta \leftarrow \beta_\text{init}$ \COMMENT{reset exploration-exploitation balance}		
				\STATE $k_s \leftarrow k_s+1$ \COMMENT{increase the stuck counter for $s$}
%				\STATE $ s \leftarrow \operatorname{INCREASE\_RADIUS}(s, (1+\alpha))$  \COMMENT{increase the radius of the current sphere}
%				\STATE $  EC_s \leftarrow \operatorname{UPDATE}(s)$ \COMMENT{extend list of ECs near the currently inflated sphere}
%				\IF[stuck due to a local minima or closed passage by uncertainty]{$ s_\text{parent} \subset s$}
				\IF{$ r_\text{next} + r_s \geq r_\text{next}*(1+\alpha)^{k_s}$}
					\STATE $s \leftarrow s_\text{parent}$	\COMMENT{backtrack to parent sphere}
					\STATE $k_s \leftarrow 0$ \COMMENT{reset sphere exploration counter for parent sphere}
				\ENDIF			
			\ENDIF			
		\ENDWHILE		
	\end{algorithmic}	
\end{algorithm}

Guided exploration is realized in the \textit{expand} method (line~7), which has steps similar to the CERRT planner but implemented radically differently. First, as opposed to random configuration sampling, we sample a frame $\vec{T}$ inside a sphere $s$: 
\begin{align*}
\vec{p}_\text{sample} &= \vec{p}_s + \mathcal{N}(0,\beta r_s),\\
R_\text{sample} &= \mathcal{U}(),
\end{align*}
where the position $\vec{p}_\text{sample}$ is a sample from a Gaussian distribution with center at the sphere $\vec{p}_s$ and $\beta r_s$ variance, and the orientation $R$ is sampled from a uniform distribution~$\mathcal{U}()$. We project the sample on a randomly selected $EC$ with $\beta$ probability. Thus, a large value of $\beta$ increases a sphere's exploration and the use of contact-exploiting action.  

%Below, we give implementation details for each step, where the numbers in parentheses refer to lines in Algorithm~\ref{alg:ceet explore step}.

%\begin{algorithm}[tbh]
%    \caption{EXPAND}
%	\label{alg:ceet explore step}
%	\begin{algorithmic}[1]		
%		\small	
%		\REQUIRE $V,s,EC,\beta$		
%		\ENSURE ${b}_\text{near},{b}_\text{new}$
%		\STATE $\vec{T}_\text{sample} \leftarrow \operatorname{SAMPLE}(s, r_s, \beta)$\COMMENT{sample target pose from $s$ and its ECs}
%		\STATE $b_\text{near} \leftarrow \operatorname{NEAREST\_NEIGBOUR}(V, \vec{T}_\text{sample})$ \COMMENT{find nearest node to sampled pose}
%		\STATE $\vec{q}_\text{target}, \:u \leftarrow \operatorname{STEER}(\bm{\mu}_{b_\text{near}}, \vec{T}_\text{sample}, \normal_{EC}, \beta)$\COMMENT{pull robot toward sampled pose}			
%		\STATE $b_\text{new} \leftarrow \operatorname{SIMULATE}(b_\text{near}, \vec{q}_\text{target}, u)$	\COMMENT{forward propagate $b_\text{near}$ toward $q_\text{target}$ with action $u$}
%		\RETURN ${b}_\text{near},\:{b}_\text{new}$
%	\end{algorithmic}	
%\end{algorithm}

Secondly, the nearest neighbor's distance metric uses $d_{\bm{\Sigma}}$ as before but combines it with the new $d_\vec{T}(b) = \|\vec{T}_b - \vec{T}_\text{sample} \|_2$ spatial distance between the sampled frame and the end-effector (or tool) frame of a belief $b$ as opposed to $d_\vec{q}$ joint space distance used in CERRT and ConCERRT. So, the nearest neighbor selection is as follows:
$$ b_\text{near} = \underset{b \in T}{argmin} \left[ \gamma\cdot d_{\bm{\Sigma}}(b) +(1-\gamma)\cdot d_\vec{T}(b) \right],$$
where $\gamma$ plays the same role as explained for CERRT in Section~\ref{sec: alg cerrt}: a large value of $\gamma$ enforces the planner to expand low uncertainty nodes and a low-value weights less state uncertainty but favor spatially closer nodes.

The third step computes a target configuration $\vec{q}_\text{target}$ and appropriate action by dragging the end-effector (or tool) frame on the robot toward the previously sampled frame $\vec{T}_\text{sample}$. We choose between a \textit{connect} or \textit{slide} action based on the initial contact state of ${b}_\text{near}$ and whether the potential field breaks contact or allows sliding. We simulate an action by moving the robot's frame in a straight line using the robot's Jacobian. With sliding actions, we maintain contact with a surface using the task projection method~\citep{stilman_task_2007}. If the contact state changes after a simulation step, we update $u$ with the action needed to reach that state. A \textit{connect} action changes to \textit{guarded move} for a new contact, and \textit{sliding} changes to \textit{guarded sliding} if the initial contact state of ${b}_\text{near}$ changes. %The algorithm returns the reached configuration and the associated action if the action changed during simulation or if the target is reached with $\epsilon_\text{position}$ precision.  

The final step is identical to the CERRT simulate method: We forward propagate $b_\text{near}$ toward the target configuration $\vec{q}_\text{target}$ by simulating the chosen action $u$ with a noisy motion model for each particle in the belief. 

In summary, workspace decomposition can provide structural context to sequence and explore task-relevant manipulation funnels. Planning becomes more efficient in complex environments by only exploring a task-relevant subspace. Note that all three conformant planners presented in this paper use some amount of workspace or task information to speed up planning: CERRT uses the least amount of information about the task with goal-biased sampling, CET uses more information to bias sampling to task-relevant regions, including the goal region, and CEET uses the most information to bias sampling and action selection. Moreover, all three conformant planners, CERRT, CET, and CEET, can be used for contingent planning by following the approach presented in Section~\ref{sec: concerrt} and obtain their contingent variants: ConCERRT, ConCET, and ConCEET. In the next section, we evaluate experimentally all three conformant planners and their contingent versions.

\section{Evaluation of ECE-Based \Cs Planners}
\label{sec: ceet eval}
%------------------------------------------------------------------------------

We evaluated all six planners (CERRT, CET, CEET, ConCERRT, ConCET, and ConCEET) in simulation to show that 1) all approaches scale to high-dimensional \Cs planning under uncertainty; 2) as an environment's structural complexity increases, leveraging structural context simplifies planning; and 3) while moderate uncertainty can be handled using manipulations funnels alone, handling an increased amount of uncertainty also needs belief partitioning with contact events. The following experiments are also supported by previous observations made in \citep{sieverling_interleaving_2017} and \citep{pall_contingent_2018}; however, we now give a complete picture of how and when to use ECE for motion planning considering uncertainty and \Cs complexity. %\footnote{link to git repo \todo{discuss it with Oliver}}

Since we want to compare our conformant and contingent belief-space planners, we define the success metric $P_\text{scussess}$ considering both the planning time and the quality of a plan: 
\begin{equation}
P_\text{success} = P(\pi)\cdot\frac{N_\text{succ}(t)}{N},
\label{eq: sucess metric bs planners}
\end{equation}
where $P(\pi)$ is the success probability of the policy (equal to 1 for conformant planners), and $\frac{N_\text{succ}(t)}{N}$ is the ratio of found solutions for $N$ samples, i.e., the planning success rate under a given time budget. We ran 20 experiments per setup for all planning problems to sample $P_\text{scussess}$ under the planning parameters shown in Table~\ref{tab:params descritption and experiment setup CEET}.     

\begin{table}[]
\begin{tabularx}{\linewidth}{|l|p{2.5cm}|X|X|}
\hline
\textbf{Param.} & \textbf{Description} & {\textbf{2DOF}  \textbf{gripper}}  & {\textbf{7DOF}  \textbf{WAM}}  \\ \hline
$t$ [min] & time budget  & 500 & 50  \\ \hline
$N$ & number of particles & 20 & 20 \\ \hline
$\delta_\text{step}$ & simulation step size &$0.05$ & $1.0$  \\ \hline
$\epsilon_\text{goal}$ & goal region & 0.1 & 0.05 \\ \hline
$\gamma$ & weighting factor for uncertainty reduction & 0.3 & 0.6  \\ \hline
$\beta_\text{init}$ & initial exploration and exploitation balance& 0.1 & 0.6  \\ \hline
$\alpha$ & rate of $\beta$ increase & 0.1 & 0.08 \\ \hline
$\bm{\sigma}_\text{init}$ & initial uncertainty & $[0.2,0.2]$ & $\vec{0}^7$  \\ \hline
$\bm{\sigma}_\text{motion}$ & motion uncertainty & $[0.2,0.2]$ & $[\sigma_m \cdot \vec{1}^6, 0.0]$  \\ \hline
\end{tabularx}
 \vspace{.5em}
\caption{Planning parameters, where $\vec{0}^n$ and $\vec{1}^n$ are an $n$ dimensional vector with zeroes or ones, respectively, and $[\vec{1}^n, x]$ is the concatenation of $\vec{1}^n$ with a scalar $x$ resulting in an $n+1$ dimensional vector.}
\label{tab:params descritption and experiment setup CEET}
 \vspace{-1.5em}
\end{table}

%-----------------------------------------------------------------
\subsection{Contact-Based Belief-Space Planners Scales to high-dimensional $\mathcal{C}$-spaces} 
%-----------------------------------------------------------------

Our first experiment considered low-dimensional \Cs planning inspired by the \textit{grasping POMDP} problem from the POMDP literature~\citep{hsiao_grasping_2007}, where a 2DOF gripper can only translate, and its fingers are tactile sensors. We considered three environments with different structural complexity (top row in Figure~\ref{fig:mean_duration cspace eval}). Since most belief space planners rely on pre-defined discretization, those planners become computationally untractable in complex environments. We evaluated our conformant planners in these low-dimensional \Cs planning problems, and the planners could compute a motion plan even in a complex and large maze-like environment. We discuss these results (bottom row in Figure~\ref{fig:mean_duration cspace eval}) in detail below. 

Our second experiment considered a higher dimensional \Cs planning problem from the Particle-RRT motion planning literature~\citep{phillips-grafflin_planning_2020}, where a robot arm, for us a 7DOF Barrett WAM arm, had to insert its end-effector into a deepening on a wall under uncertainty. To reduce uncertainty, the robot could measure the contact normal of the surfaces with a ball-shaped end-effector. The results for this planning problem (Figure~\ref{fig:7d success vs uncertainty}) proved that the conformant CERRT, CET, and CEET planners and their contingent variants ConCERRT, ConCET, and ConCEET were efficient enough to compute policies directly in a 7-dimensional configuration space under motion uncertainty. Without a fixed discretization of the contact space, these problems are not solvable for POMDP-based motion planners \citep{hsiao_grasping_2007, koval_pre-_2016}, which easily become intractable in high-dimensional spaces and complex environments. Our contingent planners handled increased amounts of uncertainty without requiring the inverse kinematics of the robot and fully observable joint states~\citep{koval_configuration_2020} or assuming reversible actions and fully observable joint states~\citep{phillips-grafflin_planning_2020}.

%-----------------------------------------------------------------
\subsection{Guided Funnel Sequencing Simplifies Planning in Complex $\mathcal{C}$-spaces} 
\label{sec: ceet exp1}
%-----------------------------------------------------------------

\begin{figure}[tbp]
\centering
\def\svgwidth{1\linewidth}
%% Creator: Inkscape inkscape 0.92.3, www.inkscape.org
%% PDF/EPS/PS + LaTeX output extension by Johan Engelen, 2010
%% Accompanies image file 'planningProblems_CS.pdf' (pdf, eps, ps)
%%
%% To include the image in your LaTeX document, write
%%   \input{<filename>.pdf_tex}
%%  instead of
%%   \includegraphics{<filename>.pdf}
%% To scale the image, write
%%   \def\svgwidth{<desired width>}
%%   \input{<filename>.pdf_tex}
%%  instead of
%%   \includegraphics[width=<desired width>]{<filename>.pdf}
%%
%% Images with a different path to the parent latex file can
%% be accessed with the `import' package (which may need to be
%% installed) using
%%   \usepackage{import}
%% in the preamble, and then including the image with
%%   \import{<path to file>}{<filename>.pdf_tex}
%% Alternatively, one can specify
%%   \graphicspath{{<path to file>/}}
%% 
%% For more information, please see info/svg-inkscape on CTAN:
%%   http://tug.ctan.org/tex-archive/info/svg-inkscape
%%
\begingroup%
  \makeatletter%
  \providecommand\color[2][]{%
    \errmessage{(Inkscape) Color is used for the text in Inkscape, but the package 'color.sty' is not loaded}%
    \renewcommand\color[2][]{}%
  }%
  \providecommand\transparent[1]{%
    \errmessage{(Inkscape) Transparency is used (non-zero) for the text in Inkscape, but the package 'transparent.sty' is not loaded}%
    \renewcommand\transparent[1]{}%
  }%
  \providecommand\rotatebox[2]{#2}%
  \newcommand*\fsize{\dimexpr\f@size pt\relax}%
  \newcommand*\lineheight[1]{\fontsize{\fsize}{#1\fsize}\selectfont}%
  \ifx\svgwidth\undefined%
    \setlength{\unitlength}{3066.16445466bp}%
    \ifx\svgscale\undefined%
      \relax%
    \else%
      \setlength{\unitlength}{\unitlength * \real{\svgscale}}%
    \fi%
  \else%
    \setlength{\unitlength}{\svgwidth}%
  \fi%
  \global\let\svgwidth\undefined%
  \global\let\svgscale\undefined%
  \makeatother%
  \begin{picture}(1,0.28683998)%
    \lineheight{1}%
    \setlength\tabcolsep{0pt}%
    \put(0,0){\includegraphics[width=\unitlength,page=1]{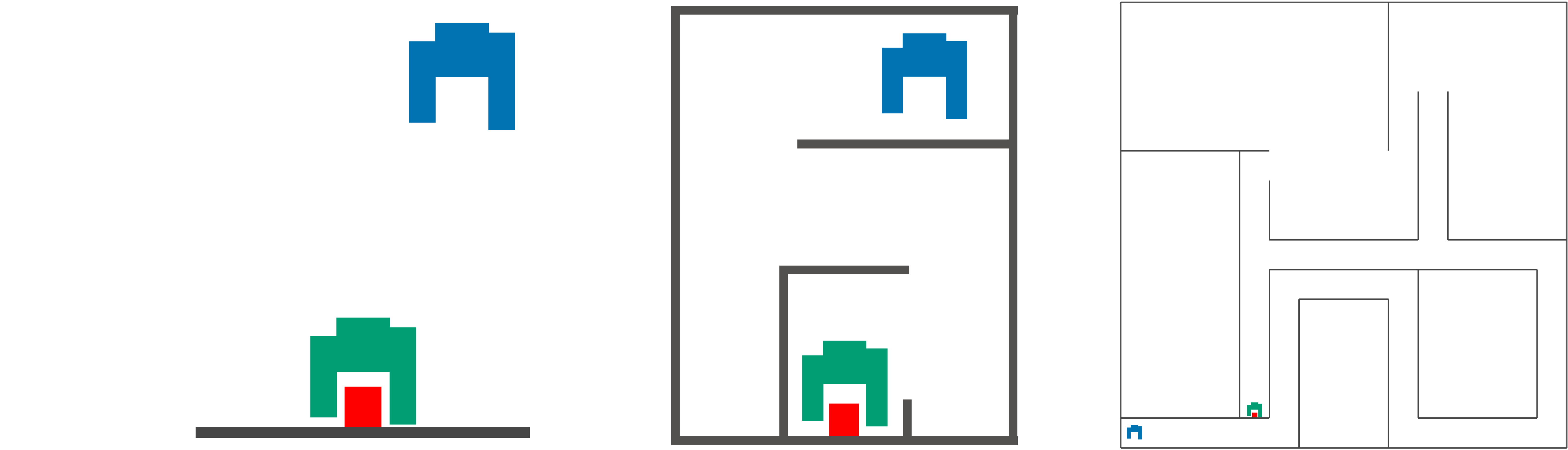}}%
    \put(0.28145136,0.25297131){\makebox(0,0)[lt]{\lineheight{1.25}\smash{\begin{tabular}[t]{l}$\mathbf{q}_0$\end{tabular}}}}%
    \put(0.2176913,0.06451086){\makebox(0,0)[lt]{\lineheight{1.25}\smash{\begin{tabular}[t]{l}$\mathbf{q}_g$\end{tabular}}}}%
    \put(0.58000747,0.24768183){\makebox(0,0)[lt]{\lineheight{1.25}\smash{\begin{tabular}[t]{l}$\mathbf{q}_0$\end{tabular}}}}%
    \put(0.52581846,0.05352636){\makebox(0,0)[lt]{\lineheight{1.25}\smash{\begin{tabular}[t]{l}$\mathbf{q}_g$\end{tabular}}}}%
    \put(0.79227517,0.03847402){\color[rgb]{0,0,0}\makebox(0,0)[lt]{\lineheight{1.25}\smash{\begin{tabular}[t]{l}$\mathbf{q}_g$\end{tabular}}}}%
    \put(0.73188387,0.00678149){\color[rgb]{0,0,0}\makebox(0,0)[lt]{\lineheight{1.25}\smash{\begin{tabular}[t]{l}$\mathbf{q}_0$\end{tabular}}}}%
  \end{picture}%
\endgroup%

~\\
\centering

\includegraphics[width=1\linewidth]{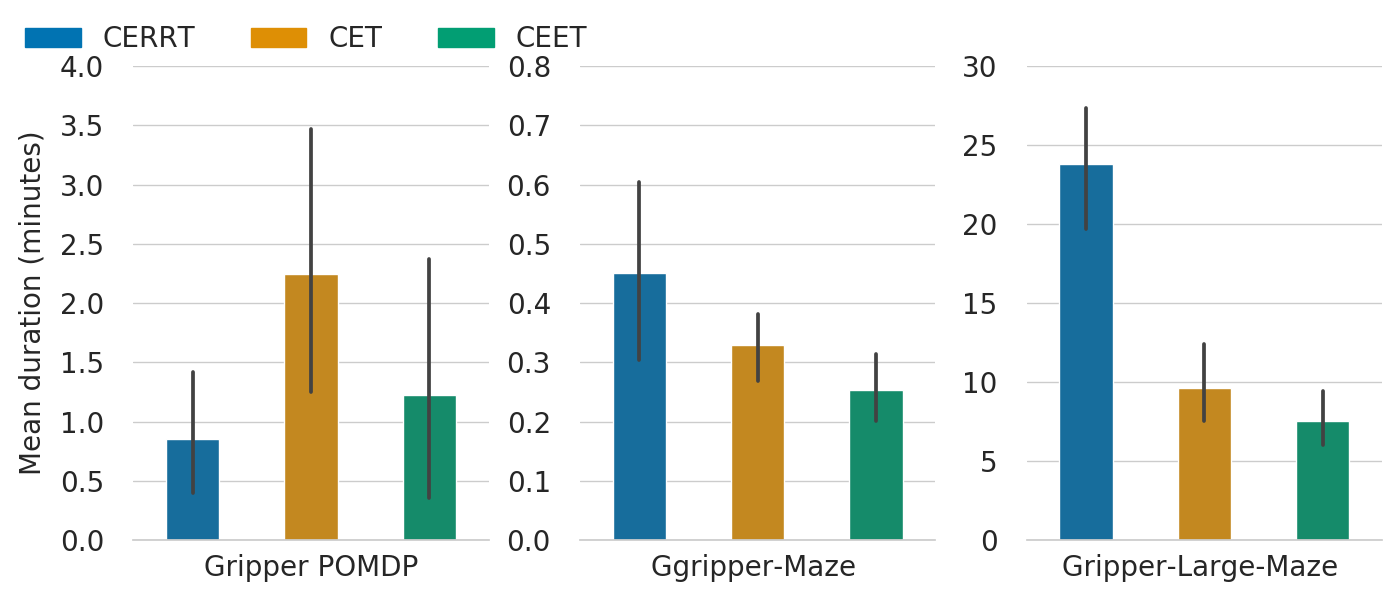}
% \vspace{.5em}
    \caption{Planning efficiency improves significantly when workspace information is used in complex environments. \textit{Top row:} three planning problems where a 2-DOF gripper has to move from $\vec{q}_0$ to $\vec{q}_g$ under initial position and motion uncertainty. The problem complexity increases from left to right by increasing the $\mathcal{C}$-space volume and the complexity of the environment. \textit{Bottom row:} comparing mean planning duration of three conformant planners with increasing use of workspace information: CERRT randomly explores $\mathcal{C}_\text{valid}$, CET randomly explores $\widetilde{\mathcal{C}}_\text{task} \subset \mathcal{C}_\text{valid}$, CEET guides exploration $\widetilde{\mathcal{C}}_\text{task} \cup \widetilde{\delta\mathcal{C}}_\text{task}$ and action selection.}
    \label{fig:mean_duration cspace eval}
 %    \vspace{-2.5em}
\end{figure}
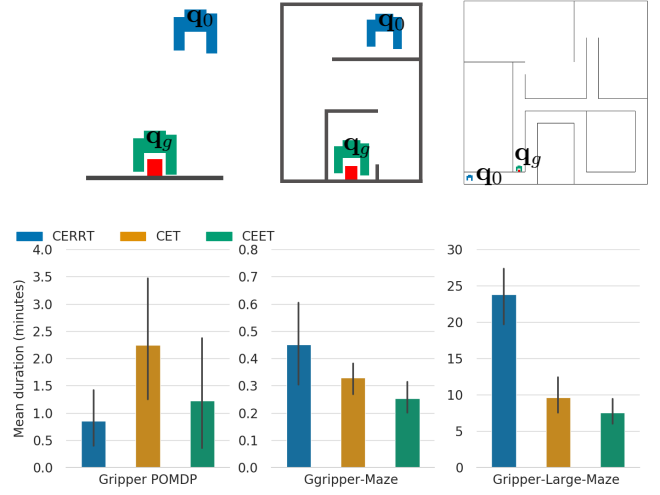

We want to show that workspace information-guided funnel sequencing improves a planner's computational efficiency in complex environments. Therefore, we compared the planning duration of CERRT, CET, and CEET  planners on three 2D planning problems with increasing $\mathcal{C}$-space volumes and increasing number of surfaces. 

In all three planning problems (top row in Figure~\ref{fig:mean_duration cspace eval}), a 2-DOF gripper had to move from $\vec{q}_0$ to $\vec{q}_g$ using only translational motion under initial position and motion uncertainty. The first planning problem is the \textit{Gripper POMDP}~\citep{hsiao_grasping_2007}, and the other two problems, \textit{Gripper-Maze} and \textit{Gripper-Large-Maze}, are variations with increasing complexity. First, we increased the volume of less relevant regions from 0\% to ~13\% and ~81\%. Secondly, we increased the environment's geometrical complexity by increasing the number of less relevant surfaces and the total number of surfaces from 0/5 to 3/21 and 32/51. We expect that leveraging structural context decreases planning duration as the environment's complexity increases.

The bottom row in Figure~\ref{fig:mean_duration cspace eval} shows the mean planning duration and the 95\% confidence interval. The results in the two maze problems show that $\mathcal{C}$-space reduction significantly improved planning efficiency, but action space reduction only provided minor improvement. However, the grasping POMDP problem results show that random exploration was better when the whole workspace was task-relevant. This is because the other three planners approximated $\widetilde{\mathcal{C}}_\text{task}$ too restrictively so that the reduced search spaces contained fewer solutions, making planning more difficult.

The planners found a solution for all three problems under the given time budget, except the CERRT planner for the large maze. In the large maze, CERRT failed two times out of ten because it spent too much time exploring the two large openings at the top of the large maze. Even though tough CEET was not restricted to searching less relevant regions when escaping a local minimum, the planner surprisingly found a solution the fastest. 

%-----------------------------------------------------------------
\subsection{Manipulation Funnels and Contact Sensing Reduce Increased Uncertainty} 
%-----------------------------------------------------------------

\begin{figure}[tbp]
  \centering
%  \def\svgwidth{0.9\linewidth}
  %    \vspace{-0.5em}
%\input{Part2/img/cs/ceet/results/all6planners_wamWALL.pdf_tex}
\includegraphics[width=1\linewidth]{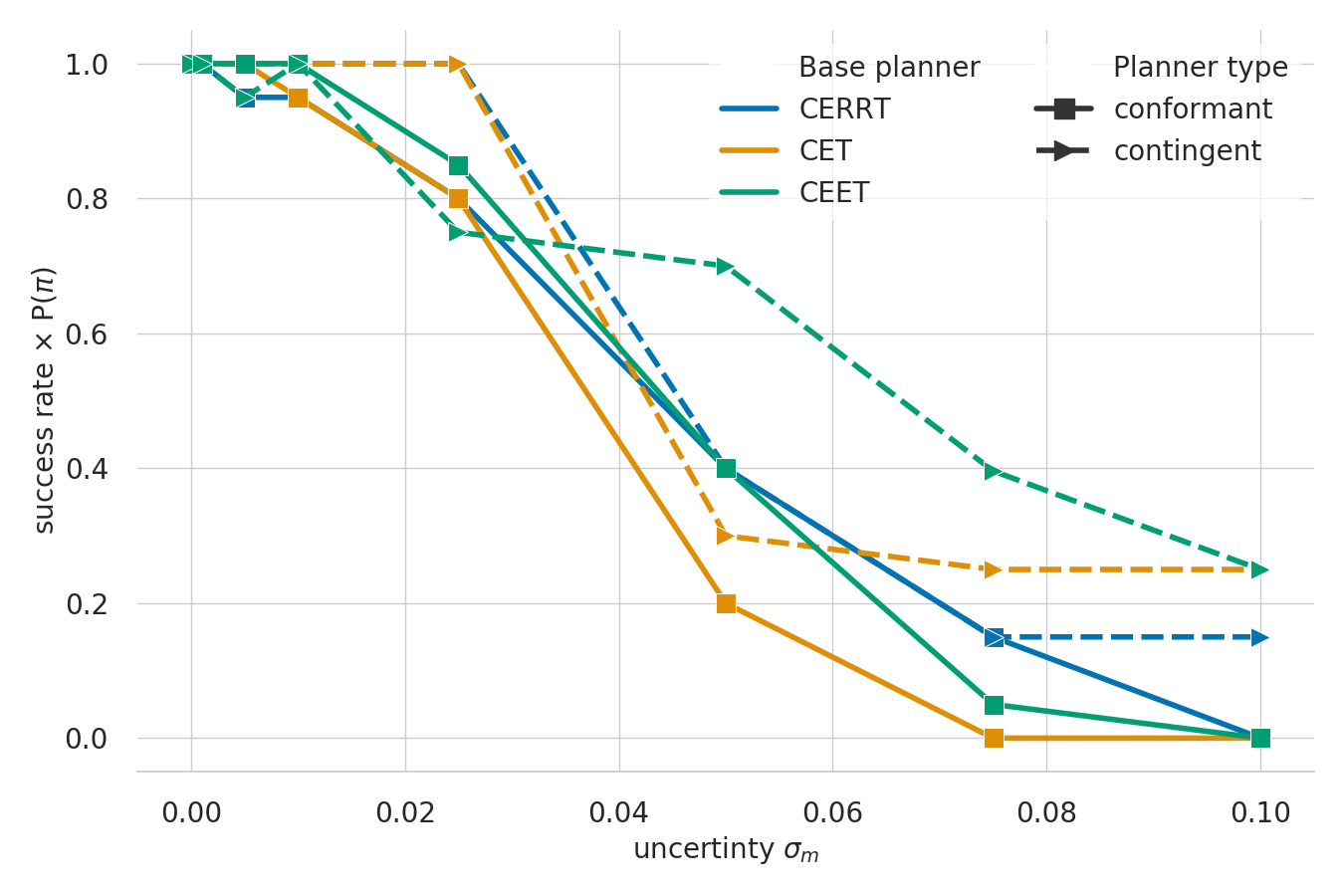}
    \caption{Planning success comparison of belief-space planners on the 7DOF planning problem. Contingent planners, using contact events and manipulation funnels, could handle the highest motion uncertainty, where the conformant planners, using only funnels, failed.} 
%{\crconcerrt} I have extended to it is not the same plot
    \label{fig:7d success vs uncertainty}
    \vspace{-1.em}
\end{figure}

We want to show that leveraging contact events for contingency planning improves planning success under increased uncertainty. Hence, we compared planning success $P_\text{scussess}$ for all conformant and contingent planners in the 7DOF WAM motion planning problem under increasing motion uncertainty $\sigma \in \{0,\:0.005,\:0.01,\:0.025,\:0.05,\:0,\:0.075,\:0.1\}$. We expect that contingent planners can handle larger motion uncertainty than conformant planners.

Figure~\ref{fig:7d success vs uncertainty} shows the planning success of all six planners. All planners are affected by increased motion uncertainty. Only the contingent planners, ConCERRT, ConCET, and ConCEET, found a solution (even if it was partial) under the 50-minute time limit, while the conformant version of these planners failed to find a solution for $\sigma_m=0.1$. Notably, guided exploration also improved contingent planning under uncertainty, indicating that the two approaches are complementary and should be used simultaneously for complex environments under increased uncertainty.  

\subsection{Real-World Applications}
%------------------------------------------------------------------------------

\cite{sieverling_interleaving_2017} evaluated the CERRT planner extensively, including the influence of $\gamma$ and of the number of particles representing a belief. They have also shown that the CERRT planner scales to high-dimensional planning problems by applying to a 3D problem using a 7 DOF arm inspired by Phillips-Grafflin and Berenson~\citep{phillips-grafflin_planning_2020} but also showed its use for real-world motion planning. We include the latter experiment below.

\begin{figure}[t]
    \centering
%    \subfloat[2DOF robot\label{subfig-1: 2dof}]{%
%	\def\svgwidth{.47\linewidth}
%	\includegraphics[width=0.46\linewidth]{example-image-a} 
%    }
%	\subfloat[7DOF arm\label{subfig-1: 7dof}]{%
%	\def\svgwidth{.47\linewidth}
	\includegraphics[width=0.9\linewidth]{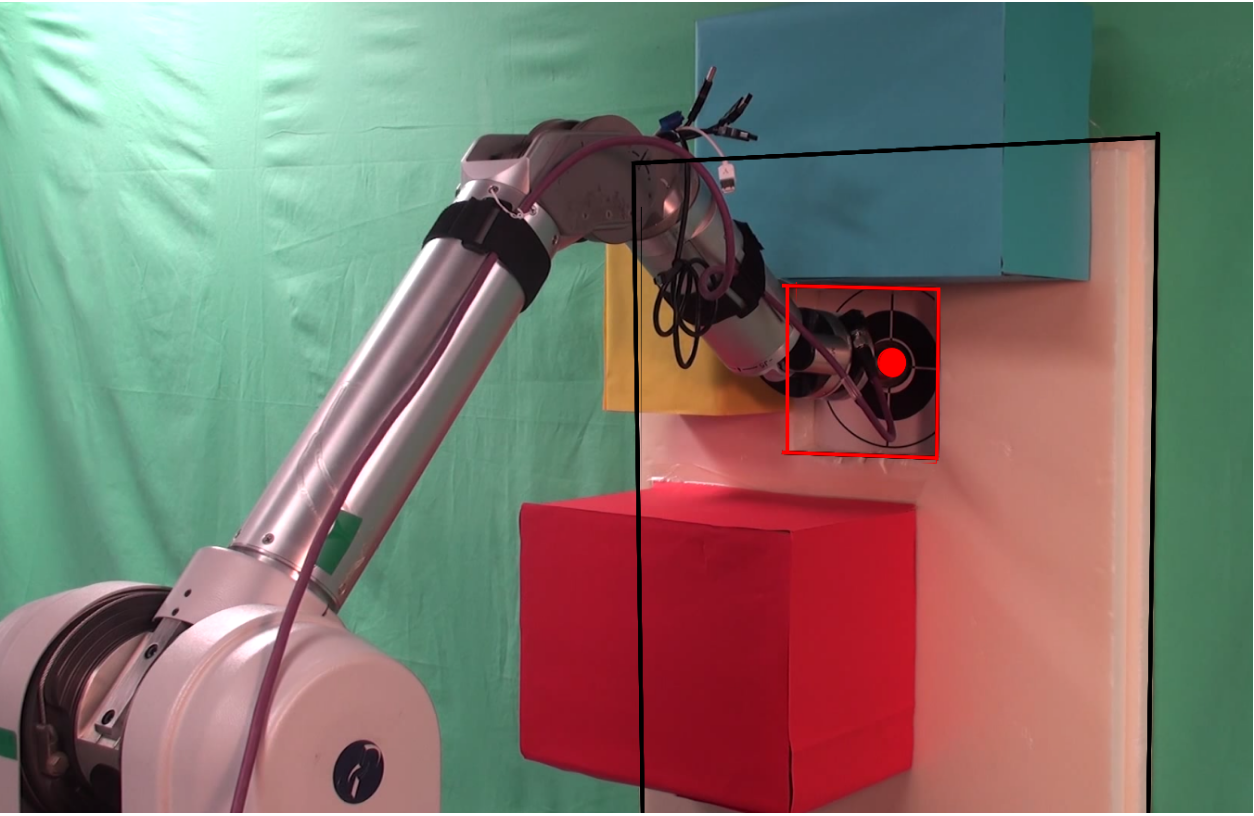} 
%    }
    \caption{Evaluation of the CERRT Planner by \cite{sieverling_interleaving_2017} with real-world experiments, where a robot used deliberate contact to reach a target under initial position uncertainty. \textcopyright 2017 IEEE.}
    \label{fig: cerrt experiment scenes}
\end{figure}

The CERRT planner was evaluated in a real-world motion planning application, proving that the planning approach is well suited for problems where the goal is relative to the environment. The task was to move the WAM 7DOF arm into a deepening on a wall as depicted in Figure~\ref{fig: cerrt experiment scenes}. Based on a given initial and motion uncertainty, a motion plan was computed with CERRT. Then, the wall was raised in front of the robot, and the previously calculated motion plan was executed without accounting for this change. Since the plan included ECE usage, the real robot could handle even the unexpected and un-modeled uncertainty from environmental change. The plan reduced the initial uncertainty of a 7 cm wall raise to a $\approx$2 cm deviation when reaching the end of its motion.

\citet{pall_contingent_2018} evaluated the ConCERRT planner extensively in simulation and in a real-world experiment as well. They also showed that the contingency branches of ConCERRT allowed them to solve problems with significantly higher uncertainty than the non-contingent CERRT planner but also showed that the ConCERRT policies are robust enough to be executed on a real robot for tactile localization of an object. We include the latter experiment below.

\begin{figure}[t]
  \centering
  \def\svgwidth{0.9\linewidth}
%   \vspace{0.0em}
%\input{Part2/img/cs/concerrt/realRobotExperiment.pdf_tex}
\includegraphics[width=0.8\linewidth]{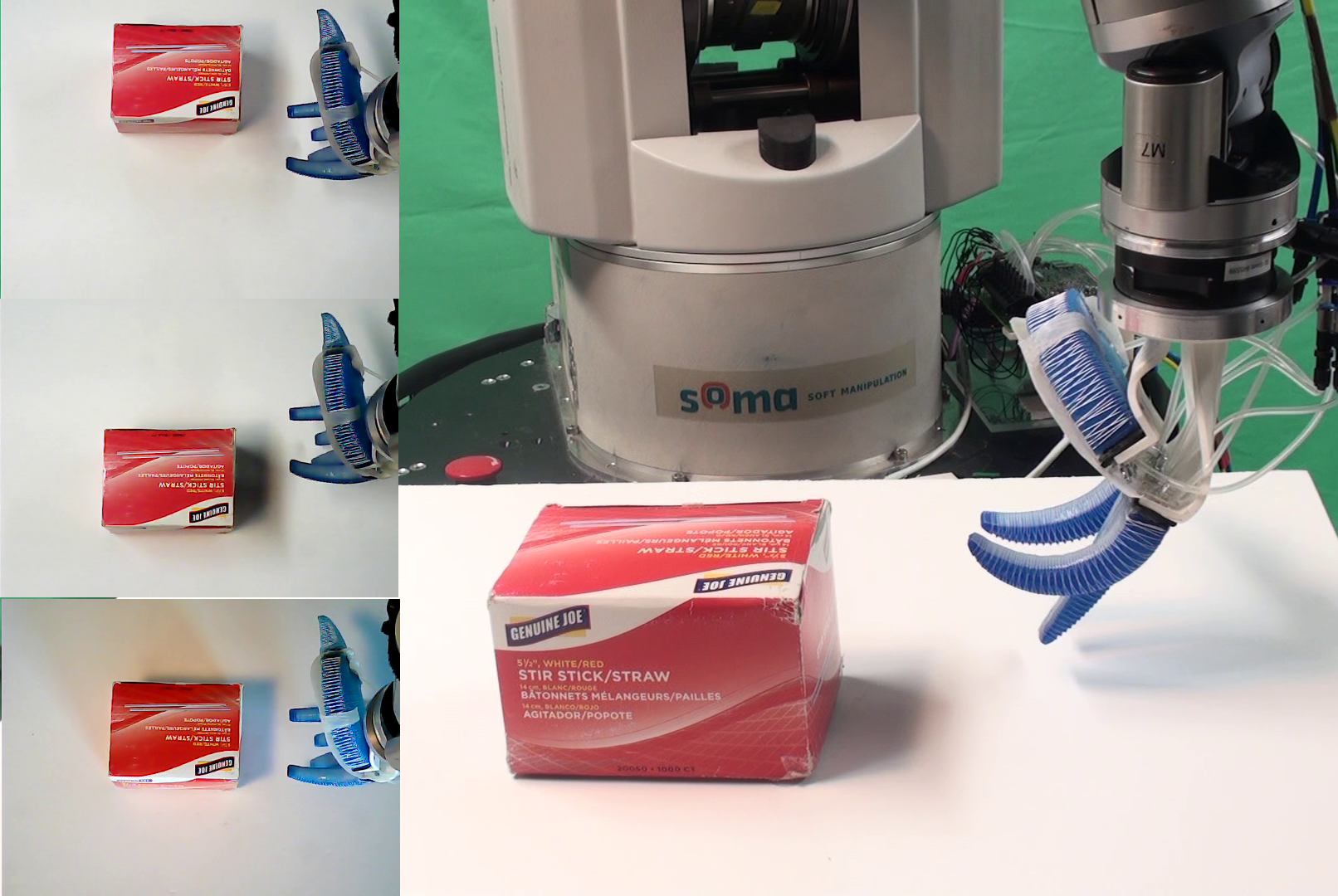}
    \caption{Real-robot object localization using pressure-based tactile sensing in the index and little fingers of the RBO~Hand~2. \textit{Left:} box positions with 0, +6, and -6 cm shifts relative to the hand. \textit{Right} the Barrett WAM ARM with the soft hand. \textcopyright 2018 IEEE}
    \label{fig:real robot experiment setup}
    %\vspace{-1.em}
\end{figure} 

To show the applicability of the ConCERRT planner, \citet{pall_contingent_2018} computed and executed a policy on a 7-DOF Barrett WAM robot arm with the RBO~Hand 2~\citep{deimel_novel_2016} as an end-effector. The experiment was inspired by the problem in Koval et al.~\citep{koval_pre-_2016}, where a robot arm with a contact-sensing hand localized an object on a tabletop. Similarly, the task was to sequence free-space and contact-exploiting motions such that the hand stops centered in front of the box as shown in Figure~\ref{fig:real robot experiment setup}. The fingers of a soft hand deform when they come into contact with the environment, resulting in a measurable pressure change. The robot used large changes in pressure as a proxy for contact sensing on the partially inflated index and little fingers as tactile sensors. The ConCERRT planner didn't use the slide action to plan a policy, as reliable sliding is hard to implement with a soft manipulator.

\begin{figure}[tbh]
  \centering
  \def\svgwidth{\linewidth}
   \vspace{0.em}
\includegraphics[width=\linewidth]{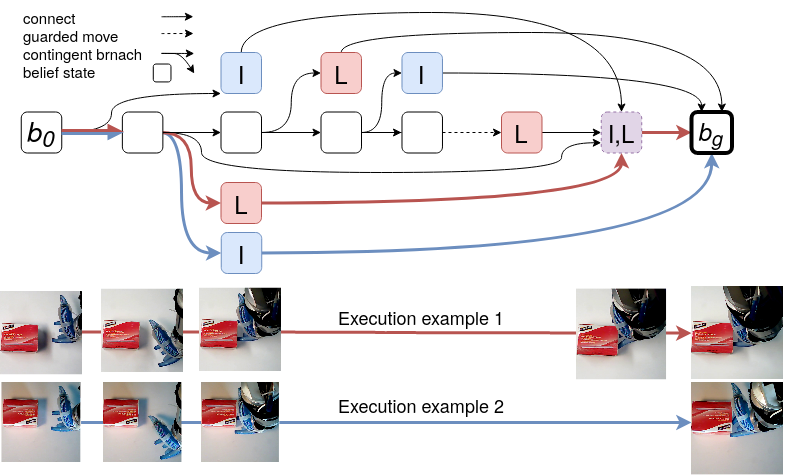}
    \caption{\textit{Top:} the policy used in the real-robot experiment, where belief states are labeled with "I", "L", and "I,L" indicating contact with the index, little, or both fingers, respectively. \textit{Bottom:} two executions showing the states when following the red and blue arrows in the policy. \textcopyright 2018 IEEE}
    \label{fig:real robot policy executed with 2 trajectories}
%    \vspace{-1.0em}
\end{figure}

ConCERRT consistently found policies in 16.66 minutes, and the selected one for execution is shown in Figure~\ref{fig:real robot policy executed with 2 trajectories}\footnote{\url{https://youtu.be/NaRppcg0CtQ}}. The most likely path through the policy makes four free-space motions and then ensures the final contact with a guarded move. Deviations to this path are anticipated with contact events on the index or little fingers. The policy was evaluated by moving the boxes 0, 2, 4, and 6 cm to the left and right relative to the hand's constant initial position and then executing the policy four times for each displacement. The policy was robust up to 4~cm uncertainty, and it could handle 6~cm to the right by reducing it $\approx$3~cm, but started to fail when the box was moved 6~cm to the left. 

The previous three sections showed that ECE is generally applicable for \Cs motion planning under uncertainty, and its benefits can be leveraged to overcome computational challenges for complex environments or under significant uncertainty. Next, we present a use-case study with a new EC that provides similar benefits but from complex contact dynamics.

%======================================================================
\section{A Use Case Study of EC-Based Grasping From Homogeneous Piles}
\label{sec: gece}
%======================================================================

This section presents a practical application of ECE in a real-world grasping problem. We use deliberate contact with the environment to simplify computation and achieve robust grasping by sequencing and executing ECEs control policies without traditional grasp or motion planning. Then, we explain how a new EC in pile dynamics enables open-loop grasping.

Traditionally, the grasping problem requires significant computation and accuracy: searching for a set of stable contact points\footnote{the fitness of contact point for grasping can be evaluated with form- or force closure analysis~\cite{}} on an object, touching the object at the desired points and clamping it with desired contact forces without making any additional contacts. In contrast, humans seemingly grasp effortlessly by simply reaching into a box and picking a piece of popcorn, even while watching a movie. Is it because they use environmental constraint exploitations?  

Our lab showed that humans use static parts of the environment as ECs~\citep{deimel_exploitation_2013} for robust grasping, and then we also proved that robots can use the same ECs for simple and robust grasping~\citep{eppner_exploitation_2015}. We refer to the use of such surfaces as \textit{geometrical ECs} since geometrical properties of the environment define the motion constraint. Later, we discovered a new EC in pile dynamics~\citep{pall_analysis_2021} that drastically simplified grasping from a pile of nearly identical objects. We refer to the use of pile dynamics as \textit{granular ECE} because solid objects in a pile provide a motion constraint similar to granular materials~\cite{}.

The novel EC provides benefits similar to geometrical ECs explained in previous sections but emerges from a more complex physical interaction. As before, the novel EC can simplify computation and ensure robust execution. By using it, our robot required no visual object detection or to compute contact points or interaction forces to achieve up to 100\% grasp success rate for round objects in a pile in our explorative study~\citep{pall_analysis_2021}.

\subsection{EC-Based Grasping Application}
\label{sec: pd gece}
%------------------------------------------------------------------------------

The following application is inspired by Ocado's real-world warehouse logistics~\citet{mnyusiwalla_bin_2020}. At Ocado, employees pack customer orders: first, an employee receives a grocery shopping list. Then, a bin arrives at their station containing one type of grocery item from the order. Next, the employee places the given number of items into a delivery bag, and the procedure repeats until the order is fulfilled. %We used their use case with minor technical modifications detailed below.

\begin{figure}[t]
\centering
%    \vspace{1.em}

		\includegraphics[width=0.9\linewidth]{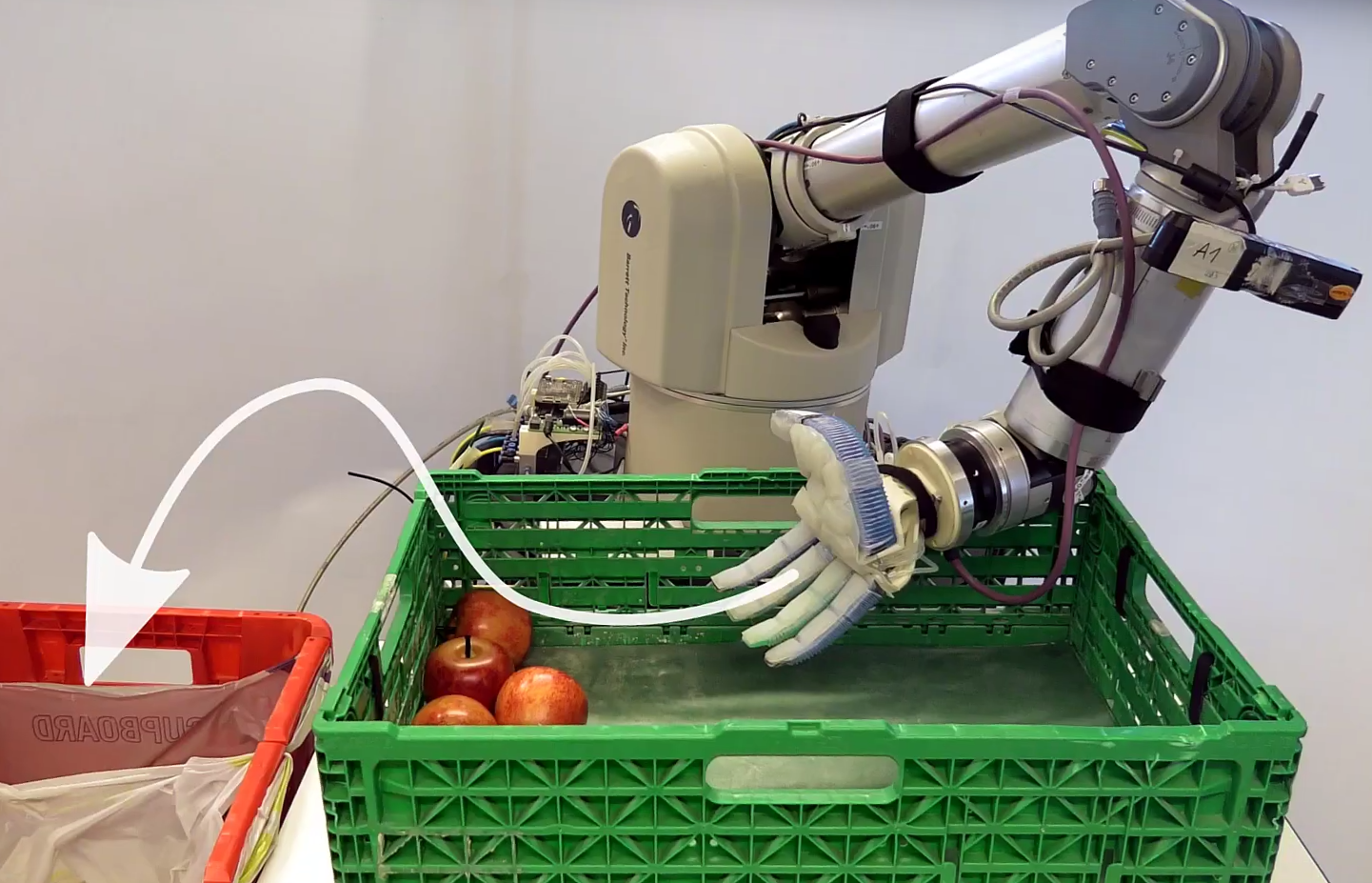}
    \caption{We use deliberate contact with the bin and pile of apples to simplify grasping in the Ocado's bin picking use cases~\citet{mnyusiwalla_bin_2020}. \textcopyright 2021 IEEE}
    \label{fig: ocado use case}
    \vspace{-1.5em}
\end{figure}

Like at Ocado, our robot had to fulfill an order by picking $N$ items from a bin and placing it into a delivery bag, as shown in Figure~\ref{fig: ocado use case}. Only one type of goods was stored in a bin, and the bin's location was known. Our robot was composed of a Barrett WAM 7DOF arm, a writs mounted ATI FTN-Gamma force-torque sensor, followed by the anthropomorphic RBO~Hand~2~\cite{deimel_novel_2016}.%, and for later experiments, we also use a rigid shovel.

In our adaption of the Ocado application, the items were apples, tennis balls, cylinders, or net bags of limes, and the order was of four items of the same type $N=4$. The key difference to Ocado's use case was that we tilted the bin $5\degree$, so the pile remained beside the same wall after each grasp attempt. This way, we can consider a pile's location known; thus, visual detection of the pile was not required. If the robot grasped more items than required, the robot dropped the grasped objects back into the pile. For example, the robot already picked and placed three apples out of four. If the next time it picks two apples in one attempt, it drops both apples back into the pile. Since we allowed the robot to grasp more than one item in one attempt and our hand can simultaneously grasp two to three objects, the ideal average grasp success $R$ is between two and four.

\begin{table}[t]
\centering
\setlength{\tabcolsep}{0.01em} % for the horizontal padding
{\renewcommand{\arraystretch}{1.2}% for the vertical padding
\begin{tabular}{c|c|c|c} 
%\hline
%\textbf{Object} &  
 \begin{minipage}{.24\linewidth}
      \includegraphics[width=\linewidth]{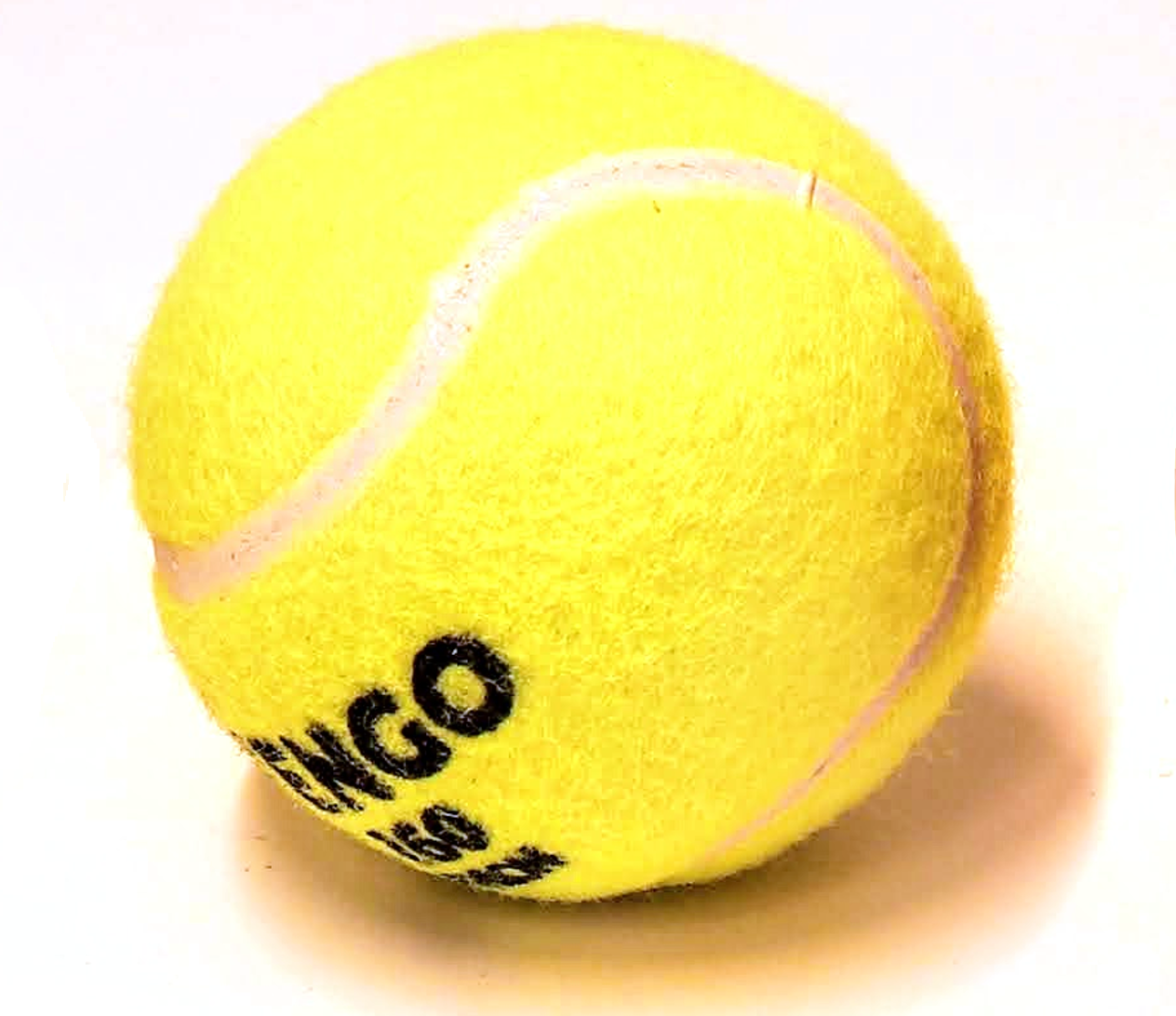}
 \end{minipage} &
 \begin{minipage}{.24\linewidth}
      \includegraphics[width=\linewidth]{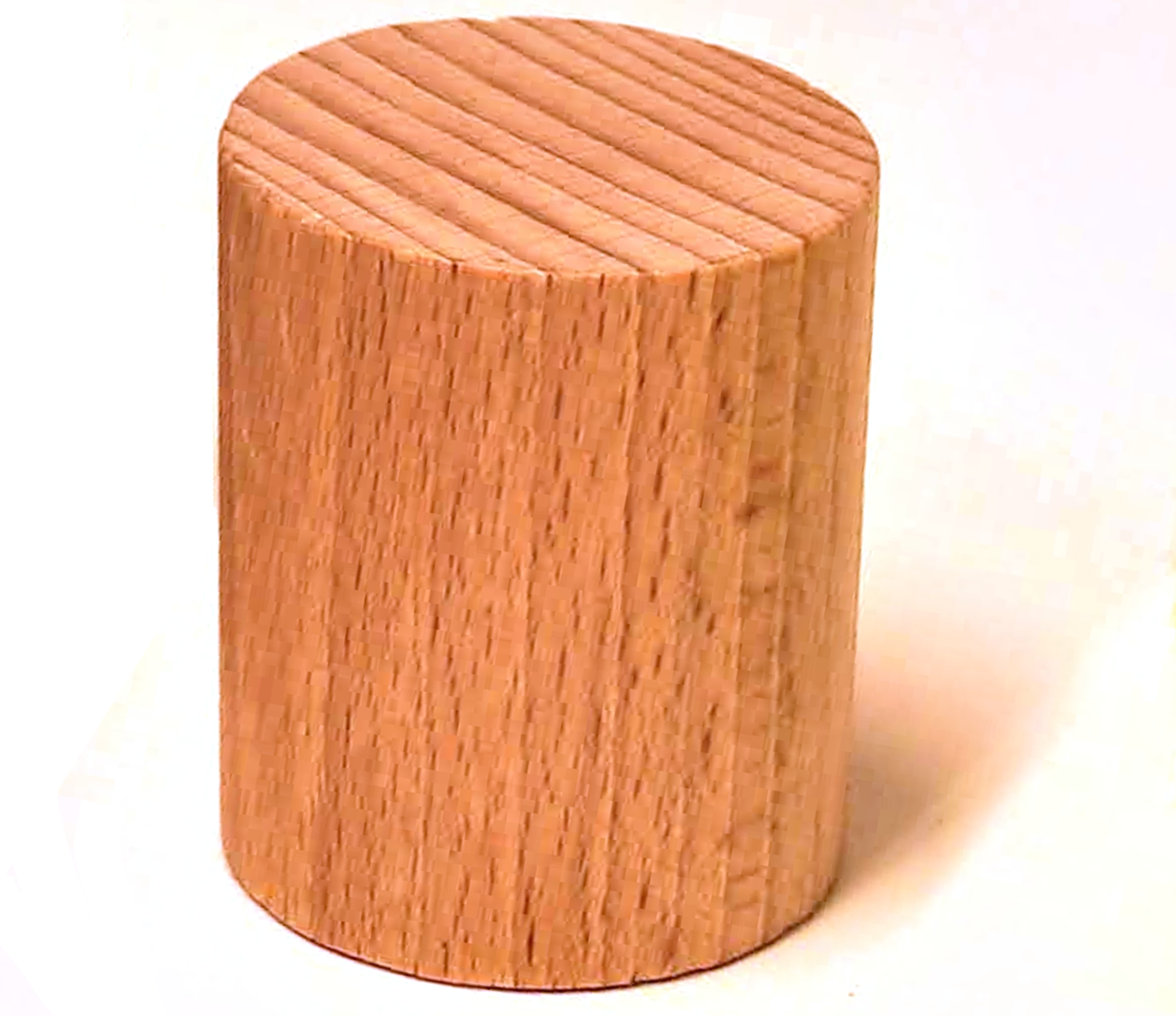}
 \end{minipage} &
% \begin{minipage}{.18\linewidth}
%      \includegraphics[width=\linewidth]{img/objectSet/cube.png}
% \end{minipage} &
  \begin{minipage}{.24\linewidth}
      \includegraphics[width=\linewidth]{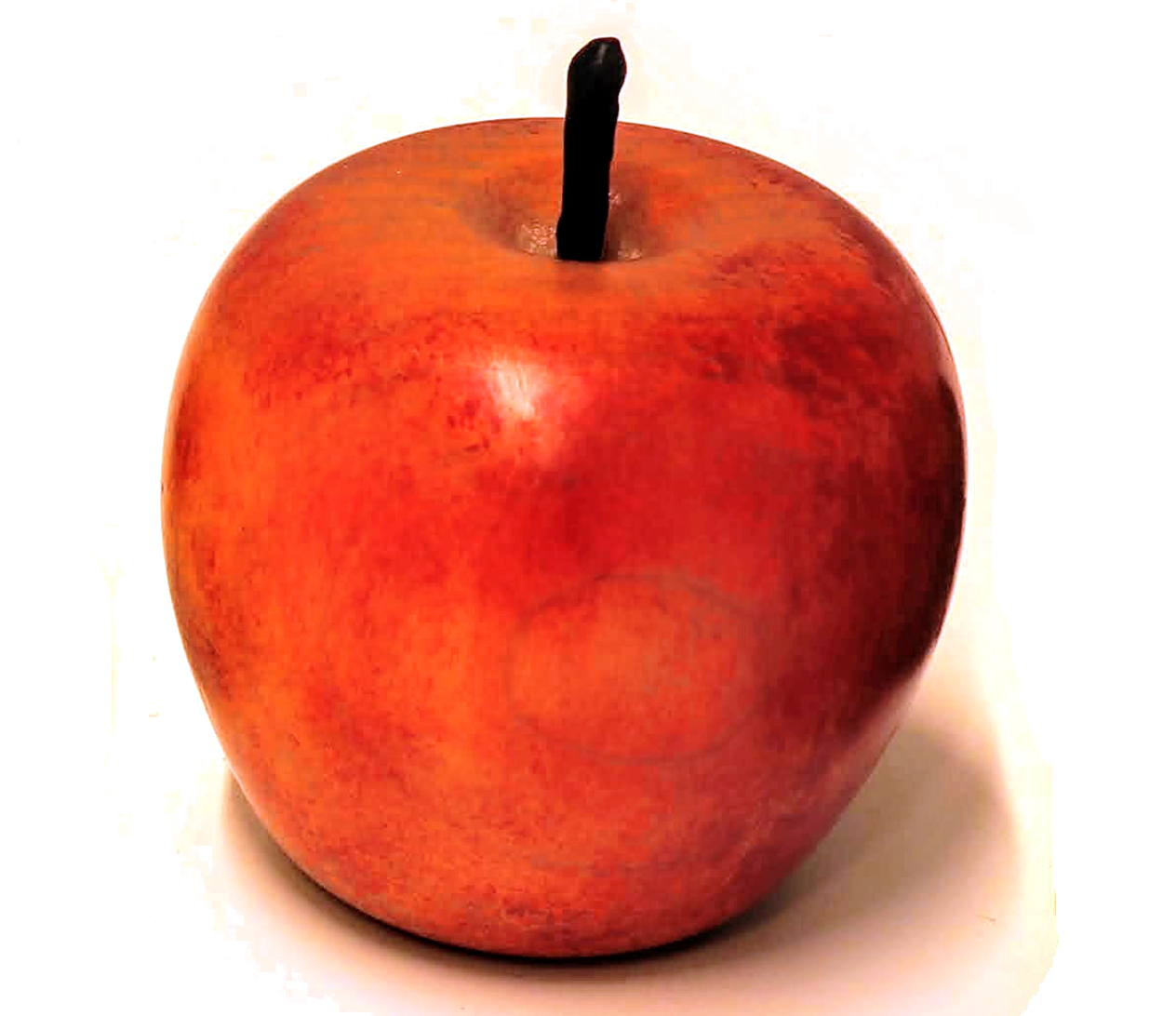}
 \end{minipage}&
 \begin{minipage}{.24\linewidth}
      \includegraphics[width=\linewidth]{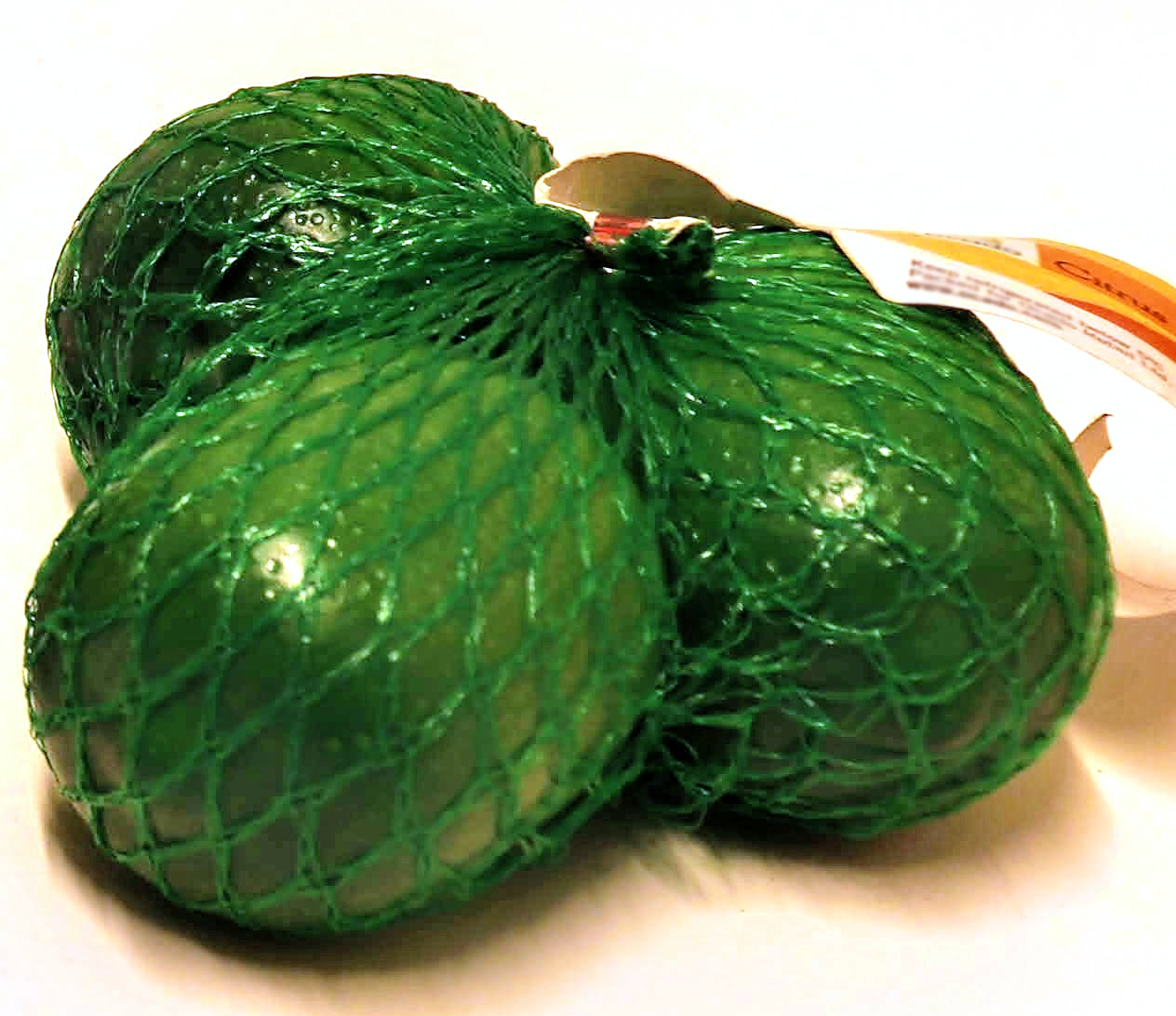}
 \end{minipage} \\   \hline
%\textbf{mass {[}g{]}} & 
tennis ball& cylinder&  apple& lime\\ \hline %cube&
60 or 180 g & 45 g&  168 g& 87 g\\ \hline % 65 g&
%\textbf{size {[}mm{]}}& 
$r=27$ mm&$r=20$ mm&$r=70$ mm&$r=49$ mm\\ %$l=45$ mm&
 & $h=50$ mm& $h=64$ mm& $h=68$ mm\\ \hline %
 $R=4.65$ & $R=9$ &$R=5.6$ &$R=7.6$ \\ % 4.3 and 5 for light and heavy balls

%\hline
\end{tabular}
}
\vspace{.5em}
\caption{Experimental object set with different object properties ($h$ is height and $r$ is radius) and results indicated by $R$ the average grasp attempts to pick and place four objects of the same type from the bin to the bag.}
\label{tab: obj properties}
\vspace{-1.5em}
\end{table}

\subsubsection{EC-Based Gasping Strategy\\}
%\label{sec: gece alg}
%------------------------------------------------------------------------------

%\todo{should include planning and the strategy too, but which order?}

\begin{figure}[t]
\vspace{0.5em}
	\centering
%\begin{subfigure}{1.\linewidth}	
%	\def\svgwidth{\linewidth}
%	\input{Part1/img/grasp_sequemce_cartoon.pdf_tex}	
\includegraphics[width=\linewidth]{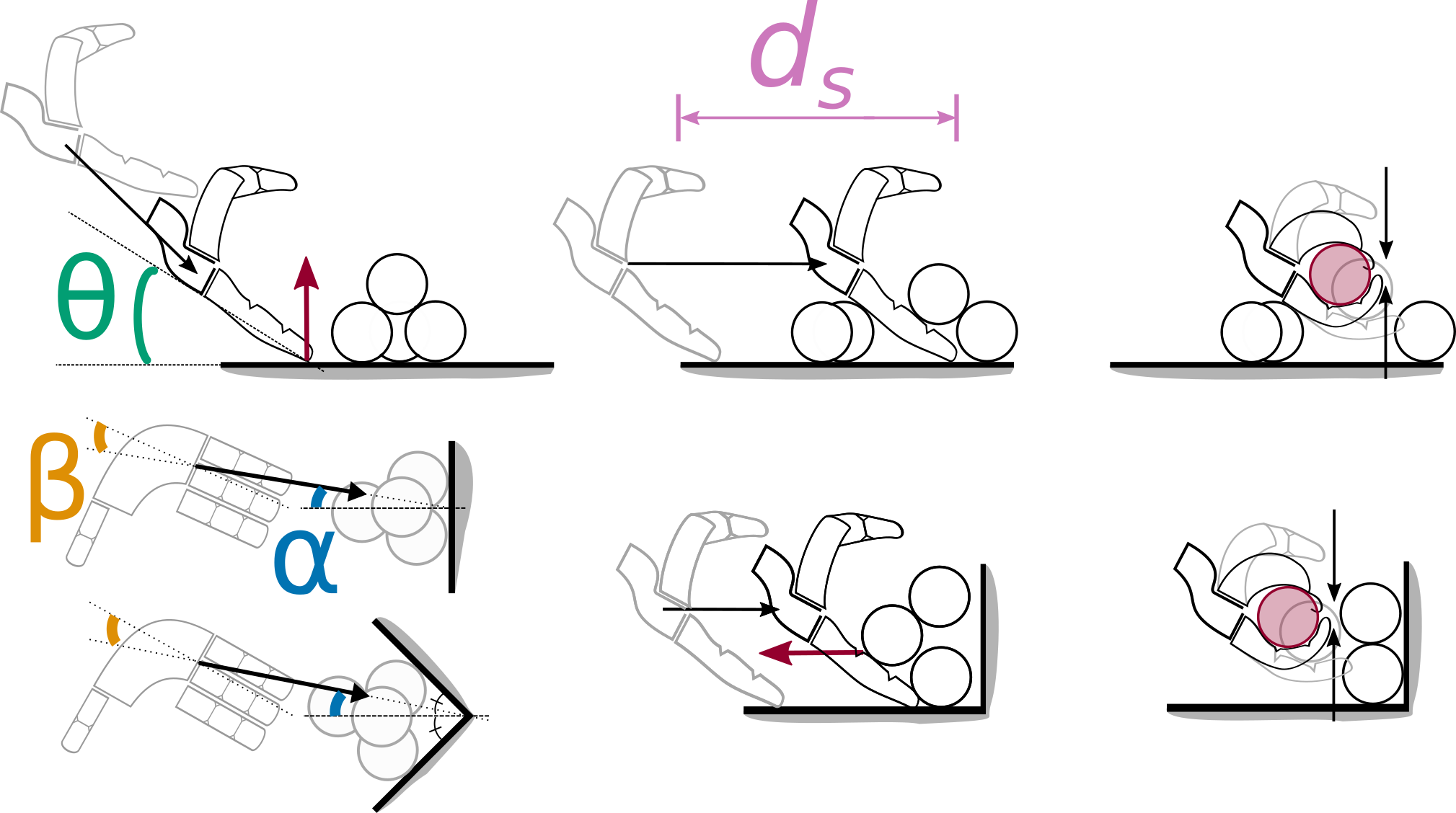}
%	\def\svgwidth{0.3\linewidth}
%	\input{img/grasp_sequemce_cartoon_1.pdf_tex}	
%	\def\svgwidth{0.3\linewidth}
%	\input{img/grasp_sequemce_cartoon_2.pdf_tex}	
%	\def\svgwidth{0.3\linewidth}
%	\input{img/grasp_sequemce_cartoon_3.pdf_tex}	
%%\end{subfigure}
	\caption{2D sketch of the grasp strategy for a pile alone (top row) or supported by vertical walls (bottom row): a hand \textit{lowert} to the tabletop from free space, where $\theta$ (green) is the finger's slope, $\alpha$ (blue) and $\beta$ (yellow) are the relative hand orientation to the pile and a supporting wall or corner, and the hand's motion is visualized with black arrows. The force along the contact normal (vertical red arrow) triggers the \textit{sliding} motion, which terminates after moving $d_s$ distance (pink) or if a contact force (horizontal red arrow) is detected. Finally, the hand \textit{grasps} an object that rolled on it.}
		\label{fig: grasp strategy sketch}
	\vspace{-1.em}
\end{figure}

In Ocado's use case, we apply our simple shoveling-like grasp strategy from~\citep{pall_analysis_2021}. It is composed of three EC exploitations, as illustrated in Figure~\ref{fig: grasp strategy sketch}. First, the hand \emph{lowers} to the tabletop with its palm facing up to align vertically with the pile. Second, the hand \emph{slides} on the horizontal support surface to maintain its vertical position. Third, the hand is pushed into the pile to roll an object onto its palm. Finally, \emph{grasping} triggers after the hand moves through the pile or reaches a static wall, and then, the hand stops moving and closes its fingers. We omitted to control any particular object's position and the applied forces because the granular EC will " control " these.

We parameterized the strategy with the hand's orientation ($\alpha=0\degree$, $\beta=0\degree$, and $\theta=30\degree$), velocity $|\vec{v}_\text{hand}|=0.1 m/s$, sliding distance $d_s$, and a force threshold $F_\text{grasp}=17N$. The hand's orientation is relative to a pile and the environment, where $\alpha$ is the relative hand orientation to the pile. When the pile is next to a wall, $\alpha$ is the angle between the hand and the wall normal. With corners, $\alpha$ is the angle between the hand and the interior bisector of the corner angle. $\beta$ is the offset angle between the fingers and the direction of motion, and $\theta$ is the slope of the fingers.

\subsubsection{Grasping Strategy Evaluation\\}

We executed an order fulfillment request per object type four times and recorded the average number of pick attempts to place four objects into the bag. Before each grasp attempt, an operator decided to leverage the granular EC with a wall or corner and executed the pile-centered strategy. The results are shown in Table~\ref{tab: obj properties} and in this \href{https://youtu.be/BtCBNdkgejU}{\color{blue}{video}\color{black}}\footnote{https://youtu.be/BtCBNdkgejU}. 

On average, the robot needed 6.3 grasp attempts to pick and place four items, indicating that our simple grasping strategy suited bin-picking. It also showed that such open-loop grasp (no feedback used on object position or applied forces) generalizes for irregular object shapes like apples and even net bags of limes. The success of this simple strategy can only be explained by the existence of an EC. It also explains why humans grasp effortlessly from piles of objects, like popcorn or nuts from a container.

\subsection{Empirical Study of the Granular EC}

Our simple grasp strategy's success can be explained by geometrical and granular ECE. While we understand geometrical ECE, granular ECE is less understood. Thus, we propose new explanations of the novel granular EC's emergence and interaction with geometric ECs. Our physics-based explanation provides a deeper understanding of why granular ECE, combined with geometrical ECE, allows grasping by simply sequencing and executing control policies without computing a motion plan, detecting individual objects, or controlling contact between contact points or contact forces. 

We argue that the granular ECs provide similar benefits in that above use case to geometrical ECs for motion planning (as detailed in previous sections) and for single object grasping shown by \citet{eppner_exploitation_2015}: 
\begin{enumerate}
\item granular ECE provides a manipulation funnel to reduce uncertainty, specifically, reducing hand-object position uncertainty for grasping.
\item it provides detectable contact events to indicate the funnel's exit, allowing funnel sequencing, and
\item it provides structural information about the environment and grasp affordances with geometrical features of the environment.
\end{enumerate}
\noindent{}We prove the existence of granular EC's benefits using abductive reasoning: First, we describe an observed motion constraint in pile dynamics when a hand pushes into a pile. Then, we propose an explanation for this constraint. Based on our explanation, we derive multiple testable hypotheses to show the existence of a manipulation funnel and validate our hypotheses with real-world and simulated experiments. 

\begin{figure*}[tb]
	\centering	
	\def\svgwidth{0.95\linewidth}
	%% Creator: Inkscape inkscape 0.92.3, www.inkscape.org
%% PDF/EPS/PS + LaTeX output extension by Johan Engelen, 2010
%% Accompanies image file 'gEC_vectorField.pdf' (pdf, eps, ps)
%%
%% To include the image in your LaTeX document, write
%%   \input{<filename>.pdf_tex}
%%  instead of
%%   \includegraphics{<filename>.pdf}
%% To scale the image, write
%%   \def\svgwidth{<desired width>}
%%   \input{<filename>.pdf_tex}
%%  instead of
%%   \includegraphics[width=<desired width>]{<filename>.pdf}
%%
%% Images with a different path to the parent latex file can
%% be accessed with the `import' package (which may need to be
%% installed) using
%%   \usepackage{import}
%% in the preamble, and then including the image with
%%   \import{<path to file>}{<filename>.pdf_tex}
%% Alternatively, one can specify
%%   \graphicspath{{<path to file>/}}
%% 
%% For more information, please see info/svg-inkscape on CTAN:
%%   http://tug.ctan.org/tex-archive/info/svg-inkscape
%%
\begingroup%
  \makeatletter%
  \providecommand\color[2][]{%
    \errmessage{(Inkscape) Color is used for the text in Inkscape, but the package 'color.sty' is not loaded}%
    \renewcommand\color[2][]{}%
  }%
  \providecommand\transparent[1]{%
    \errmessage{(Inkscape) Transparency is used (non-zero) for the text in Inkscape, but the package 'transparent.sty' is not loaded}%
    \renewcommand\transparent[1]{}%
  }%
  \providecommand\rotatebox[2]{#2}%
  \newcommand*\fsize{\dimexpr\f@size pt\relax}%
  \newcommand*\lineheight[1]{\fontsize{\fsize}{#1\fsize}\selectfont}%
  \ifx\svgwidth\undefined%
    \setlength{\unitlength}{588.9235465bp}%
    \ifx\svgscale\undefined%
      \relax%
    \else%
      \setlength{\unitlength}{\unitlength * \real{\svgscale}}%
    \fi%
  \else%
    \setlength{\unitlength}{\svgwidth}%
  \fi%
  \global\let\svgwidth\undefined%
  \global\let\svgscale\undefined%
  \makeatother%
  \begin{picture}(1,0.46183983)%
    \lineheight{1}%
    \setlength\tabcolsep{0pt}%
    \put(0,0){\includegraphics[width=\unitlength,page=1]{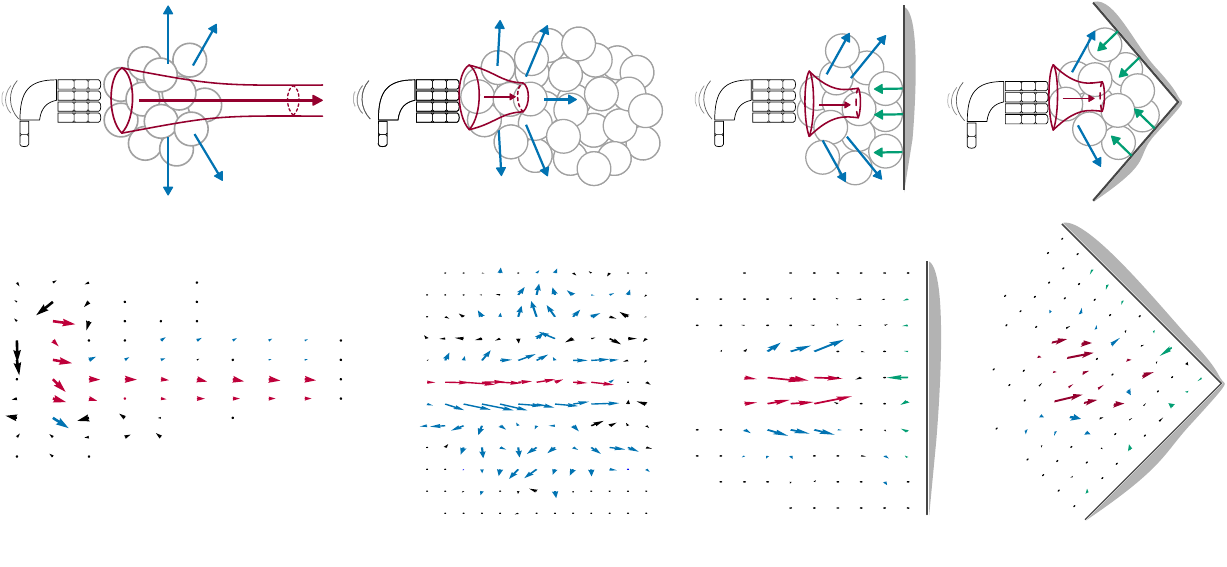}}%
    \put(0.39500537,0.28633922){\color[rgb]{0,0,0}\makebox(0,0)[lt]{\lineheight{1.25}\smash{\begin{tabular}[t]{l}a) sketch of force patterns \\\end{tabular}}}}%
    \put(0.39508469,-0.00975214){\color[rgb]{0,0,0}\makebox(0,0)[lt]{\lineheight{1.25}\smash{\begin{tabular}[t]{l}b) force patterns in simulation \\\end{tabular}}}}%
%    \put(0,0){\includegraphics[width=\unitlength,page=2]{gEC_vectorField.pdf}}%
  \end{picture}%
\endgroup%
	
	\caption[Illustrated and simulated force patterns in piles of objects]{All images show an interaction pattern in piles when pushed by an end-effector. \textit{Top row:} The four sketches illustrate the object stabilization and rolling into a hand with red funnels. The red arrows show that objects inside the funnel move with the hand. The blue arrows depict the radial spread of the other objects outside the funnel. The green arrows show the contact normals on static walls in the direction of pushing. Interaction forces roll an object on the hand for large enough piles or piles supported by walls, and the pile expands less. \textit{Bottom row:} we show a vector field depicting the interaction forces averaged over the duration of multiple simulations. We manually colored some of the arrows indicating similar patterns to the respective sketch. Even though the interactions are complex, a similar pattern to that depicted in the sketches can also be observed in the simulation.}	
	\label{fig: pile behaviour}
	\vspace{.em}
\end{figure*}

%-----------------
\subsubsection{Observed Motion Patterns in Piles\\} 
%-----------------

We observed an interaction regularity when a hand on a tabletop slides into a pile. Some objects in front of the hand moved together with the hand, while the other objects spread radially. With larger piles, one or multiple objects are rolled into the hand, depending on the hand width and object size. With smaller piles, an object rolled into the hand when the environment constrained the pile's expansion. 

We depict the observed regularity as a funnel in the top row of Figure~\ref{fig: pile behaviour}. The funnel entrance is a region where objects start moving together with the hand, the walls represent the shrinking of this region. The stabilization region's shrinking symbolizes that the number of stabilized objects decreases, sometimes it decreases to a single object that becomes centered on the hand. Finally, the funnel's exit is the point where an object rolls into the hand. Note that the funnel has no ending on the first sketch indicating that objects move together with the hand and no object rolls into the hand for small piles.

We observed a similar regularity in simulated pile dynamics. In simulation, we used a shovel-like end-effector and pushed it into a pile of spheres until it passed through the pile or an object rolled onto it. The results are visualized with four vector fields of interaction forces inside piles at the bottom of Figure~\ref{fig: pile behaviour}. For each cell in the field, we averaged the forces acting on objects in that cell over the whole duration of a simulation and repeated averaging for 100 simulations for each of the four cases. Then, we manually colored arrows that indicate object stabilization (red), radial expansion of other objects (blue), and objects' interaction with static vertical walls (green). The black arrows cannot be associated with any of the three cases mentioned before, indicating the complexity of the interaction forces.

Based on our real-world and simulation-based observations, we propose three effects of granular ECE: a) objects tend to stabilize in front of the hand, b) an object becomes centered on the hand, and c) geometrical and granular ECs interact. 

Next, we propose a physics-based explanation for all three effects. Even though we can provide an explanation using physics, it is difficult to model pile dynamics accurately. Small changes in applied forces on an object produce very different object motions. Hence, we refrain from defining mathematical formulas to describe the EC but use our explanations below to find relevant properties of piles, objects, and the environment that affect grasping with granular ECE.  

%------------------------------------------------------------
\subsubsection{Explanation of the Granular ECE Effects\\}
%\label{sec: physics explanation of gEC}
%------------------------------------------------------------

We provide a physics-based explanation for the three effects and derive testable hypotheses by assuming that objects are solid, so the respective pile can be considered a granular material~\citep{duran_sands_2012}, and use properties of granular materials in our reasoning. Moreover, we assume that the hand and object sizes allow grasping one object at a time with granular ECE. However, we think our explanations hold when the hand is wider or the objects are smaller: then, multiple objects stabilize, center, and roll into the hand.

%-----------------
\textbf{a) Object stabilization effect:} 
%-----------------
Granular materials possess the \emph{penetration resistance} property~\citep{stone_getting_2004}. Penetration resistance provides an opposing force on a pusher and, consequently, on a pushed object as well. This opposing force manifests because the objects' kinetic energy dissipates inside granular materials, similar to the effect of the friction force between two objects. Kinetic energy dissipates due to friction between objects and the objects' inertia. Objects move together with the pusher if the pusher's front part is concave or flat. 

Objects stabilize in front of a hand when pushed due to the penetration resistance. The concavities between the fingers also help this stabilization. When a hand pushes an object into a pile, the loss of kinetic energy generates an opposing force on the pushed object. If the opposing force is large enough, the pushed object rolls into the hand, after which the hand can close its fingers to grasp the object. 

We analyze the conditions for grasping when leveraging the object stabilization effect of granular ECs. Since kinetic energy loss affects the amount of opposing force, we want to characterize the relationship between the cardinality of a pile (number of objects in a pile) and grasp success. We propose the following hypothesis:\\
\textbf{Hypothesis 1:} As the number of objects in a homogeneous pile increases, granular ECE better supports grasping, assuming that the size and mass of the objects relative to the hand allow grasping one object at a time. 

%-----------------
\textbf{b) Object centering effect:} 
%-----------------
The second key property of granular materials concerns how an external force propagates through the material. An external pushing force propagates through short and temporal \emph{force chains}, transferring the load. Force chains are parallel or radial for structured- and unstructured granular materials, respectively~\citep{tordesillas_force_2014}. The force chains carry the opposing force, creating compression in front of the pusher. 

With unstructured piles, a hand applies forces at multiple locations. Thus, a compression region develops in front of the hand. In this region, objects are pushed toward the hand and roll away outside the region. Since the force chains are temporal, the region's width changes relative to the hand. Therefore, objects can shift laterally in front of the hand, and only a centered object remains stabilized. The changing nature of the compression region causes an object to center in front of the hand. To analyze the object-centering effect on grasping, we analyze the importance of hand alignment to an object.\\
\textbf{Hypothesis 2:} A robot can grasp successfully from a homogeneous pile even without initially aligning its hand with an object when leveraging the granular EC, assuming that the size and mass of the objects relative to the hand allow grasping one object at a time.

With structured piles, the object-centering effect does not emerge because the force chains are parallel and constant. Therefore, the compression region is also constant in front of the hand. If the structure breaks, the compression region changes, and the object-centering effect manifests. Since we are interested in the object-centering effect of granular ECs, our study focuses on unstructured piles.

%-----------------
\textbf{c) Geometrical ECs' affect granular ECE:} 
%\label{sec: physics explanation of gEC}
%-----------------
Static parts of the environment (i.e., geometrical ECs) increase the penetration resistance of a granular material~\citep{stone_getting_2004}. The penetration resistance increases near static obstacles because of the contact normals and friction forces between the environment and objects. 

When the environment constrains a smaller pile's expansion, the pile propagates the interaction forces between objects and the environment. We can leverage the interaction between objects and the environment for grasping from smaller piles.\\
\textbf{Hypothesis 3:} A robot can grasp with granular ECE from smaller homogeneous piles, which alone provides insufficient opposing force, if the pile is beside static walls assuming that the size and mass of the objects relative to the hand allow grasping one object at a time from large enough piles not supported by geometrical ECs.

The last two sketches in Figure~\ref{fig: pile behaviour} illustrate when a small pile is constrained by a wall and a corner. In both cases, the penetration resistance increases, and a corner further constrains a pile's expansion than a wall, which we illustrated by having fewer blue arrows on the leftmost sketch.

\subsection{Hypotheses Evaluation of Granular ECE Effects}
\label{sec: gece eval}
%------------------------------------------------------------------------------
We want to experimentally prove the existence and characterize the novel granular EC emerging from pile dynamics. Since pile dynamics is challenging to model or measure accurately, but force patterns manifest on a high level, we followed an empirical approach. Above, we proposed explanations for the observed regularities and derived a hypothesis. Now, we evaluate our hypotheses, showing that: 
\begin{enumerate}
\item granular EC provides an opposing force for grasping,% by evaluating Hypothesis~1, 
\item granular EC centers an object for grasping, and% by evaluating Hypothesis~2, and
\item granular EC's opposing force improves when combined with geometrical ECE.
\end{enumerate}

We analyze the granular EC  by observing the outcome of grasp attempts in real-robot and simulated experiments. We omit human grasp experiments to avoid biases like human reflexes. Humans can react uncontrollably to visual and tactile stimuli that affect pile dynamics. On the other hand, a robot's motion control can be open-loop by design. All experiments used the same parameters and followed the protocol described next unless stated otherwise.

In each experiment, first, we built a random pile to have a random hand-object alignment to sample a grasp success rate from twenty executions. Since objects' configuration in a pile can affect the pile dynamics, we built unstructured piles with random object configurations by dropping the objects into a cube-like frame. The frame was placed at a predefined location on a table. When vertical walls constrained a pile, the frame was placed next to a wall or corner. The frame size linearly increased with the pile's cardinality $|\text{Pile}|$ as follows: $0.77\times|\text{Pile}|+11.15$. We chose this frame size to build piles with a minimum of two layers of objects. We removed the frame, waited until the pile stabilized, and then executed the strategy. 

Table~\ref{tab: obj properties} gives the actual object's properties. All simulated objects had the same friction properties using the tennis ball's properties~\citep{cross_measurements_2002} for comparability between simulated and real-world experiments. The restitution coefficient (COR) of all simulated objects was a linear function of the mass, where COR(60g)$\:= 0.72$~\citep{cross_measurements_2002}, and COR(180g)$\:= 0.37$ using our experimental observation on our sand-filled heavy tennis balls.

We executed our grasp strategy using the same hardware setup as for the use case application but sometimes with a simple and rigid shovel-like end-effector. We used only the shovel-like end-effector in a simulation because the accurate simulation of the soft hand is challenging, and we expected to observe the granular EC even with this simple shovel. The simulated end-effector had lateral friction~0.5, spinning\nobreakdash-, rolling-friction~0.001, and COR$\:=\:0.8$. We parameterized the strategy with $\alpha=\beta=0\degree$, $\theta=15\degree$, $|\mathbf{v}_\text{hand}| = 0.1 m/s$, $F_\text{grasp} = 17 N$, and the sliding motion terminated after 30 cm. 

Finally, we observed the \emph{Roll Success} (RS) and \emph{Grasp Success} (GS) in each experiment. We observed the roll and grasp successes because measuring interaction forces inside a pile is difficult, and force-torque measurements are noisy. \emph{Roll Success} was true if an object rolled on the hand such that the object's center of mass was above the fingers. \emph{Grasp Success} was true if an object was grasped after the fingers closed. We sampled these success rates from twenty observations on the robot and fifty in the simulation.

%-------------------------------------------------
\subsubsection{Larger Piles Roll Objects on the Hand for Grasping\\}
\label{sec:eval p1 valing}

\begin{figure}[t]
\centering
	\fontsize{9pt}{11pt}\selectfont
	\def\svgwidth{1.\linewidth}
	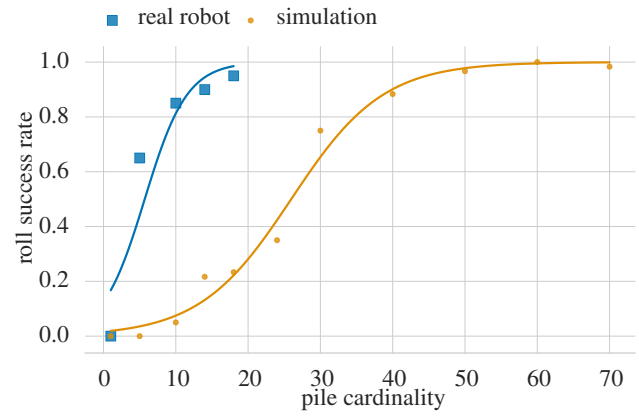
% 	\vspace{.3em}    
    \caption{Piles with more objects, larger cardinality, enable grasping with granular EC exploitation. The pile's opposing force increases as the pile cardinality increases, which we observed in the increase in roll success rates both in real-world (blue boxes) and in simulation (yellow dots), and with the fitted logical regression (solid lines) in both cases.}
    \label{fig:p1h1}
    \vspace{-1.em}
\end{figure}

We want to analyze the conditions under which a pile provides enough opposing force on a pushed object for grasping concerning a pile's cardinality. If this force is large enough to roll an object on the hand, then granular EC supports grasping and produces a contact event on a robot's end-effector. Previously, we established that the opposing force results from kinetic energy loss. Since friction dissipates kinetic energy, if we increase the number of objects, the number of contacts between objects increases, so more kinetic energy dissipates. Therefore, if we increase a pile's cardinality, the granular EC better supports grasping. We formally express our first hypothesis as:\\ \textbf{Hypothesis 1:} There is a positive correlation between grasp success rate and pile cardinality when granular EC enables open-loop grasping. 

To test our hypothesis, we observed the \emph{roll} success for piles with different cardinality, $|\text{Pile}| \in \{1,5,10,14,15\}$ of tennis balls of 180 grams on a horizontal surface. Since the object-centering effect of the EC is not in the scope of this experiment, we considered an attempt successful if one or more objects rolled on the hand.%or grasped.  

\begin{figure}[!th]
\centering
	\fontsize{9pt}{11pt}\selectfont
	\def\svgwidth{1.\linewidth}
 \includegraphics[width=\linewidth]{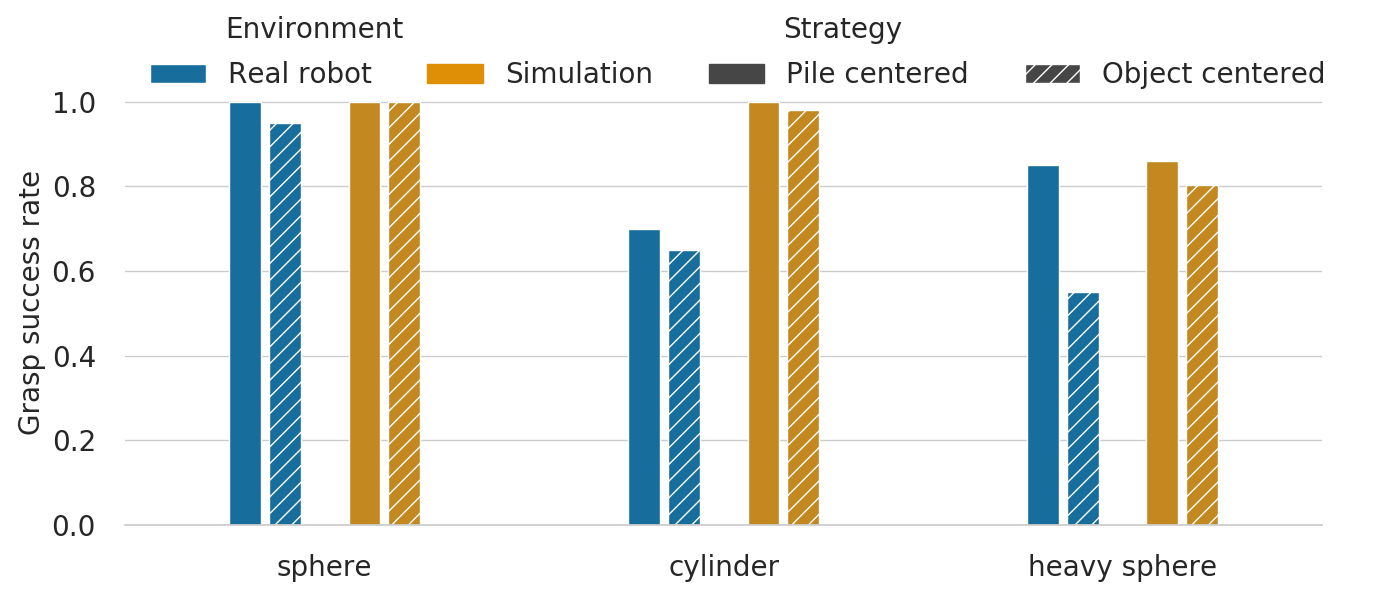}
% 	\vspace{.3em}    
    \caption[Evaluation of pile- and object-centered grasping.]{Both real robot (blue) and simulated (yellow) experiments strongly indicate that granular ECs can replace controlling object centering. The pile-centered grasp strategy (full bars) performs as well as the object-centered strategy (dashed bars), but pile-centered is somewhat better for heavy spheres.}
    \label{fig:p2h1}
%    \vspace{-1.2em}
\end{figure}

Figure~\ref{fig:p1h1} shows the roll success rate (markers) for different pile cardinalities and the estimated logistic regression model (solid line). The roll success rate monotonically increased with the pile cardinality. The positive correlation with the Spearman coefficient of 0.99 is a very strong indication that the opposing force increases for larger piles. 

The results also indicate a minimum pile cardinality for which the opposing force was large enough to support grasping with a simple straight pushing motion. We observed the same behavior in the simulation, but the minimum pile's cardinality was larger. We account for this difference to inaccuracies in friction and kinetic energy loss.
 
%-------------------------------------------------
\subsubsection{Pile Dynamics Centers Object for Grasping\\}
\label{sec: exp type 1 shifting computation}
We want to show the existence of the granular EC's object-centering effect. If an object gets centered on the hand without actively controlling its relative position, then granular EC provides a manipulation funnel that reduces the uncertainty of object-hand positioning.

To analyze the object-centering effect, we define two instances of the above-described strategy: the \emph{pile-centered} and the \emph{object-centered} instances. The two instances differ in the way the hand approaches a pile. With the pile-centered strategy, a robot approaches a pile object agnostic by centering the hand on the pile without considering any object's position. It assumes object centering and stabilization occur without actively controlling which object to be pushed on the pile's perimeter and without controlling the motion of the pushed object. Hence, the hand is lowered to a predefined location on a table in front of a pile. Even if the hand's pose is constant (same position, $\alpha$, $\beta$, and $\theta$ angles) between executions, the hand's relative pose to objects is random because we build random piles.

In contrast, the object-centered strategy aligns the hand with an object during the approach phase. This strategy assumes that the centered object stabilizes and is grasped. In advance, an expert tested iteratively different hand alignments with 18 tennis balls in a pile. A chosen ball's center was aligned with each gap between the fingers and the tip of the index or middle finger. The best alignment was when the index and middle finger gap aligned with the object's center. This outcome was used for piles of any size and all object types when executing the object-centered strategy. In each experiment, the expert chose an object on the pile's perimeter and then horizontally aligned the hand to the object before the hand slid toward the pile. 

We executed the pile- and object-centered strategies for various piles and considered an attempt successful only if one object is grasped. Identical grasp success rates between the two strategies would indicate that object centering can be replaced by granular ECE. Therefore, we propose the following \textbf{null hypothesis 2}: The \emph{object-centered} strategy is better than the \emph{pile-centered} strategy\\
$$\mu(\text{GS}_\text{object centered}) > \mu(\text{GS}_\text{pile centered}),$$
where $\mu(\text{GS})$ is the mean grasp success (GS).

We expect two possible outcomes. First, \textbf{Hypothesis 2a:} There is no significant difference between the grasp success rates for the two strategies:
$$\mu(\text{GS}_\text{object centered}) = \mu(\text{GS}_\text{pile centered}).$$
Secondly, \textbf{Hypothesis 2b:} The pile-centered strategy better performs than the object-centered one:
$$\mu(\text{GS}_\text{object centered}) < \mu(\text{GS}_\text{pile centered}).$$

We sampled the grasp success rate for three objects: light and heavy tennis balls and cylinders. The piles had 18 balls or 14 cylinders. Note that the hand can grasp two tennis balls and more than three cylinders. We built the pile beside a wall to increase the strength of the opposing force, which we analyze in detail below.%in Section~\ref{sec:eval p3}.  

The results in Figure~\ref{fig:p2h1} show no statistical differences in grasp success rates between pile- or object-centered strategies for light tennis balls and cylinders, supporting that the granular ECE provided object-centering. Moreover, we found weak evidence indicating that the pile-centered strategy was better than the object-centered one for heavy tennis balls, where the p-value is 0.041 from a single-sided Fisher's exact test. This suggests that it was difficult for the expert to properly model the pile's dynamics and center the hand relative to an object, while the granular ECE  provided better object centering.

%-------------------------------------------------
\subsubsection{Combining Geometrical and Granular ECE Enables Grasping From Small Piles\\}
\label{sec:eval p3}

\begin{figure}[t]
\centering
	\fontsize{9pt}{11pt}\selectfont
	\def\svgwidth{1.\linewidth}
	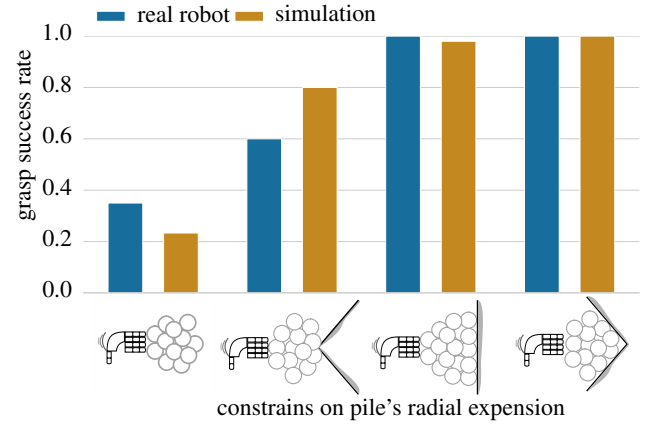
% 	\vspace{.3em}    
    \caption[Evaluation of the interaction between geometrical and granular ECs.]{Static walls  (geometrical ECs) increase the opposing force in small piles and enable grasping with granular EC exploitation.}
    \label{fig:p3h1}
    \vspace{-1.2em}
\end{figure}

We want to analyze the interaction between geometrical and granular ECs to identify beneficial combinations for grasping. If we can identify a beneficial transition between the two ECs, the granular EC complements the contextual information of geometrical EC about grasp affordances. 

We proposed that a pile's penetration resistance increases when the hand moves toward a vertical wall because objects lose kinetic energy via environmental interactions. So, we test \textbf{Hypothesis 3:} Grasp success rate increases when leveraging the granular EC in a pile that alone does not support grasping if the pile's expansion is constrained increasingly in opposition to the hand motion.

%To analyze our hypothesis, 
We sampled grasp success rates from increasingly constrained piles to expand. We constrained piles' expansion with two walls and changed the angle between the walls to increase the constraining effect. When there are no walls, a pile is the least constrained, 270$^{\circ}$ corner angle is less constrained than 180$^{\circ}$, and 90$^{\circ}$ corner angle is the most constrained case, as shown at the bottom in Figure~\ref{fig:p3h1}. We grasped from piles of 18 light tennis balls using the pile-centered strategy, which involves sliding the hand along a corner's bisector on the horizontal surface. We slid the hand along the bisector so that the wall's normal forces were symmetrical on both sides of the hand. We expect that grasp success will increase as the pile's expansion is increasingly constrained.

The results in Figure~\ref{fig:p3h1} show an increase in grasp success rate as the environment constrains the pile increasingly. Both real-world and simulation results show a strong correlation between grasp success and environment support, with Spearman coefficient $>0.94$. Interestingly, a single vertical wall (or two walls with 180$^{\circ}$) provided enough additional forces to achieve robust grasping from small piles.

In summary, simple and complex contact dynamics can provide ECs. We used geometrical and granular ECs for bin picking, simplifying computationally visual perception, grasp planning, and motion generation. Our study also illustrates a problem category where simply sequencing and executing ECE policies solves the motion generation problem without computing motion plans. 

\section{Conclusion}
\label{sec:conclusion}
%%======================================================================

This paper integrated Environmental Constraint Exploitation (ECE) into robot motion planning to address the challenges of computational complexity and uncertainty in high-dimensional and complex configuration spaces. By rethinking traditional collision-free planning constraints, we demonstrated how deliberate contact with the environment simplifies motion planning through dimensionality reduction and interleaving free-space and contact-exploiting exploration. Our methods, such as the newly enhanced Contact Exploiting RRT (CERRT) and Contingency CERRT planners, utilize ECE to structure the search space, making motion planning problems in complex environments computationally tractable while maintaining robustness under motion and proprioception uncertain.

Our results show that ECE-based motion planning simplifies the search process and improves execution robustness through contact-based uncertainty reduction. Leveraging ECE graphs to guide exploration toward task-relevant regions minimizes unnecessary computation and enhances efficiency. The practical utility of ECE was further validated through a real-world bin-picking application, where we characterized a novel environmental constraint and demonstrated its benefits in grasping scenarios with complex spatial interactions.

Looking forward, ECE presents a promising paradigm that extends beyond motion planning to other facets of robotics, including control and perception. Future research could investigate the integration of ECE with advanced sensing and learning frameworks to further enhance adaptability in dynamic environments.

%% use section* for acknowledgment
%\section*{Acknowledgment}
%\todo{should include Alumni's who's work helped: Arne Sieverling, Clemens Eppner, Can Erdogan, Max Winkelmann.}
%
%We gratefully acknowledge the funding provided by by the European Commission (SOMA, \mbox{H2020-ICT-645599}) and the Deutsche Forschungsgemeinschaft (DFG, German Research Foundation) under Germany's Excellence Strategy - EXC 2002/1 "Science of Intelligence" - project number 390523135. 

%======================================================================
% Acknowledgments
%======================================================================
\begin{acks}
The authors wish to thank several laboratory member who conducted pilot studies and experiments: Arne Sieverling, Clemens Eppner, Can Erdogan, and Max Winkelmann.
\end{acks}

%======================================================================
% Funding Information
%======================================================================
\begin{funding}
We gratefully acknowledge the funding provided by by the European Commission (SOMA, \mbox{H2020-ICT-645599}) and the Deutsche Forschungsgemeinschaft (DFG, German Research Foundation) under Germany's Excellence Strategy - EXC 2002/1 "Science of Intelligence" - project number 390523135.
\end{funding}

%======================================================================
% Author Disclosure Statement}
%======================================================================
\begin{dci}
The authors declare that there is no conflict of interest.
\end{dci}

%======================================================================
% Bibliography
%======================================================================
%\pagebreak

%\balance
\flushend

\bibliographystyle{SageH}

% weird things happening on manuscriptcentral...
% so I had to copy-paste the .bbl content into this tex file
\bibliography{Thesis.fixed}

% that's all folks
\end{document}